\documentclass{article} 
\usepackage{iclr2026_conference,times}

\usepackage{amsmath,amsfonts,bm}

\def\eqref#1{equation~\ref{#1}}

\def\1{\bm{1}}

\DeclareMathAlphabet{\mathsfit}{\encodingdefault}{\sfdefault}{m}{sl}
\SetMathAlphabet{\mathsfit}{bold}{\encodingdefault}{\sfdefault}{bx}{n}

\usepackage{hyperref}
\usepackage{url}

\usepackage{multirow}
\usepackage{amsmath}
\usepackage{amssymb}
\usepackage{booktabs}
\usepackage{graphicx} 
\usepackage{enumitem}
\usepackage{xcolor}
\usepackage[dvipsnames]{xcolor}
\usepackage{kotex}
\usepackage{float}
\usepackage{caption}
\usepackage{subcaption}
\usepackage{booktabs}
\graphicspath {{figures/}}
\usepackage{natbib}
\usepackage{cuted}
\usepackage{tcolorbox}
\usepackage{xspace}
\usepackage{ntheorem}
\usepackage{wrapfig}
\theoremseparator{.}
\newtheorem{theorem}{Theorem}

\makeatletter
\DeclareRobustCommand\onedot{\futurelet\@let@token\@onedot}
\def\@onedot{\ifx\@let@token.\else.\null\fi\xspace}
\def\eg{\emph{e.g}\onedot} 
\def\ie{\emph{i.e}\onedot}

\makeatother

\title{Impact of Regularization on Calibration and Robustness: From the Representation Space Perspective}

\author{
    \hspace{-0.5em}
    \textbf{Jonghyun Park, Juyeop Kim, Jong-Seok Lee}\\
    Yonsei University, Korea\\
    \texttt{\{jongpark.seoul, juyeopkim, jong-seok.lee\}@yonsei.ac.kr} \\
}

\iclrfinalcopy 
\begin{document}

\maketitle

\begin{abstract}
Recent studies have shown that regularization techniques using soft labels, \eg, label smoothing, Mixup, and CutMix, not only enhance image classification accuracy but also mitigate miscalibration due to overconfident predictions, and improve robustness against adversarial attacks. However, the underlying mechanisms of such improvements remain underexplored. In this paper, we offer a novel explanation from the perspective of the representation space (\ie, the space of the features obtained at the penultimate layer). Based on examination of decision boundaries and structure of features (or representation vectors), our study investigates confidence contours and gradient directions within the representation space. Furthermore, we analyze the adjustments in feature distributions due to regularization in relation to these contours and directions, from which we uncover central mechanisms inducing improved calibration and robustness. Our findings provide new insights into the characteristics of the high-dimensional representation space in relation to training and regularization using soft labels.
\end{abstract}


\section{Introduction}
\label{sec:introduction}

The motivation to improve the performance of classification models has led to the development of various regularization methods that use soft labels instead of one-hot encoded hard labels for classification targets. Representative methods include label smoothing \citep{szegedy2016rethinking}, Mixup \citep{zhang2018mixup}, and CutMix \citep{yun2019cutmix}. These techniques have demonstrated significant success in improving classification accuracy across various benchmarks.

However, their impact goes beyond accuracy improvement.
Studies have shown that these techniques contribute to better-calibrated models, aligning predicted probabilities more closely to actual accuracy \citep{guo2017calibration, muller2019does}.
Furthermore, they strengthen model robustness against gradient-based adversarial attacks, where imperceptible noise is added to input data to mislead models \citep{Goodfellow2014ExplainingAH, yun2019cutmix, Fu2020LabelSA, zhang2021how}.

While the benefits of soft labels are evident, the underlying mechanisms by which they achieve these improvements remain largely unexplained. 
In this paper, we offer a deeper understanding of how soft labels mitigate overconfident predictions and enhance adversarial robustness by \emph{examining the model’s representation space}.
Intuitively, data points that are correctly classified with lower confidence are located near decision boundaries, making them more vulnerable to adversarial perturbations \citep{hein2019relu, kim2024curved}.
Therefore, there exists a contradiction. If regularization alleviates overconfident predictions, features are expected to be located closer to the decision boundaries. Then, how is robustness to adversarial attacks enhanced?

To resolve this contradiction, we investigate the core mechanisms underlying model behavior in terms of calibration and robustness. For calibration, we explore the distribution of confidence contours and features, \ie, the outputs of the penultimate layer within the decision boundaries. This is crucial as calibration is a measurement on the overconfidence or underconfidence of predictions.
For robustness to gradient-based adversarial attacks, we focus on gradient directions and feature distributions, given that perturbations are generated based on gradients of the loss function.

We analyze the characteristics of decision boundaries, confidence contours, and gradients, in both visualizable low-dimensional representation spaces and the original high-dimensional spaces.
Based on the results, we study how the feature distribution in the representation space is modified by the use of soft labels, and how such changes can improve calibration and robustness simultaneously. Our analysis spans a wide range of models in order to obtain consistent findings.

Our work can be summarized as follows:

\begin{enumerate}
\item Building on prior work that focused on decision boundaries, we turn our attention to confidence contours and gradient directions—the characteristics that have a direct impact on confidence calibration and robustness to gradient-based adversarial attacks. Our analysis shows that decision regions and their \textbf{confidence contours form cone-shaped structures around the origin}, while \textbf{gradients for the cross-entropy loss radiate outward from the minimal-loss point}.

\item Observing how regularization modifies the feature distribution within the representation space, we find that the magnitudes of the features are reduced, leading to tighter clustering. Using the findings above, we explain why regularization using soft labels leads to less confident predictions and improved robustness.
    We show that \textbf{feature vectors with smaller magnitudes improve model calibration}, as reducing the feature magnitude acts similarly to temperature scaling, a common post-hoc calibration method.
    Furthermore, by analyzing gradient directions in the representation space, we show that \textbf{smaller features tend to be distributed in robust regions}, which align better with the class center vector.
\end{enumerate}

\section{Related Work}
\label{sec:related_work}

\textbf{Calibration and robustness.} Calibration refers to the alignment between a model's confidence and its actual accuracy.
\citet{guo2017calibration} found that modern neural networks exhibit overconfidence, leading to miscalibrated predictions.
To address this, various techniques have been proposed, including temperature scaling \citep{guo2017calibration}.


Simultaneously, neural networks’ vulnerability to adversarial attacks—imperceptible input perturbations causing misclassification—has gained attention \citep{Szegedy2013IntriguingPO}. An example is the Fast Gradient Sign Method (FGSM) \citep{Goodfellow2014ExplainingAH}, which prompted extensive research into various attacks and defenses \citep{carlini2017towards, madry2018towards, croce2020reliable, deng2024understanding}.


\noindent \textbf{Regularization techniques.} Regularization techniques such as label smoothing, Mixup, and CutMix have been shown to enhance model generalization. Label smoothing softens targets by distributing probability mass uniformly across labels \citep{szegedy2016rethinking}. Mixup linearly interpolates inputs and labels to create virtual training examples \citep{zhang2018mixup}, and CutMix replaces image regions with patches from other images, adjusting labels proportionally \citep{yun2019cutmix}. These methods have demonstrated improvements not only in generalization but also in calibration and adversarial robustness \citep{yun2019cutmix, Fu2020LabelSA, zhang2021how}.


Several studies explored reasons behind these improvements. \citet{thulasidasan2019mixup} showed that data augmentation in Mixup alone using hard labels does not improve calibration, highlighting the importance of soft labels. Recent visualizations indicate that Mixup clusters data near decision boundaries, reducing model overconfidence and thus improving calibration \citep{fisher2024pushing}. 
Regarding adversarial robustness, \citet{zhang2021how} demonstrated that minimizing Mixup loss approximates minimizing adversarial loss, enhancing robustness.
However, a comprehensive explanation linking soft labels to simultaneous calibration and robustness improvements from a representation perspective remains unexplored, motivating this study.

\noindent \textbf{Representation space.}
Several studies explored representation spaces in deep learning using 2-dimensional visualizations, uncovering radial feature distributions \citep{wen2016discriminative, wang2017normface, liu2017sphereface, chiranjeev2024hyperspacex}. \citet{luo2019strong} described cone-shaped decision regions due to radial distributions, emphasizing angularity between features. 
However, most studies omit the bias term in the classification layer, as the bias can be used to discriminate different classes over angularity \citep{wang2017normface}.
In our study, we demonstrate that these radial and cone-shaped structures persist regardless of bias terms if the number of classes is significantly smaller than the feature dimensionality.
Another notable concept is Neural Collapse, which shows that feature and weight vectors converge to an equiangular tight frame \citep{papyan2020prevalence}. 
Regarding the effect of regularization on the representation space, it was observed that label smoothing brings features closer together, reducing overall prediction confidence \citep{muller2019does}. 
However, no prior studies have resolved the contradiction between less confident predictions and stronger robustness from such representations.

\section{Representation Space}
\label{sec:3}

In this section, we investigate the representation space to show how decision regions, confidence contours, and gradient directions are formed and explain the mechanisms behind the formation of these shapes and directions.

\subsection{Decision Regions}
\label{sec:3.1}

\begin{tcolorbox}[
    colback=gray!15,
    colframe=gray!70!black,
    boxrule=1pt,
    arc=5pt,
    left=3pt,
    right=3pt,
    top=2pt,
    bottom=1pt
]
\textbf{Key Takeaway:} Decision regions of classification models form cone-like shapes centered around the origin when the dimensionality of the representation space is sufficiently high.
\end{tcolorbox}

A classification model typically consists of a feature extractor that maps inputs into features and a classification layer that uses those features to make decisions. The representation space of the model refers to the space where the output of the feature extractor, or more specifically, the output of the penultimate layer of the model resides \citep{kim2024curved}. It usually has high dimensionality (\eg, 2048 for ResNet50 \citep{he2016deep} and 768 for Swin-T \citep{liu2021swin}), making it challenging to visually analyze its characteristics. To address this, we first conduct intuitive analysis by transforming the representation space to 2D, then generalize the analysis to the original space. Implementation details regarding the transformation process for visualization can be found in Appendix~\ref{appendix:visualization}.

\begin{wrapfigure}{r}{0.5\textwidth}
  \centering
  \vspace{-1em}
  \begin{subfigure}[t]{.45\linewidth}
    \includegraphics[width=\linewidth]{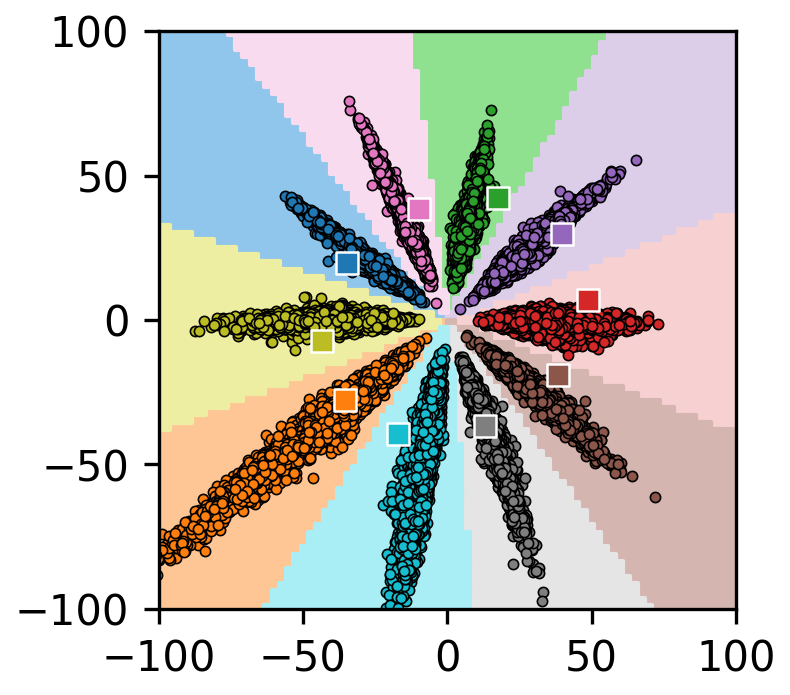}
    \caption{Without bias}
    \label{fig:without_bias}
  \end{subfigure}
  \begin{subfigure}[t]{.45\linewidth}
    \includegraphics[width=\linewidth]{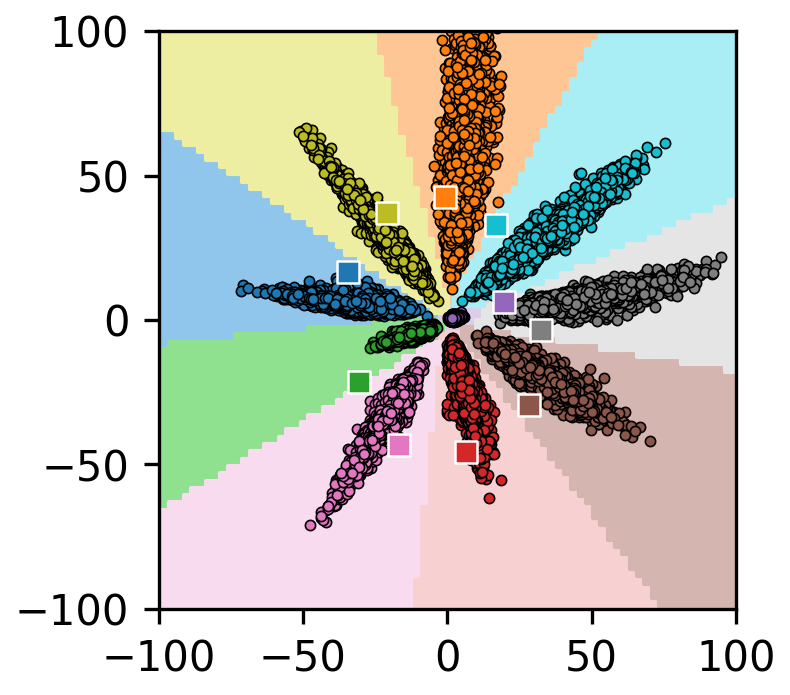}
    \caption{With bias}
    \label{fig:with_bias}
  \end{subfigure}
  
  \caption{%
    Decision regions and feature distributions:
    (a) without bias in the classification layer, and
    (b) with bias.  Without bias the regions and features are cone-shaped,
    radial for every class; with bias one class (purple) sits in the center
    with a circular region.  Squares are weight vectors.}
  \label{fig:boundary_comparison}
  \vspace{-1em}
\end{wrapfigure}
Fig.\ref{fig:boundary_comparison} visualizes ResNet50 trained on CIFAR-10 \citep{krizhevsky2009learning}, comparing results with and without bias terms in the classification layer. Without biases, decision regions form perfect circular sectors (\textit{cone-like shapes}) with radially distributed features (Fig.\ref{fig:without_bias}), which is consistent with prior studies \citep{wen2016discriminative, luo2019strong, chiranjeev2024hyperspacex}. When biases are included, however, this pattern does not hold for certain classes (Fig.~\ref{fig:with_bias}), prompting some studies to exclude biases to enforce radial distributions \citep{wang2017normface, liu2017sphereface}. Nevertheless, our analysis below demonstrates that radial distributions naturally emerge when feature dimensionality is sufficiently high, regardless of bias terms.


\noindent \textbf{How are decision regions shaped?}
To analyze the shape of decision regions, we examine how features are processed by the classification layer, focusing on factors influencing logit calculations. This is because a feature is assigned to the class with the highest logit value. Given a feature $\mathbf{f}$, the logit $\ell_c(\mathbf{f})$ for class $c$ is expressed as follows.
\begin{equation}
\label{eq:1}
\ell_c(\mathbf{f}) = \mathbf{w}_c^T \mathbf{f} + b_c = || \mathbf{w}_c || \cdot || \mathbf{f} || \cos\theta + b_c,
\end{equation}
where $\mathbf{w}_c$ and $b_c$ are the weight vector and the bias of the classification layer for class $c$, and $\theta$ is the angle between $\mathbf{f}$ and $\mathbf{w}_c$.
Thus, the elements that affect prediction results are $||\mathbf{w}_c||$, $\cos \theta$, and $b_c$.


In the study of \citet{papyan2020prevalence}, it was observed that class weight norms $||\mathbf{w}_c||$ become similar, as shown in Fig.~\ref{fig:boundary_comparison}. Thus, without bias terms, predictions mainly depend on $\cos \theta$. In other words, decision boundaries are formed based on the degree of alignment with the weight vector, and consequently, the decision regions take the shape of cones centered at the origin. 
This phenomenon is observed regardless of weight initialization (see Fig. \ref{fig:2} in Appendix~\ref{appendix:initialization}).
Further verifications of these cone-shaped regions are included in Appendix~\ref{appendix:more_example_2d}.

\begin{figure*}[!t]
    \centering
    \subfloat[]
    {
        {\includegraphics[width=0.25\textwidth]{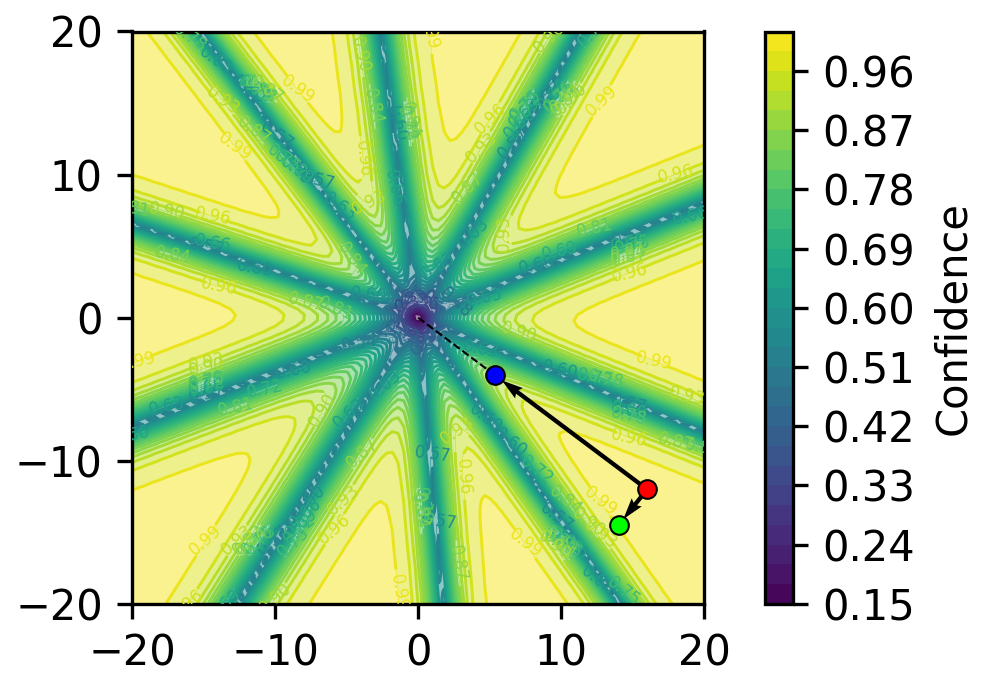}}
        \label{fig:conf_grad_a}
    }
    \subfloat[]
    {
        {\includegraphics[width=0.24\textwidth]{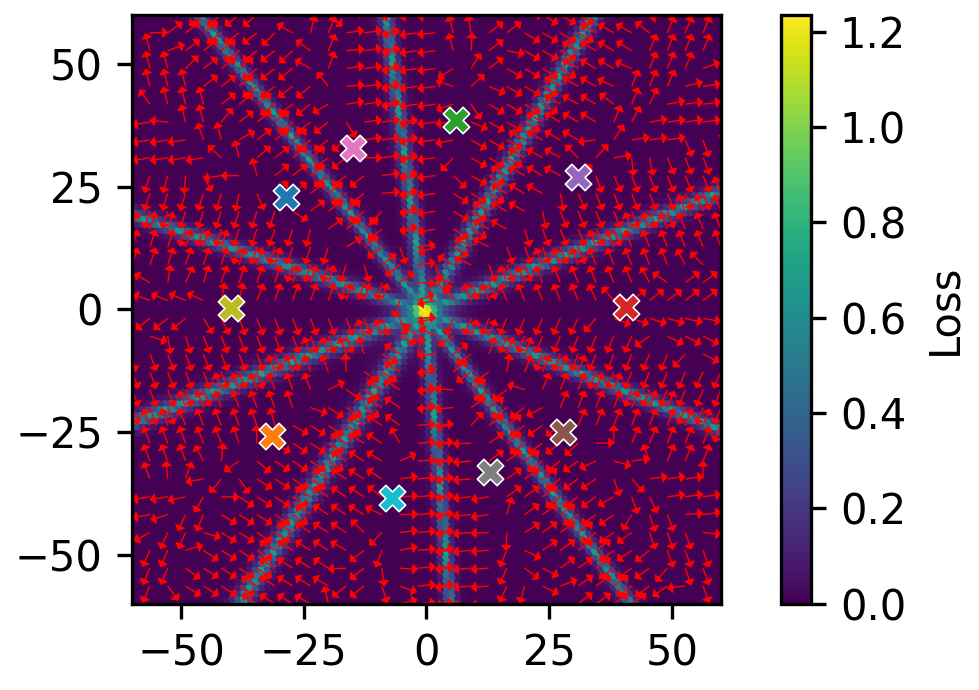}}
        \label{fig:conf_grad_b}
    }
    \subfloat[]
    {
        {\includegraphics[width=0.225\textwidth]{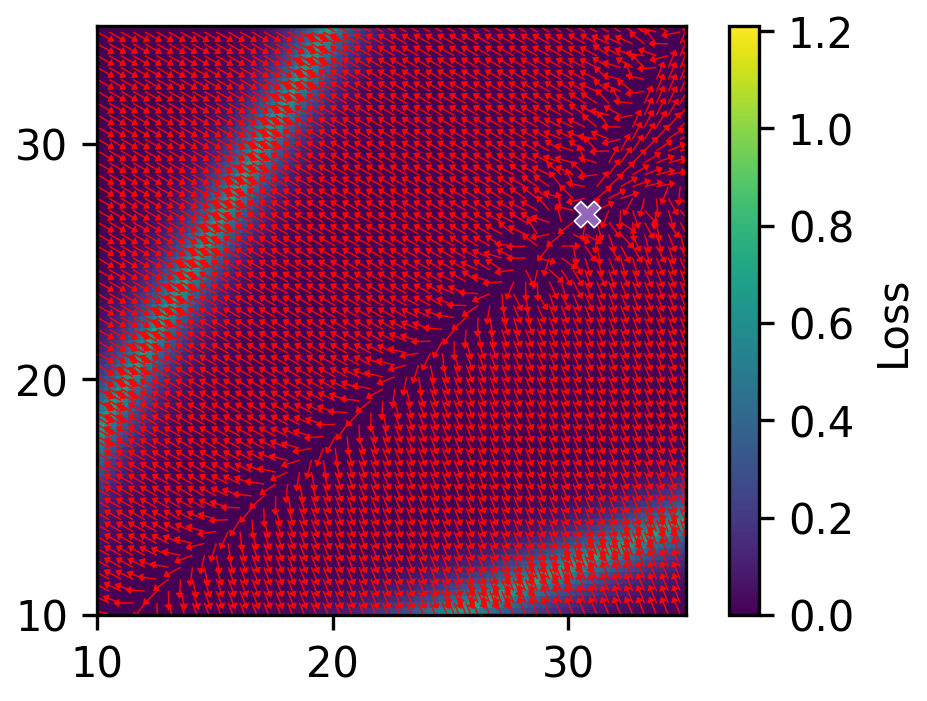}}
        \label{fig:conf_grad_c}
    }
    \subfloat[]
    {
        {\includegraphics[width=0.18\textwidth]{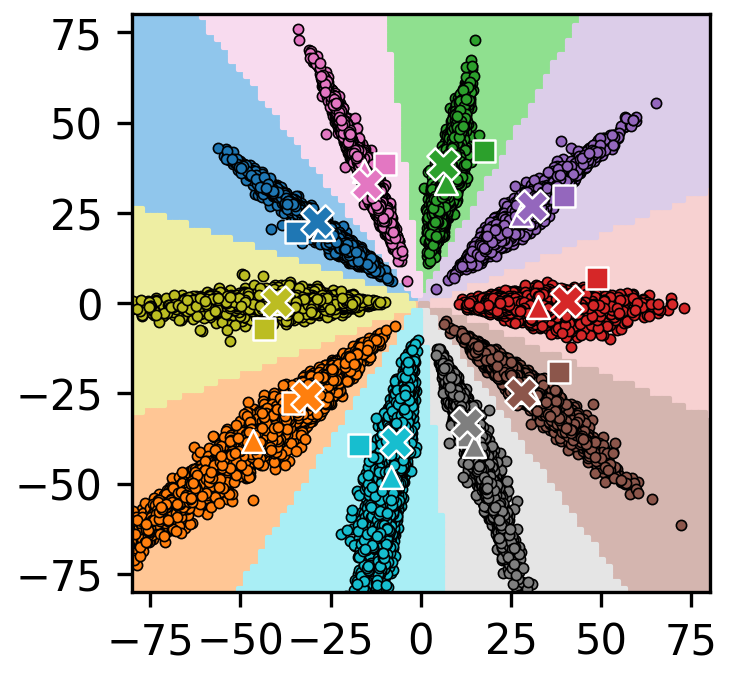}}
        \label{fig:class_center_options}
    }
    
    \caption{2D representation space of ResNet50 on CIFAR-10. 
    Each cross mark represents the location with the lowest loss for each class.
    (a) Confidence contours.
    (b) Loss and gradient directions.
    (c) Enlarged version of (b).
    (d) Features (circles), class means (triangles), and weight vectors (squares). A cross mark indicates the minimum loss point for each class.
    }
    \vspace{-1em}
    \label{fig:conf_grad}
    
\end{figure*}

For models with bias terms in the classification layer, the shape of decision regions can differ, \eg, a circular decision region at the center in Fig.~\ref{fig:with_bias}. In this case, because bias values can be used for prediction, the purple and gray classes can be correctly classified although their weight vectors have similar directions.
The appearance of a non-cone-shaped decision boundary is likely due to the difficulty that models face when trying to fit multiple classes into a cone-like structure within a constrained dimensionality, such as fitting 10 classes (or even 100 classes for CIFAR-100 \citep{krizhevsky2009learning}) into the 2D space.

Therefore, for original models, where the dimensionality of the representation space is sufficiently high, the cone shapes of the decision regions hold consistently even with the presence of bias terms. Since visual inspection is not possible in the original high-dimensional space, we take an alternative approach to verify it, whose results are presented in Tabs.~\ref{tab:1} and \ref{tab:appendix_bias_check} in Appendices~\ref{appendix:original_boundaries} and \ref{appendix:bias}.

\subsection{Confidence Contours, Gradient Directions}
\label{sec:3.2}

\begin{tcolorbox}[
    colback=gray!15,
    colframe=gray!70!black,
    boxrule=1pt,
    arc=5pt,
    left=3pt,
    right=3pt,
    top=2pt,
    bottom=1pt
]
\textbf{Key Takeaway:} Confidence contours and gradient directions radiate outward from minimal-loss points, indicating the convexity of the loss landscape.
\end{tcolorbox}

Although research on decision boundaries has been conducted \citep{wen2016discriminative, wang2017normface, liu2017sphereface, luo2019strong, chiranjeev2024hyperspacex}, the shape of confidence contours and loss gradient directions within these boundaries have been rarely examined. It is essential to study these characteristics because calibration measures how overconfident or underconfident predictions are and gradient-based adversarial attacks create perturbations based on the loss function.

\noindent \textbf{Confidence Contours}. In Fig.~\ref{fig:conf_grad_a}, we present the confidence contours in each decision region in the 2D representation space. If the logit for a particular class is large in comparison to the others, the confidence in the prediction is high. From Eq.~\ref{eq:1}, the logit for a particular class is large if the feature norm is large or if the feature is well-aligned with the weight vector. This can be confirmed from the confidence contours shown in Fig.~\ref{fig:conf_grad_a}. The confidence of a feature (red dot) can be lowered in two ways: 1) by moving radially toward the origin (blue dot), which reduces the feature norm, or 2) by moving toward the nearest decision boundary (green dot), which deteriorates alignment with the weight vector.

\noindent \textbf{Gradient Directions}. The cross-entropy is convex with respect to logits \citep{boyd2004convex}. Since the logits in Eq.~\ref{eq:1} are obtained through an affine transformation of features, the cross-entropy loss is also convex with respect to the features.
To illustrate this, we examine the gradient directions in the 2D space of ResNet50 on CIFAR-10 in Figs.~\ref{fig:conf_grad_b} and \ref{fig:conf_grad_c}.
In areas with \textbf{high cosine similarity} to these minimal-loss points (\ie, points on the line between the origin and the minimal-loss point), gradient directions point toward the origin.
However, in regions with \textbf{low cosine similarity}, gradient directions appear nearly orthogonal to the direction of the origin, pointing toward the decision boundaries.
Thus the gradient direction for a feature differs depending on the cosine similarity between the feature and the minimal-loss point.
Overall, the gradient directions radiate outward from the minimal-loss points, highlighting the convex structure of the loss landscape.
Note that, to locate these minimal-loss points, we use gradient descent.

\section{Effect of Regularization}
\label{sec:4}

Based on the characteristics observed in Section~\ref{sec:3}, we examine how regularization modifies feature distributions in a way to reduce overconfident predictions and enhance robustness against adversarial attacks simultaneously. Through this analysis, we aim to address the contradiction against the common expectation that if a feature with lower confidence is located closer to the decision boundary, it is likely to be more vulnerable to adversarial perturbations.

\subsection{Measuring Feature Distributions}
\label{sec:3.3}

In a low-dimensional space, it is easy to identify the approximate distribution of features, as shown in Fig.~\ref{fig:boundary_comparison}. In a high-dimensional space, however, visualization becomes challenging, making it difficult to understand how features are distributed. \citet{papyan2020prevalence} measured the angular distance between features and the mean of features for each class, while some others \citep{liu2017sphereface, chiranjeev2024hyperspacex} measure angular distance of features from weight vectors. However, we argue that the angular distance should instead be measured between features and minimal-loss points.

To describe how features are distributed within the decision regions, it is natural to measure the proximity of a feature to the decision boundary based on its confidence. Following this idea, the cosine similarity of a feature with the center of the decision region of a class, noted as \textit{class center}, should have the strongest correlation with its confidence, as the center of the decision region is expected to have the highest confidence (farthest from the decision boundary). We examine three candidates for the class center: 1) the mean of correctly classified features within a class (\textit{class mean}), 2) the \textit{weight vector} of the classification layer, and 3) the point where the classification loss is the lowest (\textit{minimum loss point}).

\begin{wraptable}{r}{0.5\textwidth}
    \vspace{-1em}
  \centering
  \caption{
    Pearson correlation coefficient between confidence and cosine similarity of features to class means, weight vectors, and minimum loss points. The case with the highest correlation among the three candidates is marked in bold. We omit ConvNeXt-T results trained with CutMix, as training failed to converge.
    See Appendix~\ref{appendix:implementation} for training details.
  }
  \vspace{-0.2em}
  \resizebox{\linewidth}{!}{ 
    \renewcommand{\arraystretch}{1.2}
    \setlength{\tabcolsep}{5pt}
    \begin{tabular}{c c c c c}
      \toprule
      \multirow{2}{*}{\bf Model} & 
      \multirow{2}{*}{\bf Method} 
      & \bf Class & 
      \bf Weight & 
      \bf Minimum \\ 
      & & \bf Mean & \bf Vector & \bf Loss Point \\
      \hline
      \multirow{4}{*}{ResNet50}
        & Baseline & 0.35 & 0.52 & \bf 0.56 \\
        & Label smoothing & 0.36 & 0.61 & \bf 0.70 \\
        & Mixup & 0.36 & 0.60 & \bf 0.75 \\
        & CutMix & 0.39 & 0.52 & \bf 0.65 \\ \hline
      
      \multirow{4}{*}{Swin-T}
        & Baseline & 0.41 & 0.55 & \bf 0.56  \\
        & Label smoothing & 0.56 & \bf 0.64 & \bf 0.64  \\
        & Mixup & 0.48 & 0.63 & \bf 0.64  \\
        & CutMix & 0.48 & \bf 0.58 & \bf 0.58  \\ \hline
      
      \multirow{4}{*}{MobileNetV2}
        & Baseline & 0.41 & 0.35 & \bf 0.44 \\
        & Label smoothing & 0.41 & 0.35 & \bf 0.44 \\
        & Mixup & 0.47 & 0.49 & \bf 0.74 \\
        & CutMix & 0.42 & 0.41 & \bf 0.70 \\ \hline
      
      \multirow{4}{*}{ConvNeXt-T}
        & Baseline & 0.26 & 0.35 & \bf 0.37 \\
        & Label smoothing & 0.73 & \bf 0.74 & \bf 0.74 \\
        & Mixup & 0.53 & \bf 0.57 & \bf 0.57  \\
      \bottomrule
    \end{tabular}
  }
  \label{tab:2}
  \vspace{-2em}
\end{wraptable}
Fig.~\ref{fig:class_center_options} illustrates the positions of these three candidates in the 2D representation space. While the positions of class means (triangles) seem reasonable, large errors could occur if outliers exist far from the feature clusters. As seen in the pink and brown classes, the weight vectors (squares) often show a significant error, making them unsuitable as the centers of the decision regions. The minimum loss points (crosses) appear to best represent the class center. More quantitative analysis is provided in Tab.~\ref{tab:2}, where the cosine similarity between features and minimum loss points shows the highest correlation with confidence. Note that the results in Tab.~\ref{tab:2} are from original models trained on ImageNet.

In addition, from the analysis regarding confidence in Section~\ref{sec:3.2}, features closer to the origin have lower confidence, as they are also closer to the decision boundaries.
Therefore, we determine how far a specific feature is from the decision boundary using two criteria: 1) \textbf{the root mean square (RMS) of the feature} (we use RMS as the feature magnitude to compensate for different dimensionalities across models) and 2) \textbf{the cosine similarity of the feature with the \textit{class center}} (crosses in Figs. \ref{fig:conf_grad_b}, \ref{fig:conf_grad_c}, and \ref{fig:class_center_options}), where the classification loss is minimal for that class.

\begin{figure*}[!t]
    \centering
    
    \subfloat[Baseline]
    {
        {\includegraphics[width=0.18\textwidth]{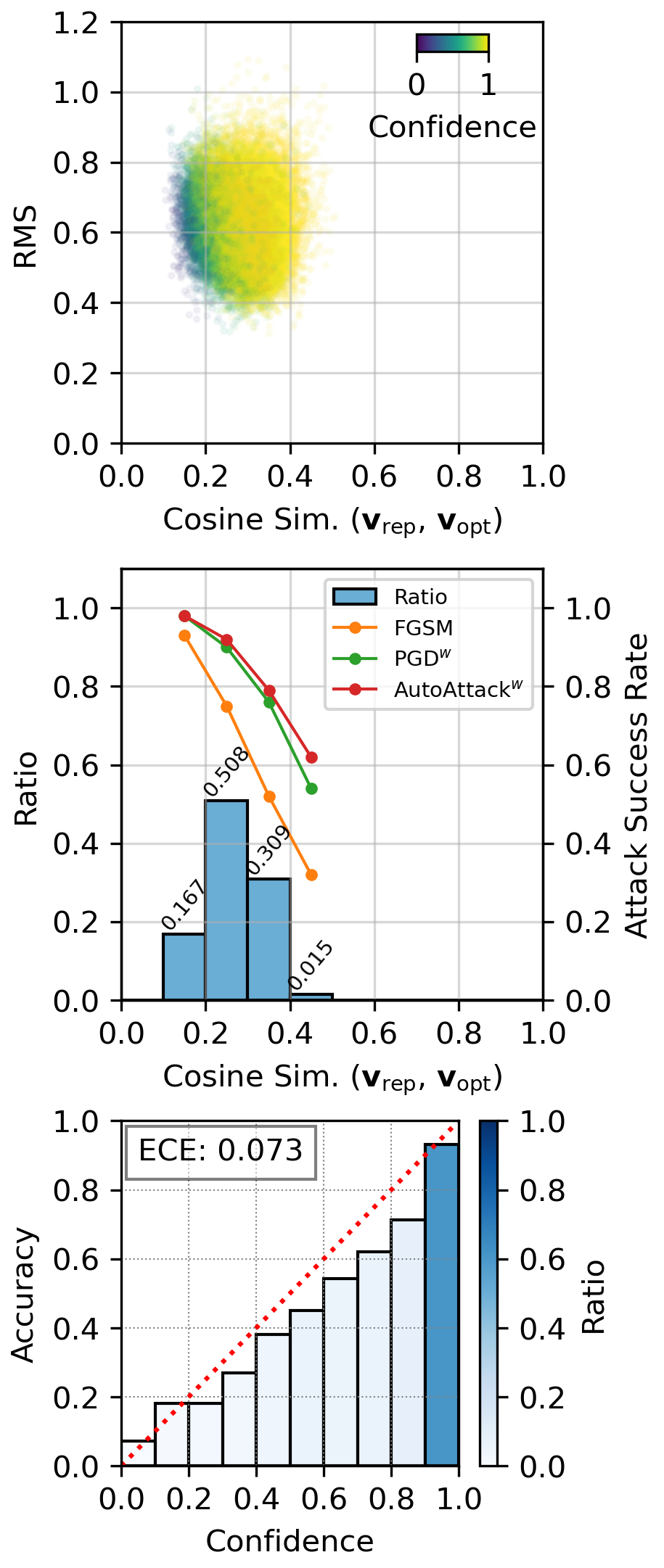}}
        \label{fig:5_a}
    }\hspace{-0.0cm}
    \subfloat[Label smoothing]
    {
        {\includegraphics[width=0.18\textwidth]{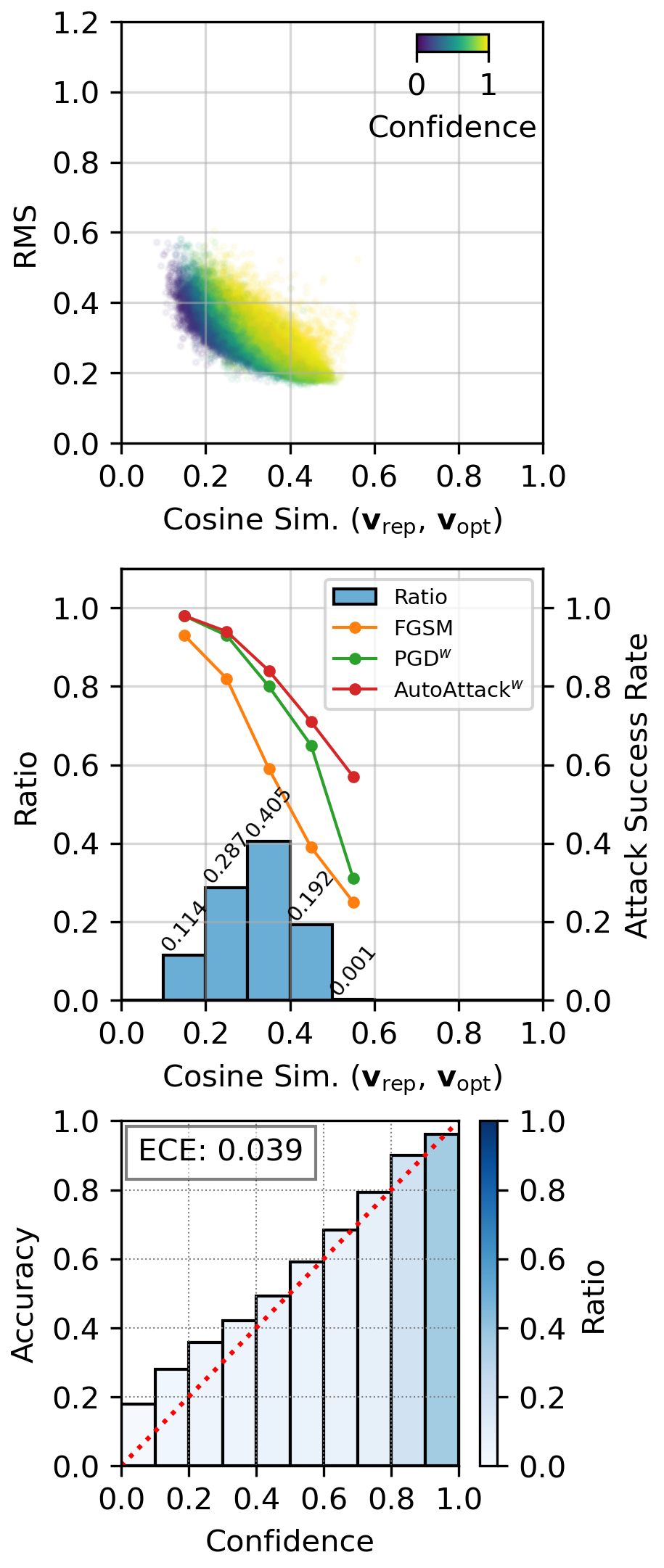}}
        \label{fig:5_b}
    }\hspace{-0.0cm}
    \subfloat[Mixup]
    {
        {\includegraphics[width=0.18\textwidth]{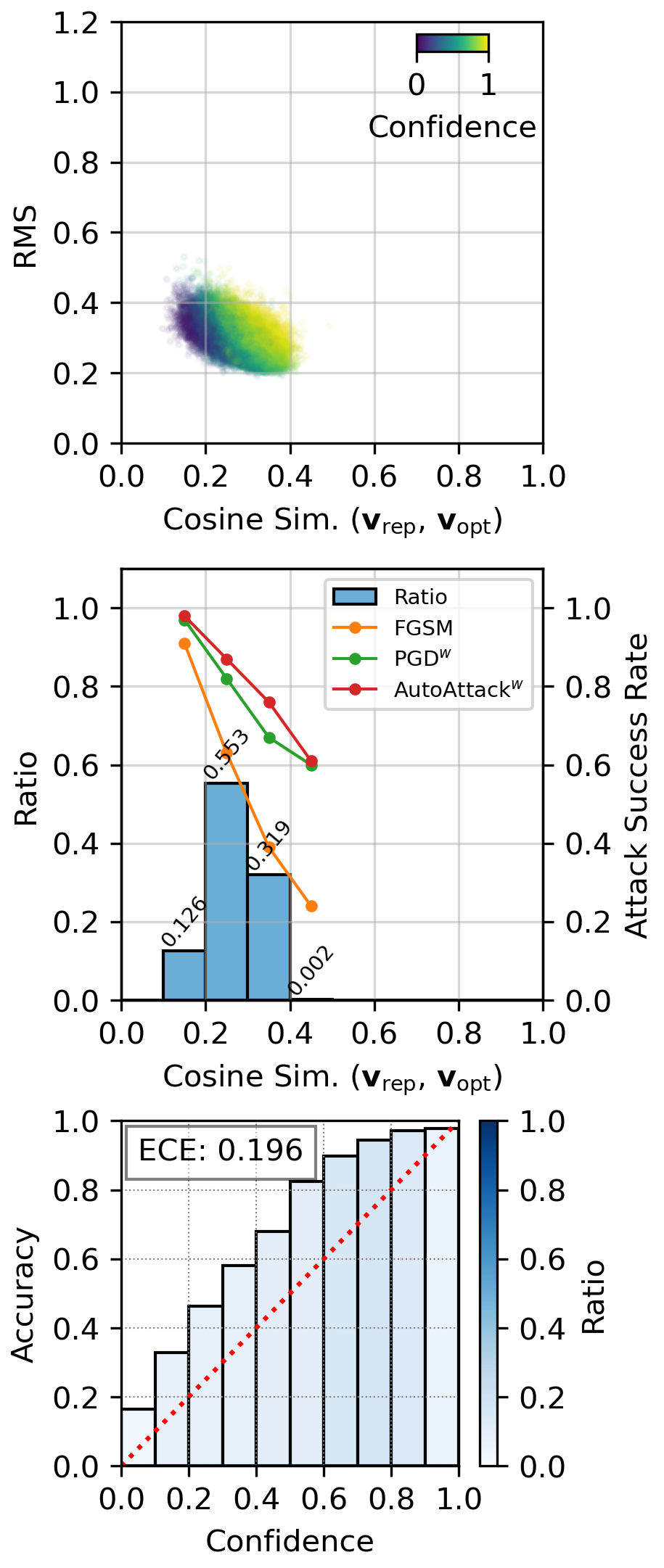}}
        \label{fig:5_c}
    }\hspace{-0.0cm}
    \subfloat[CutMix]
    {
        {\includegraphics[width=0.18\textwidth]{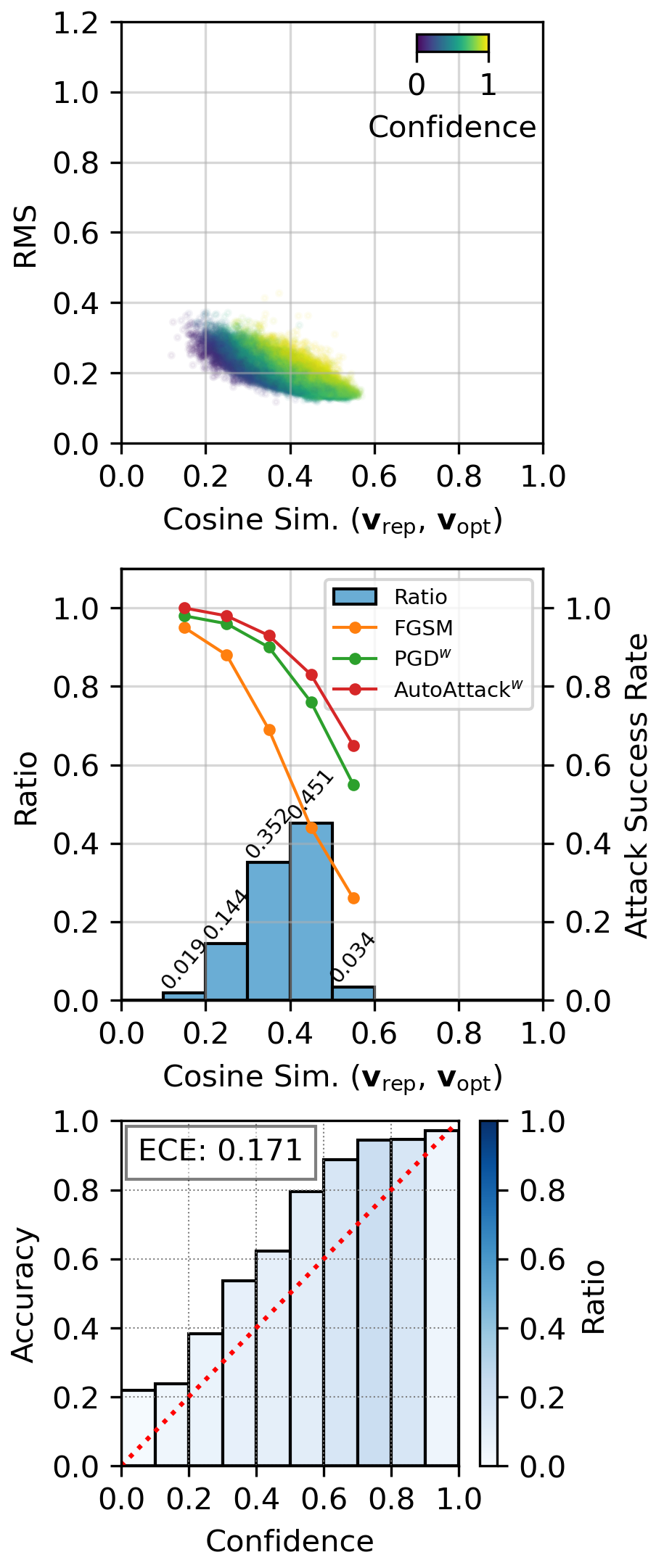}}
        \label{fig:5_d}
    }
    \caption{
    Evaluation results of ResNet50 on the ImageNet validation data.
    \textbf{Top.} Scatter plots of feature RMS and cosine similarities of features ($\mathbf{v}_\text{rep}$) with the class center ($\mathbf{v}_\text{opt}$). Colors represent confidence values.
    \textbf{Middle.} Histograms of cosine similarities of features to class centers, along with the attack success rates of FGSM, PGD$^\text{w}$, and AutoAttack$^\text{w}$ for each bin (hyperparameters settings for the attacks can be found in Section~\ref{sec:4_4}). 
    For results on PGD$^\text{s}$ and AutoAttack$^\text{s}$, see Fig.~\ref{fig:stronger_settings1} in Appendix~\ref{appendix:more_result_regularization}. 
    \textbf{Bottom.} Reliability diagrams, where the transparency of bars represents the ratio of data in each confidence bin. Expected calibration error (ECE) \citep{guo2017calibration} values are shown for each case.
    }
    \label{fig:5}
    \vspace{-1em}
\end{figure*}

We show the relationship of the confidence vs. the RMS of features and the cosine similarity of features with the class center in the original representation space in the top row of Fig.~\ref{fig:5} for ResNet50. As expected, the smaller the RMS of features or the lower the cosine similarity is, the lower the confidence is, indicating proximity to the decision boundary. The results for Swin-T, MobileNetV2, and ConvNeXt-T can be found in Appendix~\ref{appendix:more_result_regularization}, showing similar trends.

\subsection{Impact on Feature Distributions}
\label{sec:4_2}

\begin{tcolorbox}[
    colback=gray!15,
    colframe=gray!70!black,
    boxrule=1pt,
    arc=5pt,
    left=3pt,
    right=3pt,
    top=2pt,
    bottom=1pt
]
\textbf{Key Takeaway:} Regularization reduces feature norms and improves alignment with class centers, producing compact representations closer to the origin.
\end{tcolorbox}

\begin{figure}[!t]
  \centering

    \begin{subfigure}[t]{.21\columnwidth}
      \includegraphics[width=\columnwidth]{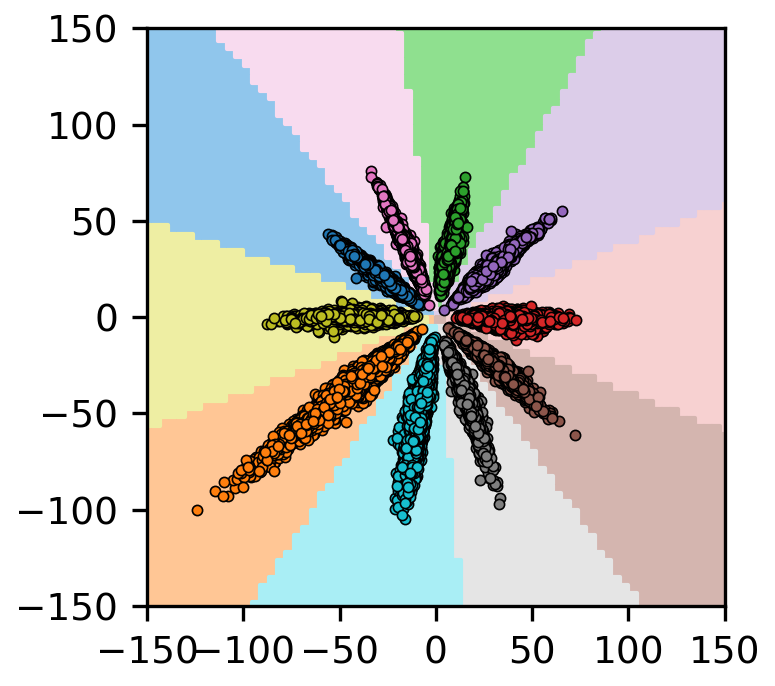}
      \caption{Baseline}
      \label{fig:rms_decrease_baseline}
    \end{subfigure}
    \begin{subfigure}[t]{.2\columnwidth}
      \includegraphics[width=\columnwidth]{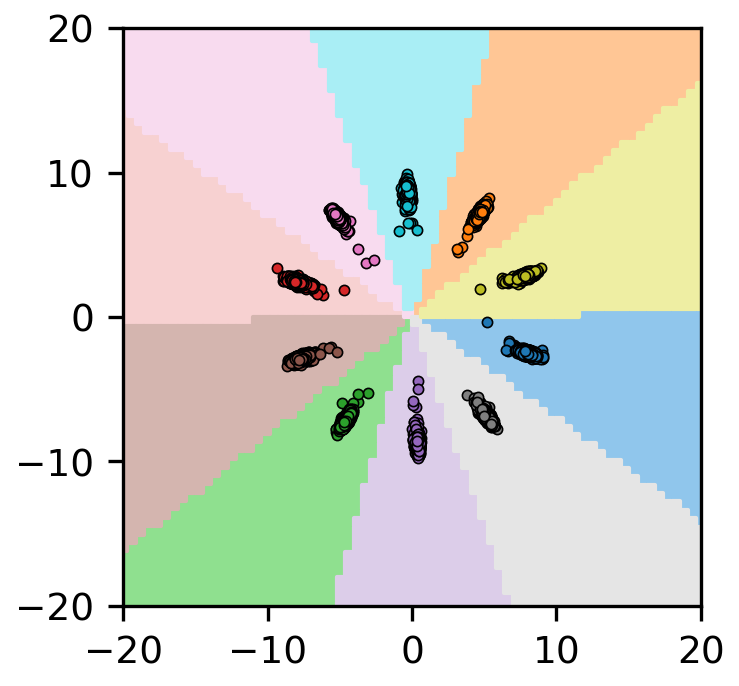}
      \caption{Label smoothing}
      \label{fig:rms_decrease_labelsmoothing}
    \end{subfigure}
    \begin{subfigure}[t]{.2\columnwidth}
      \includegraphics[width=\columnwidth]{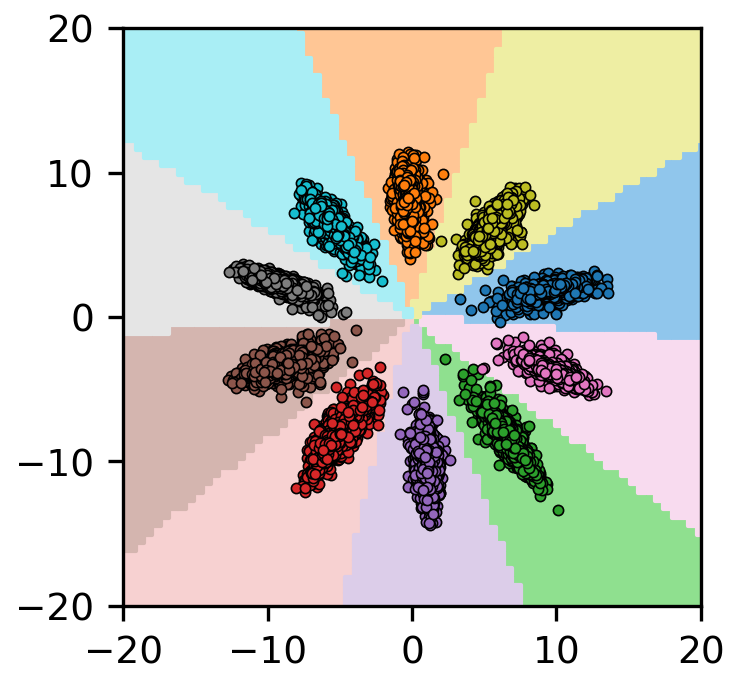}
      \caption{Mixup}
      \label{fig:rms_decrease_mixup}
    \end{subfigure}
    \begin{subfigure}[t]{.2\columnwidth}
      \includegraphics[width=\columnwidth]{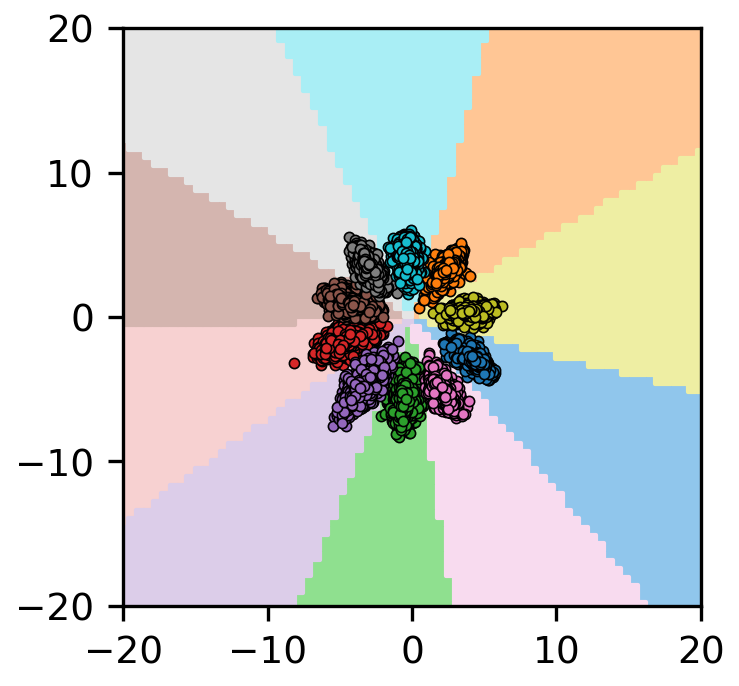}
      \caption{Mixup}
      \label{fig:rms_decrease_mixup}
    \end{subfigure}

  \caption{%
    Features in the 2D representation space for ResNet50 on CIFAR-10. Note that the scales differ across figures. See Appendix~\ref{appendix:more_example_2d} for further results.
    }
  \label{fig:rms_decrease_2d}
  \vspace{-1em}
\end{figure}

In Fig.~\ref{fig:rms_decrease_2d}, we present the results of training with and without regularization (label smoothing and Mixup) in the 2D representation space.
See Appendix~\ref{appendix:more_result_regularization} for complete results on different regularizations and models.
Training with regularization results in two notable changes. First, the RMS of features significantly decreases (Fig.~\ref{fig:rms_decrease_2d} and the top row of Fig.~\ref{fig:5}), bringing them closer to the origin. Second, from the middle row of Fig.~\ref{fig:5}, the proportion of data with high cosine similarity between the feature and the class center increases in the regularized models. (Note that the results in Fig.~\ref{fig:5} are from the original representation space, and these changes are consistently observed across various model architectures, as will be shown in Section~\ref{sec:4_5}.)  Detailed analysis is as follows.

\noindent \textbf{Decrease in RMS.} 
The following theorem explains the phenomenon of RMS decrease due to regularization.

\begin{theorem}
Assume $b_c \approx 0$ and $||\mathbf{w}_{c}|| \approx ||\mathbf{w}||$ for all classes $c$. Then, the optimal solution of training $\mathbf{f}$ with cross-entropy using hard labels, $\mathbf{f}^*_{hard}$, has a larger magnitude than that using soft labels, $\mathbf{f}^*_{soft}$, i.e.,
\begin{align}
    ||\mathbf{f}^*_{hard} || > ||\mathbf{f}^*_{soft} ||. \notag
\end{align}
\end{theorem}
The proof can be found in Appendix~\ref{appendix:proof}.

\begin{figure}[t]
\centering
\begin{minipage}[t]{0.45\textwidth}
  \centering
  \begin{subfigure}[t]{.48\linewidth}
    \includegraphics[width=\linewidth]{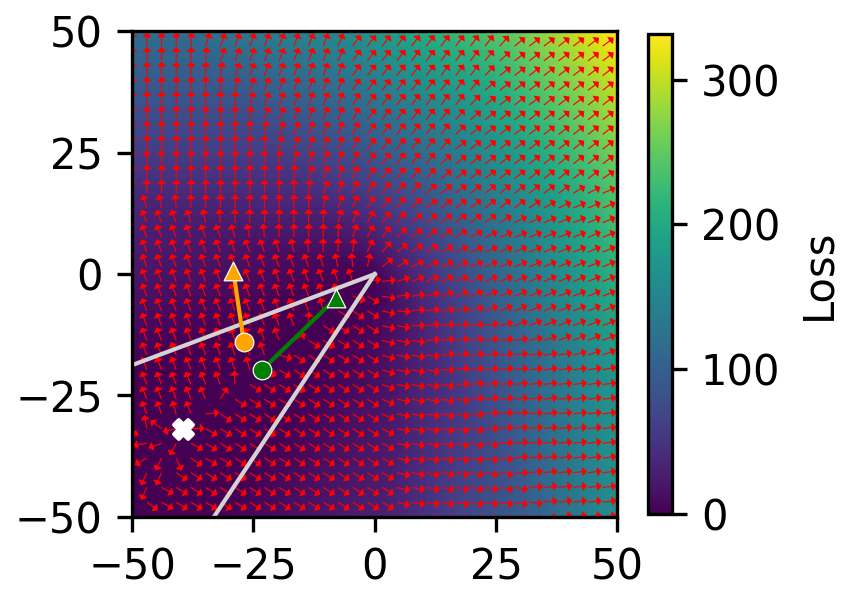}
    \caption{Baseline}
    \label{fig:7_a}
  \end{subfigure}
  \begin{subfigure}[t]{.48\linewidth}
    \includegraphics[width=\linewidth]{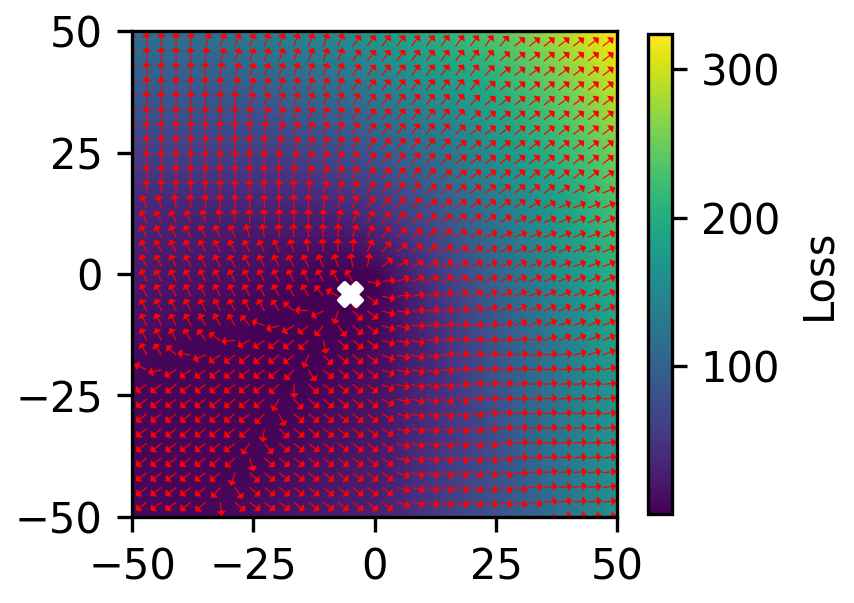}
    \caption{Mixup}
    \label{fig:7_b}
  \end{subfigure}
  \caption{Loss and gradient directions for a certain class in the 2D representation space of ResNet50 on CIFAR-10. 
  White crosses indicate the locations with the smallest loss. 
  Circles and triangles represent the features of clean and perturbed data, respectively. 
  White lines depict the decision boundary.}
  \label{fig:7}
\end{minipage}
\hfill
\begin{minipage}[t]{0.53\textwidth}
  \centering
  \begin{subfigure}[t]{.39\linewidth}
    \includegraphics[width=\linewidth]{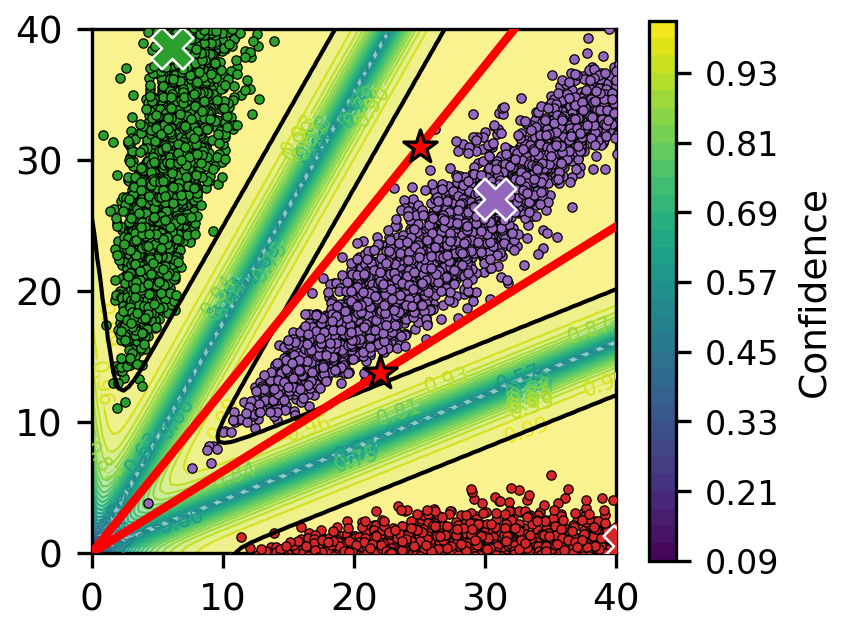}
    \caption{Baseline}
    \label{fig:8_a}
  \end{subfigure}
  \begin{subfigure}[t]{.39\linewidth}
    \includegraphics[width=\linewidth]{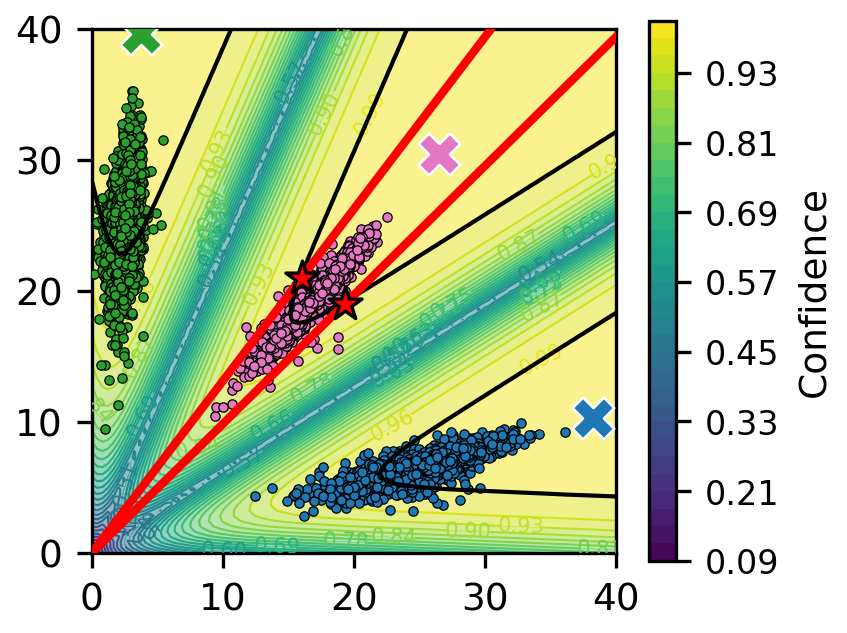}
    \caption{Mixup}
    \label{fig:8_b}
  \end{subfigure}
  \caption{Confidence contours and features (circled dots). 
  The 0.99 confidence contour is shown as a black line. 
  Crosses indicate the minimum loss points. 
  Red stars represent the features with confidence higher than 0.99 but with the poorest alignment to their class center in terms of cosine similarity; red lines connect them with the origin.}
  \label{fig:8}
\end{minipage}
\end{figure}

To confirm this, we visualize the cross-entropy loss and gradient directions for hard and soft labels in a 2D representation space in Figs.~\ref{fig:7_a} and \ref{fig:7_b}, respectively. The point with the smallest loss is marked with a white cross. We can see that for hard labels, the location of the cross mark is far from the origin, while for soft labels, it is near the origin.
The same phenomenon is observed in the original representation space. In Tab.~\ref{tab:3}, the RMS values consistently decrease when soft labels are used.

\noindent \textbf{Increase in cosine similarity.}
Fig.~\ref{fig:8} shows the confidence contours in the 2D representation space of ResNet50 trained with and without regularization on CIFAR-10. For both models, we compare how well a feature needs to be aligned with the class center (crosses) to achieve a certain confidence level. Specifically, we search for features with confidence higher than 0.99 but with the worst alignment to their class center in terms of cosine similarity, in both clockwise and counterclockwise directions (red stars). By connecting these features to the origin (red lines), we can see that the angle between the lines in the regularized model is smaller (Fig.~\ref{fig:8_b}). 
This occurs because near the origin, where the norm $|| \mathbf{f} ||$ is small in Eq.~\ref{eq:1}, a larger $\cos{\theta}$ is required to reach the same level of confidence (since confidence is positively correlated with logits). Consequently, the angular region that achieves a given confidence (\eg, 0.99) becomes narrower, leading to the observed smaller angle.
Therefore, features closer to the origin must be well-aligned with the class center to reach a given confidence level.
Conversely, a feature located far from the origin can still achieve high confidence without being as closely aligned with the class center as a feature located near the origin (Fig.~\ref{fig:8_a}).
As a result, in the regularized models producing features with small RMS, the cosine similarity of features with the class center is relatively high compared to the baseline models.
Tab.~\ref{tab:3} also confirms the increase of the cosine similarity by soft labels in the original representation spaces.

\subsection{Impact on Calibration}
\label{sec:4_3}

\begin{tcolorbox}[
    colback=gray!15,
    colframe=gray!70!black,
    boxrule=1pt,
    arc=5pt,
    left=3pt,
    right=3pt,
    top=2pt,
    bottom=1pt
]
\textbf{Key Takeaway:} Regularization through soft labels reduces overconfidence by scaling features closer to the origin, effectively calibrating predictions.
\end{tcolorbox}

In the bottom row of Fig.~\ref{fig:5}, models become less confident when regularized with soft labels, reducing miscalibration due to overconfidence in the baseline training. This can be explained in relation to the RMS decrease mentioned in Section~\ref{sec:4_2}.

When the magnitude of a feature $\mathbf{f}$ is decreased (red dot becoming the blue dot in Fig.~\ref{fig:conf_grad_a}) by a factor of $T$ due to training with regularization, its corresponding logit for class $c$ can be expressed as $\frac{{\mathbf{w}_c^T \mathbf{f}}}{T} + b_c$.
Due to the cone-shaped boundaries, vectors located on the line connecting a feature and the origin are mostly classified into the same class as the feature (Tab.~\ref{tab:1} in Appendix~\ref{appendix:original_boundaries}). Furthermore, features with smaller RMS have lower confidence values (Section~\ref{sec:3.2}). Therefore, if the magnitude of a feature vector is scaled down and the feature moves closer to the origin, the confidence of the feature decreases, but the prediction remains unchanged. 
We verify that the model confidence can be reduced, or even increased by solely scaling feature vectors without compromising classification accuracy in Fig.~\ref{fig:14} in Appendix~\ref{appendix:feature_scaling}.
This finding indicates that \textbf{decreasing magnitudes of features through soft labels enables calibration adjustment while preserving classification accuracy}.

In fact, the effect of feature scaling due to regularization is similar to the post-processing technique known as temperature scaling. Temperature scaling adjusts calibration by scaling the logit values by a factor of $T$, resulting in the logit expression $\frac{{\mathbf{w}_c^T \mathbf{f}+b_c}}{T}$, which is similar to the case of feature scaling. Mathematically, there is a difference of $\frac{T-1}{T} b_c$, but as discussed in Appendix~\ref{appendix:bias}, most bias values are close to zero and do not affect the ranking of logits.

\begin{table*}[!t]
\centering
\caption{Overall performance and feature statistics (mean and standard deviation values) across various models and training methods. Red ECE values indicate overconfidence, while blue ECE values indicate underconfidence.
For PGD and AutoAttack, we present results under two hyperparameter settings, with detailed configurations provided in Section~\ref{sec:4_4}.
}
\resizebox{\textwidth}{!}{
\setlength{\tabcolsep}{6pt}
\begin{tabular}{c c c c c c c c c c c}

\toprule
\multirow{2.5}{*}{\bf Model} & \multirow{2.5}{*}{\bf Method} & \multirow{2.5}{*}{\shortstack{\bf Validation \\ \bf Accuracy (\%)}} & \multirow{2.5}{*}{\bf RMS} & \multirow{2.5}{*}{\shortstack{\bf Cosine \\ \bf Similarity}} & \multirow{2.5}{*}{\bf ECE} & \multicolumn{5}{c}{\bf Attack Success Rate (\%)}  \\
\cmidrule(lr){7-11} 
& & & & & & \bf FGSM & \bf PGD$^\text{w}$ & \bf PGD$^\text{s}$ & \bf AutoAttack$^\text{w}$ & \bf AutoAttack$^\text{s}$ \\
\midrule

\multirow{4}{*}{ResNet50}
 & Baseline        & 76.1 & 0.62 \(\pm\) 0.09 & 0.27 \(\pm\) 0.06 & {\color{red}0.074}
   & 67.2 & 84.8 & 99.6 & 87.1 & 99.9
 \\
 & Label smoothing & 77.1 & 0.29 \(\pm\) 0.07 & 0.32 \(\pm\) 0.09 & {\color{blue}0.042}
   & 61.6 & 80.5 & 97.6 & 84.0 & 99.7
 \\
 & Mixup           & 76.6 & 0.30 \(\pm\) 0.04 & 0.27 \(\pm\) 0.05 & {\color{blue}0.202}
   & 55.0 & 76.8 & 97.5 & 83.3 & 99.7
 \\
 & CutMix          & 78.0 & 0.20 \(\pm\) 0.04 & 0.38 \(\pm\) 0.08 & {\color{blue}0.169}
   & 55.2 & 81.8 & 95.3 & 86.9 & 99.8
 \\ \midrule

\multirow{4}{*}{Swin-T}
 & Baseline        & 75.8 & 1.38 \(\pm\) 0.06 & 0.36 \(\pm\) 0.09 & {\color{red}0.095}
   & 83.9 & 85.0 & 99.7 & 85.5 & 99.8
 \\
 & Label smoothing & 76.3 & 0.67 \(\pm\) 0.14 & 0.44 \(\pm\) 0.13 & {\color{blue}0.033}
   & 79.0 & 82.4 & 99.2 & 83.6 & 99.8
 \\
 & Mixup           & 78.2 & 0.66 \(\pm\) 0.13 & 0.46 \(\pm\) 0.11 & {\color{red}0.013}
   & 71.9 & 80.9 & 98.8 & 84.9 & 98.5
 \\
 & CutMix          & 78.7 & 0.72 \(\pm\) 0.13 & 0.51 \(\pm\) 0.13 & {\color{red}0.050}
   & 77.1 & 83.2 & 99.6 & 84.8 & 99.7
 \\ \midrule

\multirow{4}{*}{MobileNetV2}
 & Baseline        & 70.8 & 0.08 \(\pm\) 0.01 & 0.22 \(\pm\) 0.04 & {\color{red}0.081}
   & 83.2 & 92.7 & 99.8 & 93.4 & 99.9
 \\
 & Label smoothing & 71.1 & 0.08 \(\pm\) 0.01 & 0.22 \(\pm\) 0.04 & {\color{red}0.078}
   & 82.3 & 91.9 & 99.8 & 92.8 & 99.9
 \\
 & Mixup           & 70.5 & 0.04 \(\pm\) 0.01 & 0.25 \(\pm\) 0.05 & {\color{blue}0.158}
   & 80.3 & 89.1 & 99.6 & 92.7 & 99.8
 \\
 & CutMix          & 70.9 & 0.03 \(\pm\) 0.01 & 0.25 \(\pm\) 0.06 & {\color{blue}0.188}
   & 80.3 & 91.8 & 99.7 & 93.1 & 99.9
 \\ \midrule

\multirow{3}{*}{ConvNeXt-T}
 & Baseline        & 70.5 & 0.44 \(\pm\) 0.08 & 0.24 \(\pm\) 0.07 & {\color{red}0.211}
   & 82.6 & 91.0 & 99.7 & 91.5 & 99.9
 \\
 & Label smoothing & 73.5 & 0.07 \(\pm\) 0.02 & 0.49 \(\pm\) 0.18 & {\color{blue}0.045}
   & 78.0 & 87.6 & 99.4 & 88.2 & 99.8
 \\
 & Mixup           & 78.0 & 0.12 \(\pm\) 0.03 & 0.28 \(\pm\) 0.08 & {\color{red}0.028}
   & 66.6 & 81.8 & 99.1 & 84.9 & 99.7
 \\

\bottomrule
\end{tabular}
}
\label{tab:3}
\vspace{-1em}
\end{table*}

\subsection{Impact on Adversarial Robustness}
\label{sec:4_4}

\begin{tcolorbox}[
    colback=gray!15,
    colframe=gray!70!black,
    boxrule=1pt,
    arc=5pt,
    left=3pt,
    right=3pt,
    top=2pt,
    bottom=1pt
]
\textbf{Key Takeaway:} Regularization improves robustness to gradient-based adversarial attacks by enhancing feature alignment with class centers.
\end{tcolorbox}

How does the use of regularization lead to better robustness against gradient-based adversarial attacks? To explain this, we examine the gradient directions in the 2D space of ResNet50 on CIFAR-10 in Fig.~\ref{fig:7_a}. 
Our goal is to investigate the distance between features and the decision boundary, as well as the direction in which these features move under adversarial perturbations. To this end, we employ FGSM, which is a single-step attack that provides a clear view of how features respond to adversarial noise.
Note that the gradients shown in Fig.~\ref{fig:7_a} are used by FGSM to perturb the data.
To visualize the gradient and perturbation directions together, we examine two sample features (green and orange) in Fig.~\ref{fig:7_a}. 
The amount of perturbation in FGSM is set to $\epsilon=8/255$.
The feature vector well-aligned with the class center (green) remains within the decision region after perturbation, as the gradient points toward the origin. On the other hand, the feature vector poorly aligned with the class center (orange) moves along the gradient direction toward the nearby decision boundary and becomes easily misclassified. Therefore, when two features, one closely aligned with its class center and the other not, are compared, the latter is more vulnerable to attacks.

We also verify the vulnerability of features to attacks with respect to the degree of alignment in the original representation space for ResNet50 trained on ImageNet. The results are shown in the middle row of Fig.~\ref{fig:5}, where the blue bars represent the histogram of cosine similarity between features and their class centers, and the orange line shows the attack success rate of FGSM attack for each confidence bin.
It is evident that as the cosine similarity between features and class centers increases, indicating better alignment, the robustness improves.

In addition to FGSM, we evaluate model robustness under stronger adversarial attacks, such as Projected Gradient Descent (PGD) \citep{madry2018towards} and AutoAttack \citep{autoattack}.
For each attack, we consider two configurations to assess the effect of perturbation strength: (i) a weaker setting (7 iterations with step size $\alpha = 0.2/255$ and perturbation limit $\epsilon = 0.4/255$ for PGD and $\epsilon = 1/255$ for AutoAttack), and (ii) a stronger setting (7 iterations with $\alpha = 2/255$, $\epsilon = 8/255$ for PGD, and $\epsilon = 8/255$ for AutoAttack, following the standard implementations). 
We denote these configurations as PGD$^\text{w}$, AutoAttack$^\text{w}$ and PGD$^\text{s}$, AutoAttack$^\text{s}$, respectively.
The results are shown in Fig.\ref{fig:5} and Tab.\ref{tab:3}.
As shown in Fig.\ref{fig:5} (middle row), better alignment (\ie, higher cosine similarity) between features and class centers consistently enhances robustness against stronger adversarial attacks as well (see Figs.~\ref{fig:swin_t}-\ref{fig:stronger_settings5} in Appendix~\ref{appendix:more_result_regularization} for results on other models).




Thus, for a model to be robust against adversarial attacks, its features should be well-aligned with their respective class centers. 
\textbf{As a greater proportion of features in regularized models exhibit high cosine similarity to their class centers} (as demonstrated in Section~\ref{sec:4_2}), \textbf{these models are more robust to adversarial attacks}.

\section{Resolving the Contradiction}

Now we are ready to resolve the contradiction mentioned in the introduction. The reduction in feature RMS due to regularization shifts features closer to decision boundaries near the origin, which mitigates overconfident predictions and thereby improving calibration. Although the features become closer to the decision boundaries, the increased cosine similarity between features and class centers enhances robustness against adversarial attacks because adversarial perturbations become directed toward the origin rather than toward decision boundaries at the sides.


\begin{wrapfigure}{r}{0.45\textwidth}
  \vspace{-1em}
  \centering
  \hspace{-10pt}
  \includegraphics[width=0.45\textwidth]{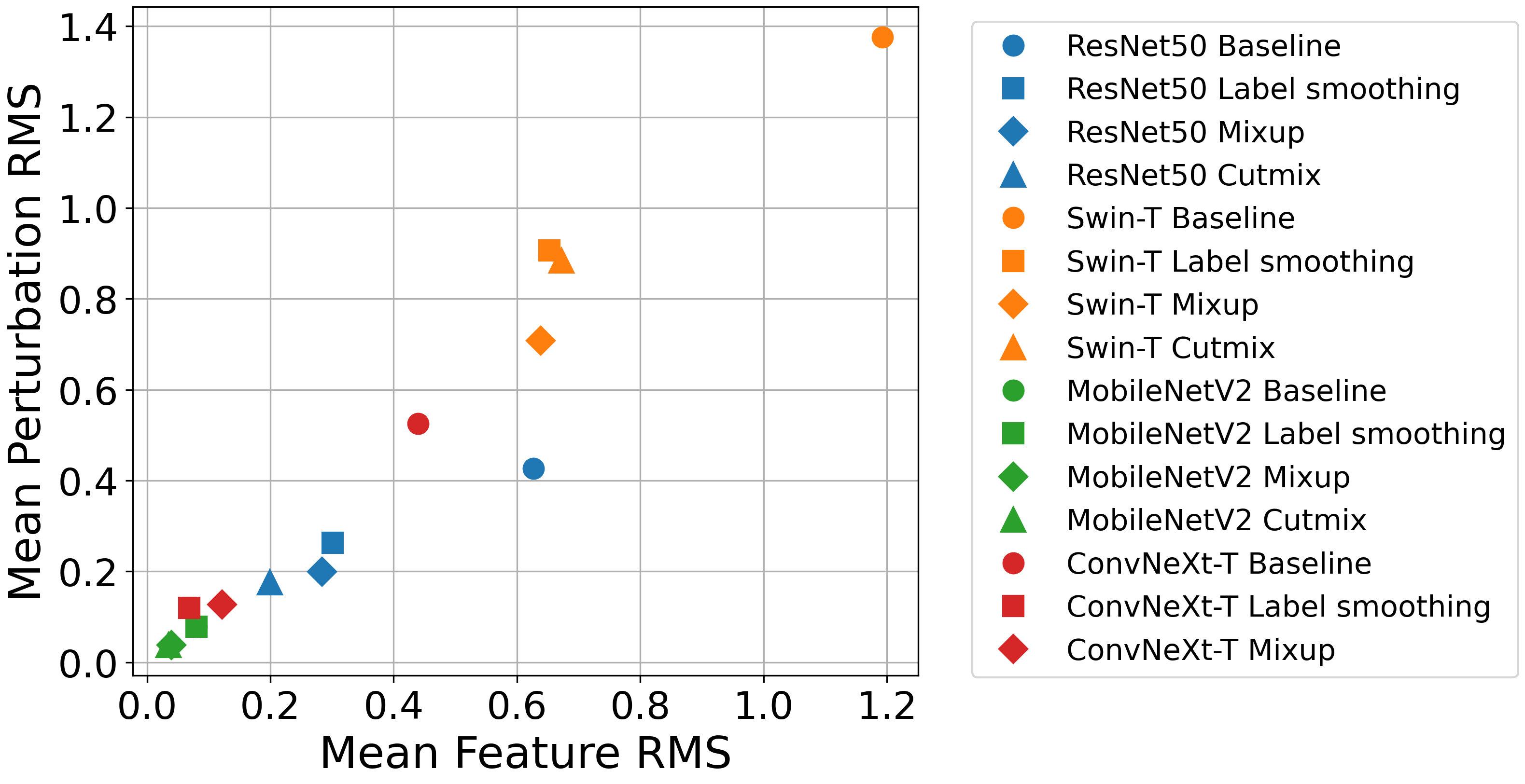}
  \caption{Feature RMS vs. perturbation RMS (the RMS of the difference between the features of clean and perturbed input images) for models trained on ImageNet.
  }
  \label{fig:rms_corr}
  \vspace{-1em}
\end{wrapfigure}
One may argue that because features become closer to the origin, it may be easier to perturb them so that they move across the origin into another decision region.
For example, when the pink features in Figs. \ref{fig:rms_decrease_baseline} and \ref{fig:rms_decrease_mixup} are compared, those in Fig.~\ref{fig:rms_decrease_mixup}, being closer to the origin, could seem more susceptible to attacks, as they could be perturbed across the origin into the gray decision region. However, this is not the case because \textbf{the scale of perturbations in the representation space is not uniform but grows with the scale of the features}.
Fig.~\ref{fig:rms_corr} compares the feature RMS and perturbation RMS for various models trained on ImageNet when $\epsilon = 8/255$. The positive correlation suggests that the features close to the origin move less than the features located far from the origin under the same amount of input perturbation. Therefore, robustness is determined mostly by the direction in which a feature moves due to perturbation, rather than by the proximity of the feature to the decision boundary near the origin.

\section{Comprehensive Evaluation}
\label{sec:4_5}

In Tab.~\ref{tab:3}, we present comprehensive results for various models trained with different methods on the ImageNet dataset (see Appendix~\ref{appendix:implementation} for training details). We consistently observe that, when regularization is applied, the RMS of features decreases, and the cosine similarity between features and class centers increases. These changes result in reduced overconfidence in predictions (leading even to underconfidence in some cases) and improved robustness to adversarial attacks.
However, a limitation to note is that, while regularization consistently improves robustness across a variety of adversarial attacks, its effect remains limited under the strongest AutoAttack configuration. 
Achieving further robustness against such strong attacks likely requires complementary strategies (\eg, adversarial training).


\section{Conclusion}
In this paper, we investigated how regularization techniques using soft labels enhance model calibration and robustness against gradient-based adversarial attacks. We analyzed decision regions, confidence contours, and gradient directions in the representation space, demonstrating that regularization reduces feature RMS and increases cosine similarity to class centers. The reduced RMS mitigates overconfident predictions, while higher cosine similarity directs perturbations toward the origin, enhancing adversarial robustness. Our findings offer new insights into the regularization dynamics in the representation space.

\bibliography{iclr2026_conference}
\bibliographystyle{iclr2026_conference}

\appendix



\appendix
\renewcommand{\thesubsection}{\thesection.\arabic{subsection}}
\setcounter{secnumdepth}{2} 

\newpage

\section{Discussion}
\label{sec:5}

Numerous studies suggest that transformer-based models outperform convolution-based models in calibration \citep{minderer2021revisiting} and adversarial robustness \citep{bai2021transformers, Benz2021AdversarialRC, paul2022vision}. However, these comparisons often overlook the use of soft label-based regularization during the pre-training and training phase of transformer-based models. 
To address this, we compare ResNet50 (25.5M parameters) and Swin-T (28.2M) using identical training conditions to achieve comparable validation accuracy. Unlike prior studies, in Tab.~\ref{tab:3}, we find that Swin-T proves to be as overconfident as ResNet50 without regularization. In addition, Swin-T is actually more susceptible to adversarial attacks, even with higher cosine similarity between features and class centers. However, because the higher cosine similarity may be due to the lower feature dimensionality of Swin-T (768) compared to ResNet50 (2048), direct comparison with respect to cosine similarity may be limited.

\section{Decision Regions of Original Models}
\label{appendix:original_boundaries}

In this section, we investigate whether, when the dimensionality of the representation space is sufficiently high (\eg, 2048 for ResNet50), the decision regions have cone shapes regardless of the presence of bias terms. One simple way to verify this is to gradually move a correctly classified feature linearly toward the origin and observe when it becomes misclassified for the first time. This process is illustrated in Fig.~\ref{fig:3}. If the decision regions are cone-shaped, the classification result will remain consistent until the feature arrives at the origin. Actually, the intersection point of the cone-shaped decision regions does not precisely coincide with the origin, but is close to the origin. Thus, the moment that the misclassification occurs will be only at the final stage of the linear movement. On the other hand, if the regions are not cone-shaped, meaning another class region lies between the feature and the origin, the feature will become misclassified early during the movement.

\begin{minipage}{\columnwidth}
  \centering
  \begin{minipage}{0.4\columnwidth} 
    \begin{figure}[H]
      \centering
      \includegraphics[width=\linewidth]{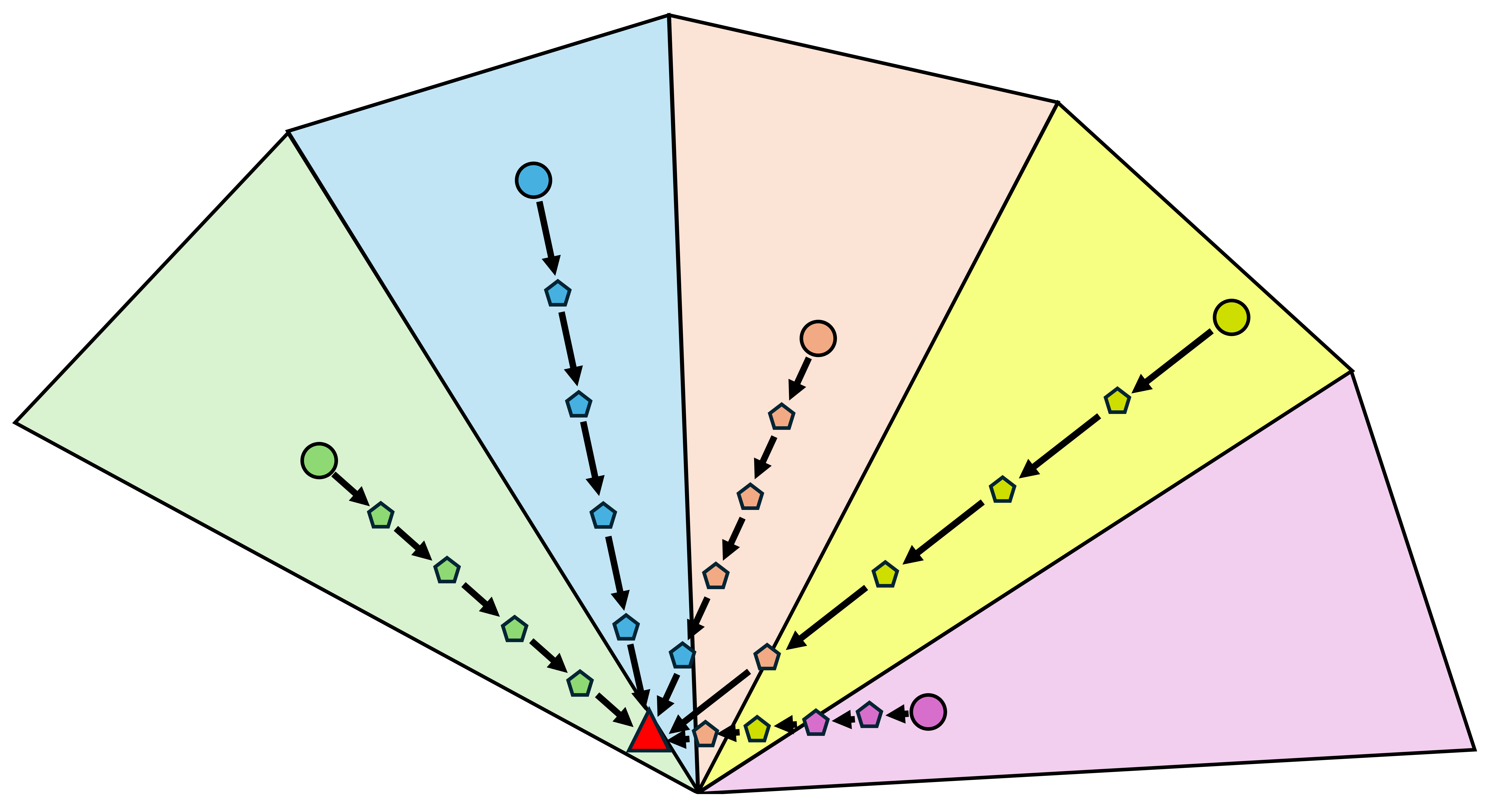}
      \caption{
        Illustration of linear movements of features (circled dots) toward the origin (red triangle). Each large triangular region represents the decision region of a specific class. Pentagons represent the intermediate positions of features as they move toward the origin. The colors within each dot indicate the class to which they are classified.
      }
      \label{fig:3}
    \end{figure}
  \end{minipage}
  \hspace{0.03\linewidth}
  \begin{minipage}{0.55\columnwidth} 
    \begin{table}[H]
      \caption{
        Accuracy and the mean ($\pm$ standard deviation) of the first movement index of misclassification for various models on different datasets.
      }
      \resizebox{\linewidth}{!}{ 
        \renewcommand{\arraystretch}{1.2}
        \begin{tabular}{c c c c}
          \toprule
          \bf Model & \bf Dataset & \bf Accuracy (\%) & \bf Index \\ \hline

          \multirow{3}{*}{ResNet50}
          & CIFAR-10 & 92.5 & 99.9 \textpm 1.6\\
          & CIFAR-100 & 71.4 & 99.4 \textpm 3.9\\
          & ImageNet & 76.1 & 99.8 \textpm 1.9\\ \hline
    
          \multirow{3}{*}{Swin-T}
          & CIFAR-10 & 89.3 & 99.9 \textpm 1.3\\
          & CIFAR-100 & 66.7 & 99.6 \textpm 2.7\\
          & ImageNet & 75.8 & 99.3 \textpm 3.2 \\ \hline
    
          \multirow{3}{*}{MobileNetV2}
          & CIFAR-10 & 92.6 & 99.8 \textpm 1.7 \\
          & CIFAR-100 & 71.7 & 98.9 \textpm 4.2 \\
          & ImageNet & 70.8 & 91.3 \textpm 10.7 \\ \hline

          \multirow{3}{*}{ConvNeXt-T}
          & CIFAR-10 & 94.5 & 99.7 \textpm 2.5 \\
          & CIFAR-100 & 70.6 & 98.1 \textpm 5.2 \\
          & ImageNet & 70.5 & 93.9 \textpm 8.1 \\

          \bottomrule
        \end{tabular}
      }
      \label{tab:1}
    \end{table}
  \end{minipage}
\end{minipage}

We verify this for ResNet50, Swin-T, MobileNetV2 \citep{howard2017mobilenets}, and ConvNeXt-T \citep{liu2022convnet} on the test sets of CIFAR-10 and CIFAR-100, and the validation set of ImageNet \citep{russakovsky2015imagenet}. For each feature in the representation space, we linearly move it toward the origin over 100 uniform steps. If the index of the first misclassified step is close to 100, it suggests that the decision region is likely cone-shaped.
The results are shown in Tab.~\ref{tab:1}. Since all indices are over 91 on average, we can confirm that decision regions are divided into cone shapes even with the presence of bias terms, if the dimensionality of the representation space is high enough. Further verification can be found in Appendix~\ref{appendix:bias}.

\section{Bias Term in the Classification Layer}
\label{appendix:bias}

In this section, we provide a more detailed discussion on the effects of the bias term on decision regions, as introduced in Section~\ref{sec:3.1}, both in the 2D and original representation spaces.

\begin{figure*}[t]
    \centering
    \subfloat[]
    {
        {\includegraphics[width=0.25\textwidth]{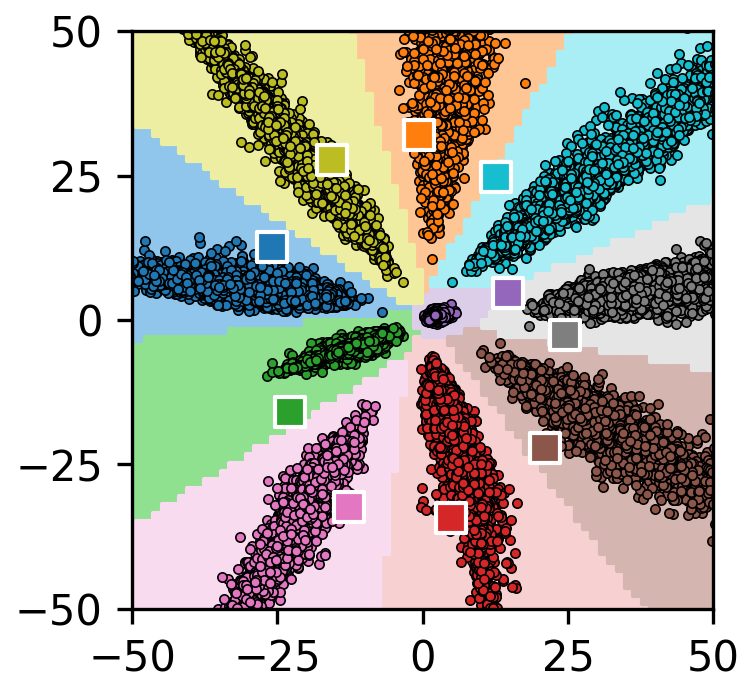}}
        \label{fig:appendix_bias_2d_a}
    }
    \subfloat[]
    {
        {\includegraphics[width=0.23\textwidth]{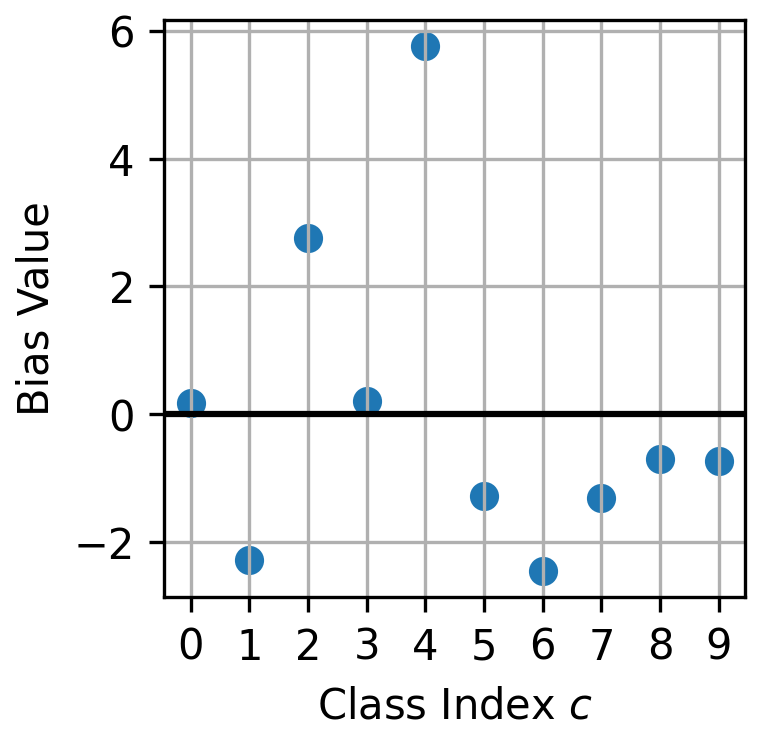}}
        \label{fig:appendix_bias_2d_b}
    }\hspace{-0.0cm}
    \subfloat[]
    {
        {\includegraphics[width=0.23\textwidth]{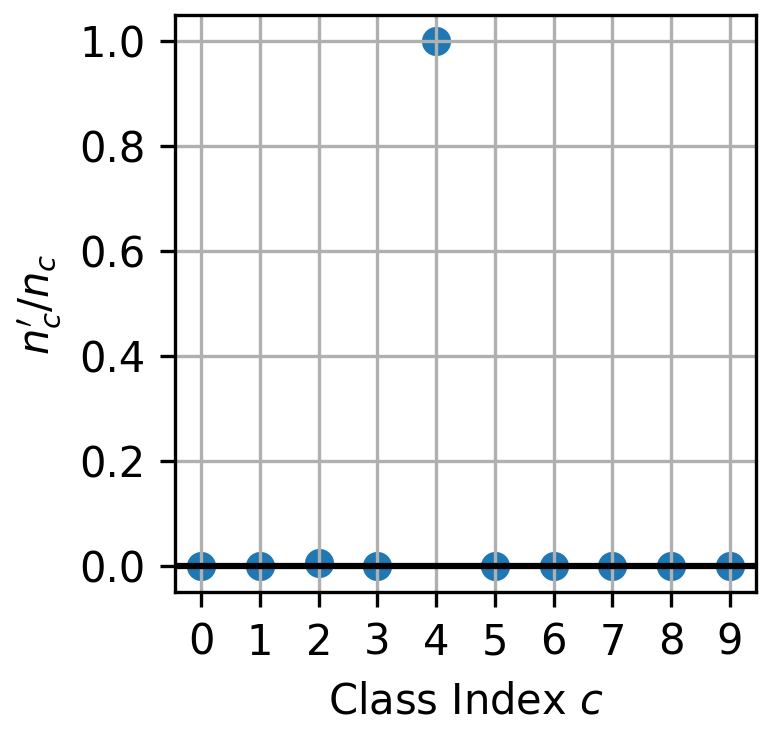}}
        \label{fig:appendix_bias_2d_c}
    }
    \caption{
        Results for ResNet50 with a 2D representation space trained on CIFAR-10.
        (a) 2D representation space. Circled and squared dots represent the features and weight vectors, respectively. Different colors indicate different class regions and classification results.
        (b) Bias values for each class.
        (c) $n_c'/n_c$ values for each class.
    }
    \label{fig:appendix_bias_2d}
\end{figure*}

\subsection{2D Representation Spaces}

Fig.~\ref{fig:appendix_bias_2d_a} shows the 2D representation space of ResNet50 with the classification layer bias (identical to Fig.~\ref{fig:with_bias}), trained on CIFAR-10.
While most classes form cone-shaped decision regions, a purple class with a circular decision region appears near the origin. As discussed in Section~\ref{sec:3.1}, cone-shaped decision regions arise because the classification result is determined by the weight vector that is aligned most closely with the feature vector. However, when two weight vectors have similar directions, as the purple and gray classes in Fig.~\ref{fig:appendix_bias_2d_a}, the final classification is determined by the biases.
To validate this, we examine the bias values for all 10 classes in Fig.~\ref{fig:appendix_bias_2d_b}. It is clear that the purple class (class 4) has a significantly higher bias value compared to the other classes. This suggests that without the bias, features from class 4 would not be correctly classified.

We calculate the ratio of instances where prediction results depend on the bias values when determining logits. To elaborate, for an arbitrary class $c$, we count the number of correctly classified samples $n_c = \sum_{i = 1}^{N} \mathbf{1} \left( \hat{y}_i = y_i = c \right)$,
where $\hat{y}_i$ is the predicted class, $y_i$ is the true class for sample $i$, and $N$ is the total number of samples.
Then, let $\mathbb{A}_c$ be the set of indices of samples that are correctly predicted into class $c$. Among such samples, we count the number of samples $n_c'$ where the prediction results would change if the logits are calculated without biases. This can be expressed as $n_c' = \sum_{j \in \mathbb{A}_c} \mathbf{1} \left( \hat{y}_j^{\mathrm{no~bias}} \neq \hat{y}_j \right)$, where $\hat{y}_j^{\mathrm{no~bias}}$ is the predicted class for sample $j$ when logits are calculated without biases.
Therefore, if the ratio $n_c' / n_c$ is large, the presence of bias values are crucial for the data correctly classified as class $c$.

Fig.~\ref{fig:appendix_bias_2d_c} shows the value of $n_c' / n_c$ for each class. The classes with cone-shaped decision regions have low $n_c' / n_c$ values, indicating that the bias terms are not necessary for correctly classifying these classes. However, for class 4 (the purple class in Fig.~\ref{fig:appendix_bias_2d_a}), the value of $n_c' / n_c$ is $1$, suggesting that non-cone-shaped decision regions rely on the bias for accurate classification. Therefore, by examining $n_c' / n_c$, we can determine whether the bias is needed for correct classification of a particular class and infer the shape of its decision region in the representation space.

More examples on the effect of the bias in the 2D representation space are provided in Appendix~\ref{appendix:more_example_2d}.

\subsection{Original Representation Spaces}

Now we verify that the bias-dependence phenomenon rarely occurs in the original high-dimensional representation spaces. Tab.~\ref{tab:appendix_bias_check} shows the mean and standard deviation of $|b_c|$ (the absolute bias values for each class $c$) and the ratios $n_c' / n_c$ for the models trained on ImageNet. The small $|b_c|$ values indicate minimal dependency of the classification results on the bias, leading to the small $n_c' / n_c$ values. Consequently, cone-shaped decision regions are formed for all classes.

\begin{table*}[h]
\centering
\caption{Mean, standard deviation, and maximum values of $|b_c|$ and $n_c'/n_c$ for models trained on ImageNet.}
\resizebox{0.7\linewidth}{!}{
    \begin{tabular}{c c c c}
    
    \toprule

    \textbf{Model} & \textbf{Method} & \textbf{$|b_c|$ Mean (\textpm std)} & \textbf{$n_c'/n_c$ Mean (\textpm std)}
    \\ \hline
    

     \multirow{4}{*}{ResNet50}
     & Baseline & 0.009 (\textpm 0.007) & 0.0005 (\textpm 0.005) \\
     & Label smoothing & 0.011 (\textpm 0.008) & 0.0007 (\textpm 0.005) \\
     & Mixup & 0.008 (\textpm 0.006) & 0.0005 (\textpm 0.005) \\
     & CutMix & 0.009 (\textpm 0.007) & 0.0003 (\textpm 0.003) \\ \hline
    
     \multirow{4}{*}{Swin-T}
      & Baseline & 0.029 (\textpm 0.022) & 0.0010 (\textpm 0.006)  \\
     & Label smoothing & 0.015 (\textpm 0.010) & 0.0009 (\textpm 0.006)  \\
     & Mixup & 0.026 (\textpm 0.020) & 0.0015 (\textpm 0.008)  \\
     & CutMix & 0.026 (\textpm 0.021) & 0.0013 (\textpm 0.007)  \\ 
     \hline

     \multirow{4}{*}{MobileNetV2}
      & Baseline & 0.339 (\textpm 0.254) & 0.0208 (\textpm 0.036) \\
     & Label smoothing & 0.331 (\textpm 0.251) & 0.0219 (\textpm 0.041)  \\
     & Mixup & 0.264 (\textpm 0.184) & 0.0246 (\textpm 0.047)  \\
     & CutMix & 0.262 (\textpm 0.190) & 0.0245 (\textpm 0.045) \\
     \hline

     \multirow{3}{*}{ConvNeXt-T}
      & Baseline & 0.463 (\textpm 0.349) & 0.0097 (\textpm 0.026) \\
     & Label smoothing & 0.212 (\textpm 0.084) & 0.0022 (\textpm 0.009)  \\
     & Mixup & 0.362 (\textpm 0.279) & 0.0165 (\textpm 0.036) \\
     \hline
    
      \multirow{2}{*}{MobileNetV2}
      & PyTorch V1 & 0.028 (\textpm 0.022) & 0.0037 (\textpm 0.014)  \\
     & PyTorch V2 & 0.053 (\textpm 0.042) & 0.0084 (\textpm 0.023) \\ \hline
    
      \multirow{2}{*}{EfficientNet-B1}
      & PyTorch V1 & 0.054 (\textpm 0.041) & 0.0037 (\textpm 0.015) \\
     & PyTorch V2 & 0.116 (\textpm 0.084) & 0.0065 (\textpm 0.021)  \\
     \hline
    
      \multirow{2}{*}{ViT-B/16}
     & PyTorch Swag Linear V1 & 0.030 (\textpm 0.026) & 0.0022 (\textpm 0.011)  \\
      & PyTorch V1 & 0.016 (\textpm 0.013) & 0.0007 (\textpm 0.005) \\

    \bottomrule
     
    \end{tabular}
}
\label{tab:appendix_bias_check}
\end{table*}

\newpage

\section{More Visualization of 2D and 3D Representation Spaces}

\label{appendix:more_example_2d}

\begin{figure}[t]
    \centering
    \subfloat[ResNet50]
    {
        {\includegraphics[width=0.21\columnwidth]{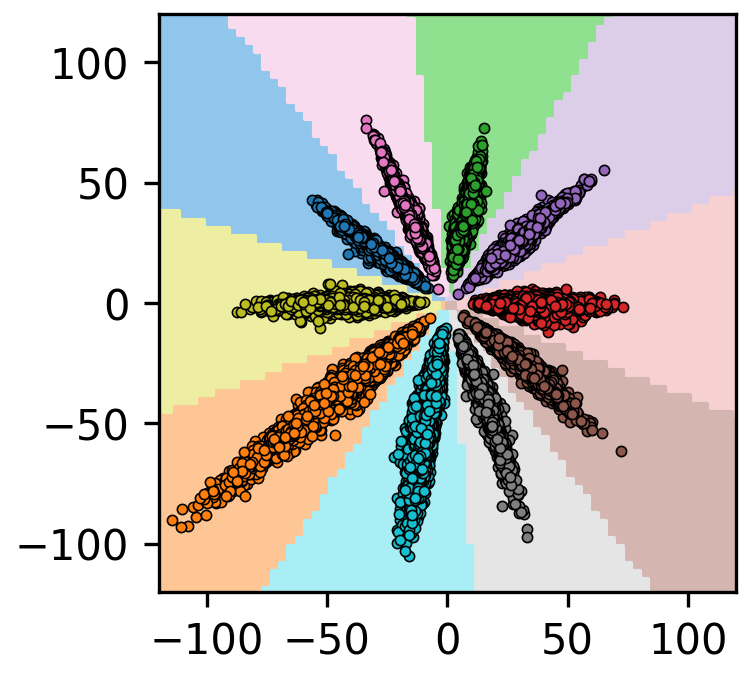}}
        {\includegraphics[width=0.24\columnwidth]{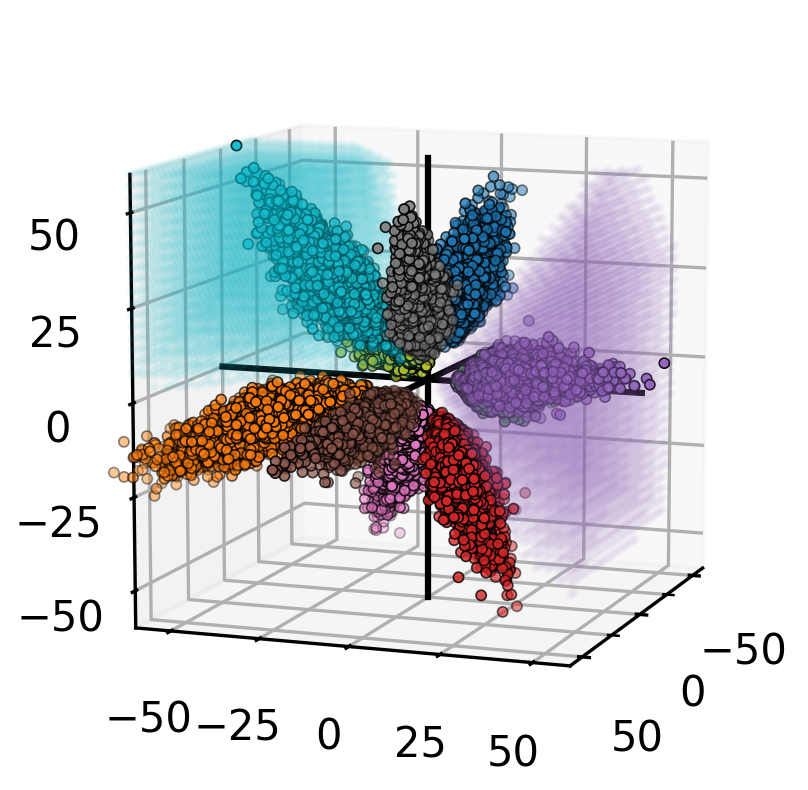}}
        \label{fig:1_a}
    }
    \subfloat[Swin-T]
    {
        {\includegraphics[width=0.21\columnwidth]{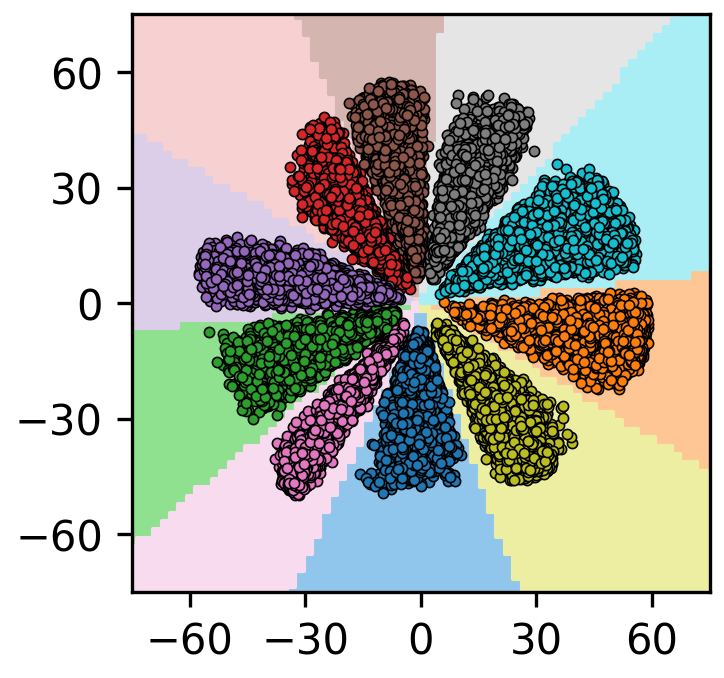}}
        {\includegraphics[width=0.24\columnwidth]{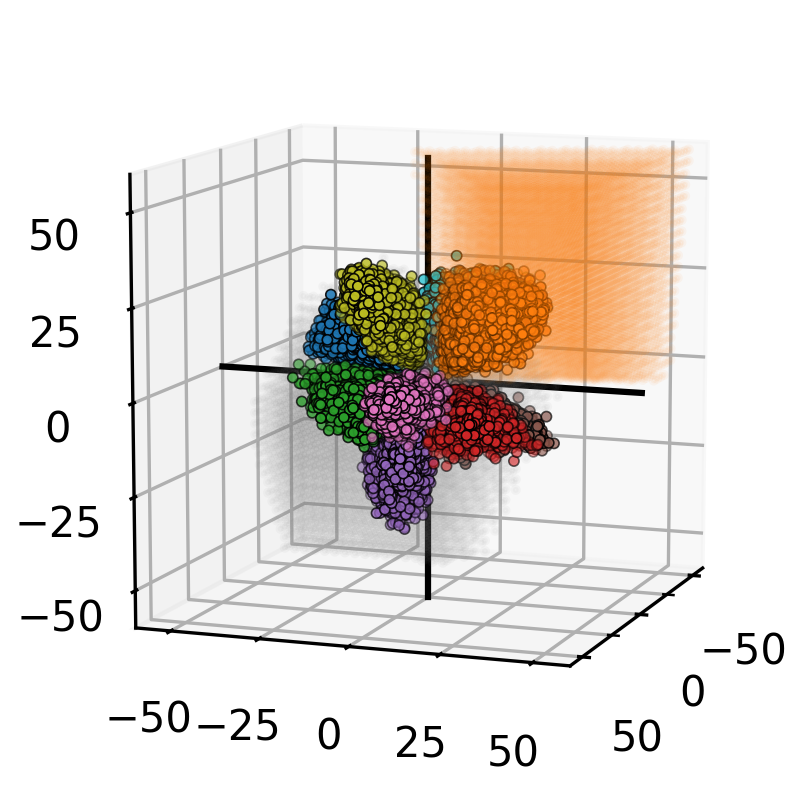}}
        \label{fig:1_c}
    }


    \caption{
        Visualization of the 2D and 3D representation spaces of ResNet50 and Swin-T on CIFAR-10. Circled dots represent output features, with different colors indicating different classes. The whole 2D planes are also colored according to the classification result of each point in the plane. In the case of 3D, regions corresponding only two classes are colored as examples for the sake of visualization. The values of the feature vectors are used as coordinates for the x, y, and z axes. Note that the scales differ across figures to best visualize the representation spaces.
    }
    \label{fig:1}
\end{figure}

Here, we present additional examples of decision regions divided around the origin, extending the discussion in Section~\ref{sec:3.1}.
Fig.~\ref{fig:1} shows the visualization results for ResNet50 and Swin-T on the CIFAR-10 dataset, when the models are trained without the bias in the classification layer.
Regardless of the differences in model structures, it can be observed that the decision regions are divided into circular sectors, \ie, \textit{cone-like shapes}, centered around the origin, with features radially distributed within these regions. Note that for the results shown in Fig.~\ref{fig:1}, regularization using soft labels is not applied.

\begin{figure*}[h]
    \centering

    \subfloat[{\begin{tabular}{c}ResNet50 \\ Baseline \\ w/o bias\end{tabular}}]
    {
        {\includegraphics[width=0.238\columnwidth]{appendix_c/resnet50_baseline_bias0_2d.png}}
    }
    \subfloat[{\begin{tabular}{c}ResNet50 \\ Label smoothing \\ w/o bias\end{tabular}}]
    {
        {\includegraphics[width=0.225\columnwidth]{appendix_c/resnet50_labelsmoothing_bias0_2d.png}}
    }
    \subfloat[{\begin{tabular}{c}ResNet50 \\ Mixup \\ w/o bias\end{tabular}}]
    {
        {\includegraphics[width=0.225\columnwidth]{appendix_c/resnet50_mixup_bias0_2d.png}}
    }
    \subfloat[{\begin{tabular}{c}ResNet50 \\ CutMix \\ w/o bias\end{tabular}}]
    {
        {\includegraphics[width=0.225\columnwidth]{appendix_c/resnet50_cutmix_bias0_2d.png}}
    }

    \subfloat[{\begin{tabular}{c}ResNet50 \\ Baseline \\ w/ bias\end{tabular}}]
    {
        {\includegraphics[width=0.238\columnwidth]{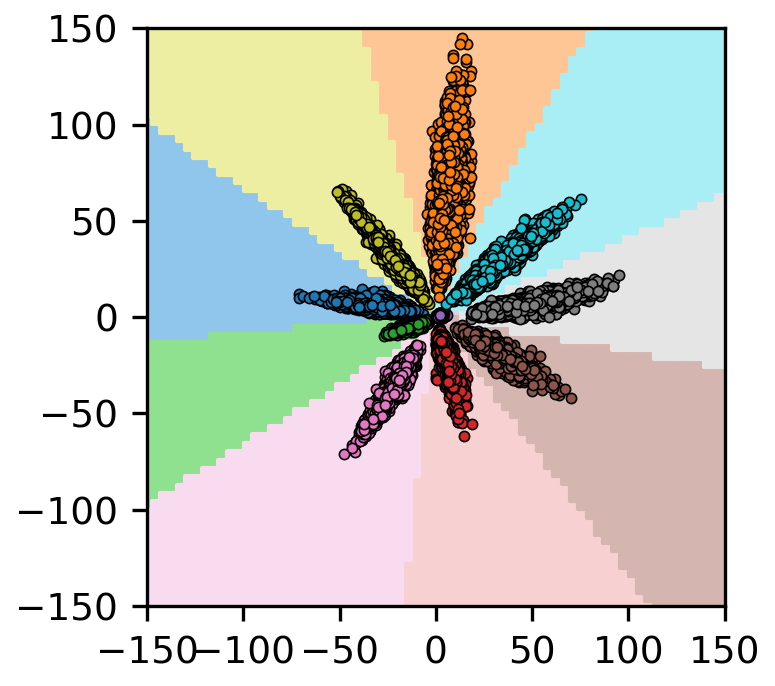}}
    }
    \subfloat[{\begin{tabular}{c}ResNet50 \\ Label smoothing \\ w/ bias\end{tabular}}]
    {
        {\includegraphics[width=0.225\columnwidth]{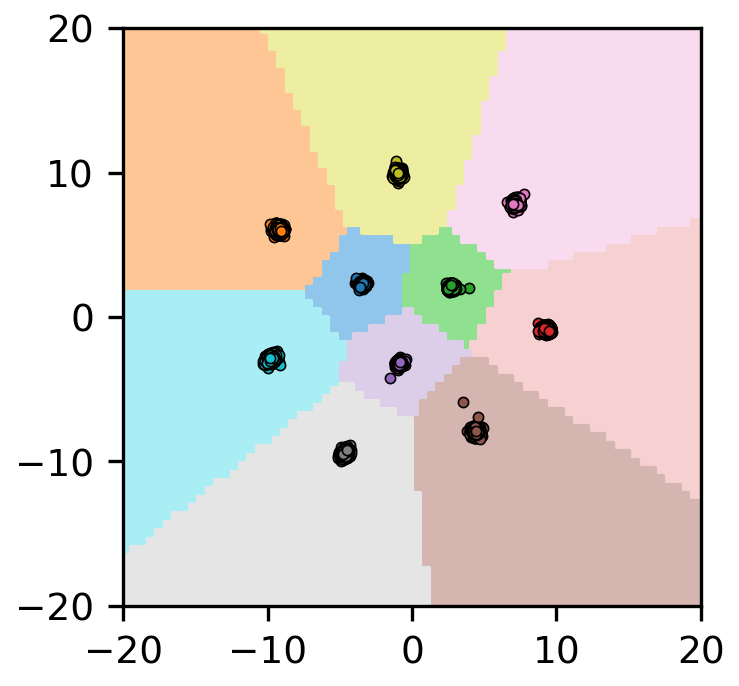}}
    }
    \subfloat[{\begin{tabular}{c}ResNet50 \\ Mixup \\ w/ bias\end{tabular}}]
    {
        {\includegraphics[width=0.225\columnwidth]{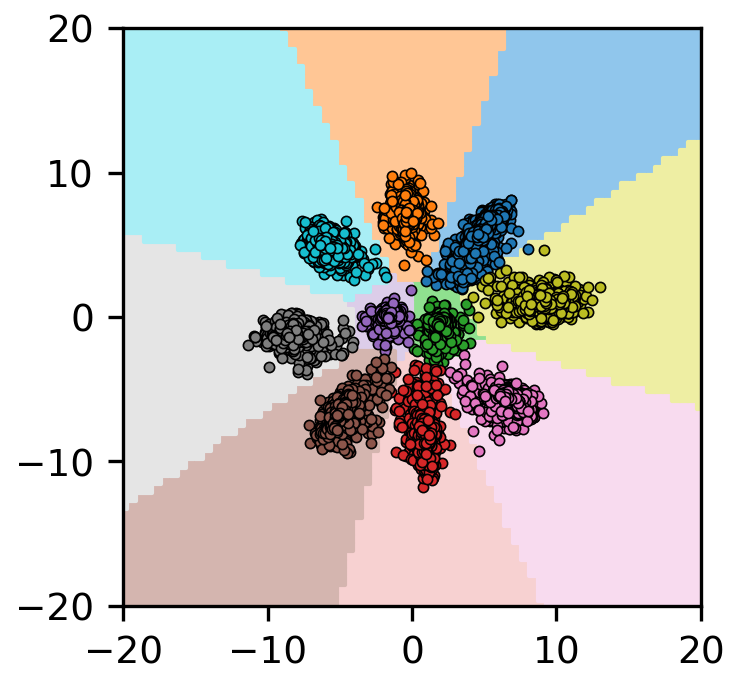}}
    }
    \subfloat[{\begin{tabular}{c}ResNet50 \\ CutMix \\ w/ bias\end{tabular}}]
    {
        {\includegraphics[width=0.225\columnwidth]{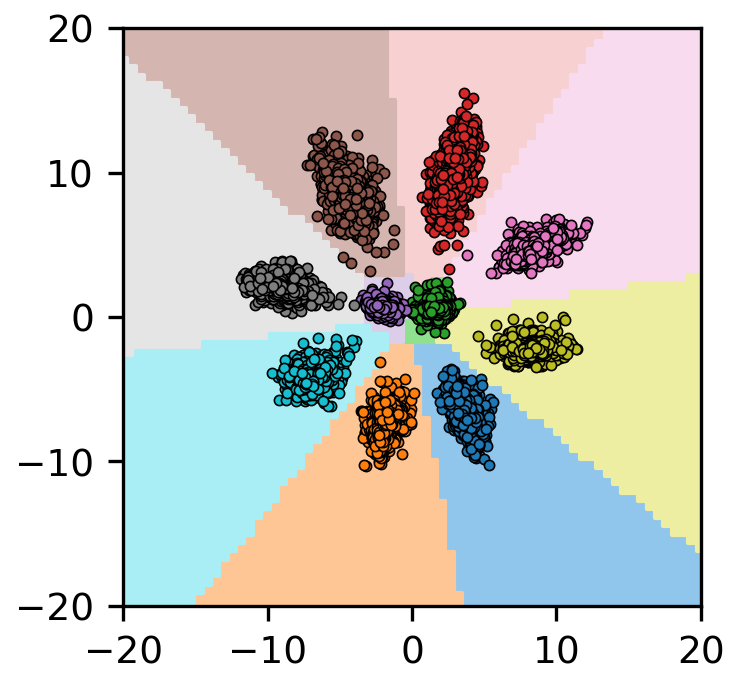}}
    }

    \subfloat[{\begin{tabular}{c}Swin-T \\ Baseline \\ w/o bias\end{tabular}}]
    {
        {\includegraphics[width=0.23\columnwidth]{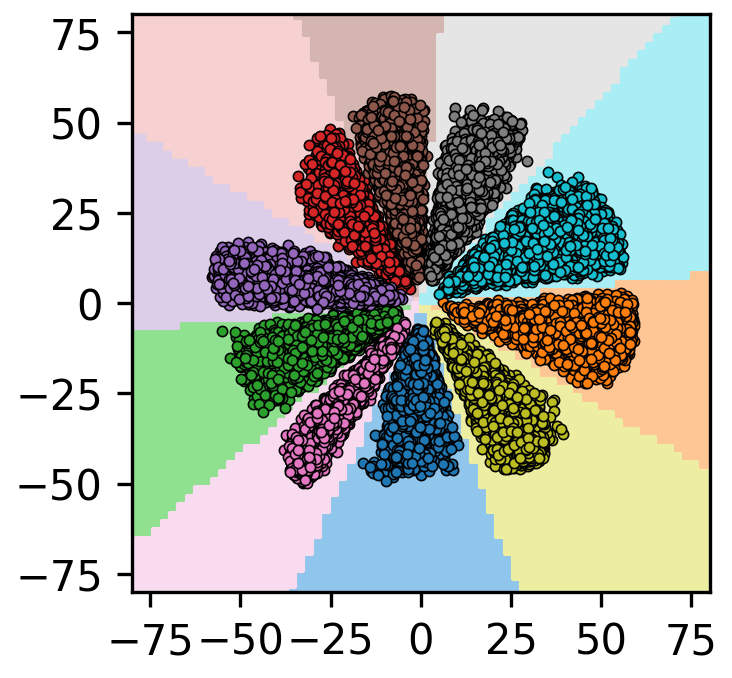}}
    }
    \subfloat[{\begin{tabular}{c}Swin-T \\ Label smoothing \\ w/o bias\end{tabular}}]
    {
        {\includegraphics[width=0.232\columnwidth]{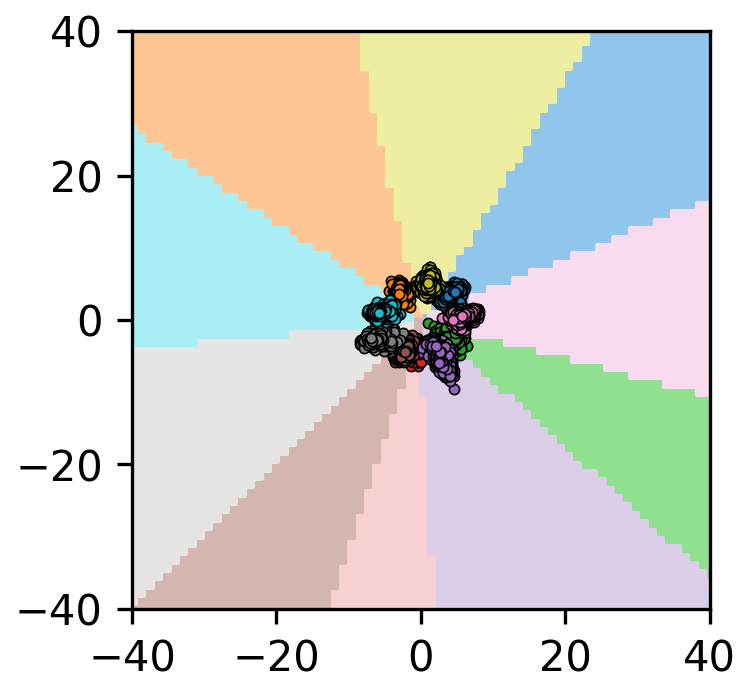}}
    }
    \subfloat[{\begin{tabular}{c}Swin-T \\ Mixup \\ w/o bias\end{tabular}}]
    {
        {\includegraphics[width=0.232\columnwidth]{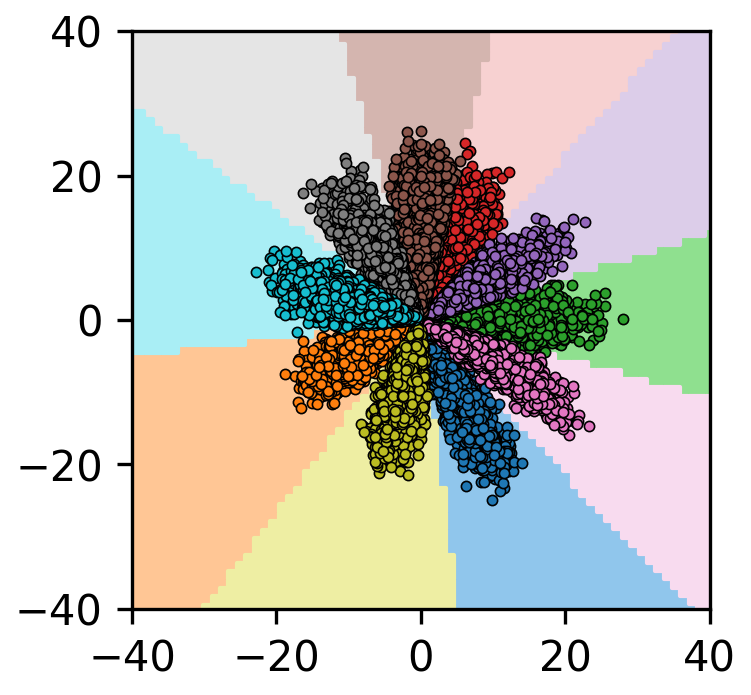}}
    }
    \subfloat[{\begin{tabular}{c}Swin-T \\ CutMix \\ w/o bias\end{tabular}}]
    {
        {\includegraphics[width=0.232\columnwidth]{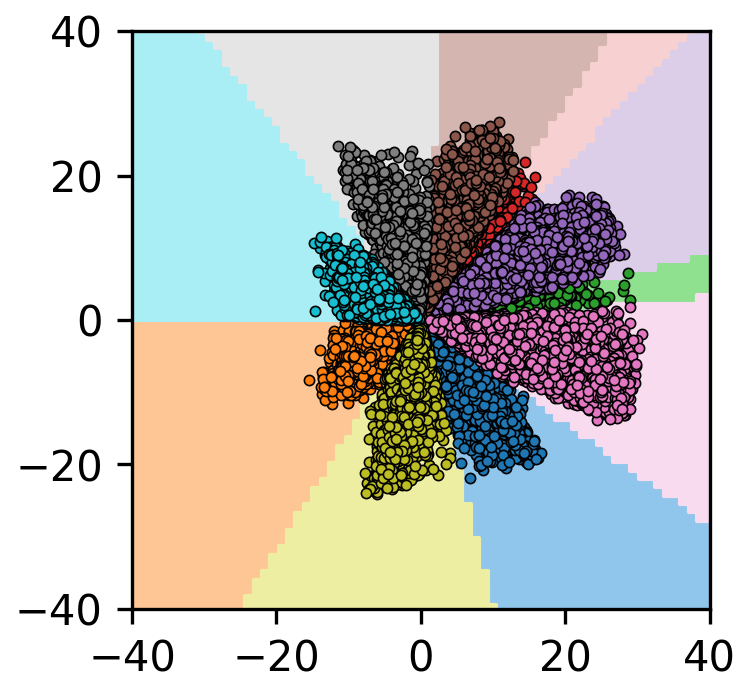}}
    }

    \subfloat[{\begin{tabular}{c}Swin-T \\ Baseline \\ w/ bias\end{tabular}}]
    {
        {\includegraphics[width=0.23\columnwidth]{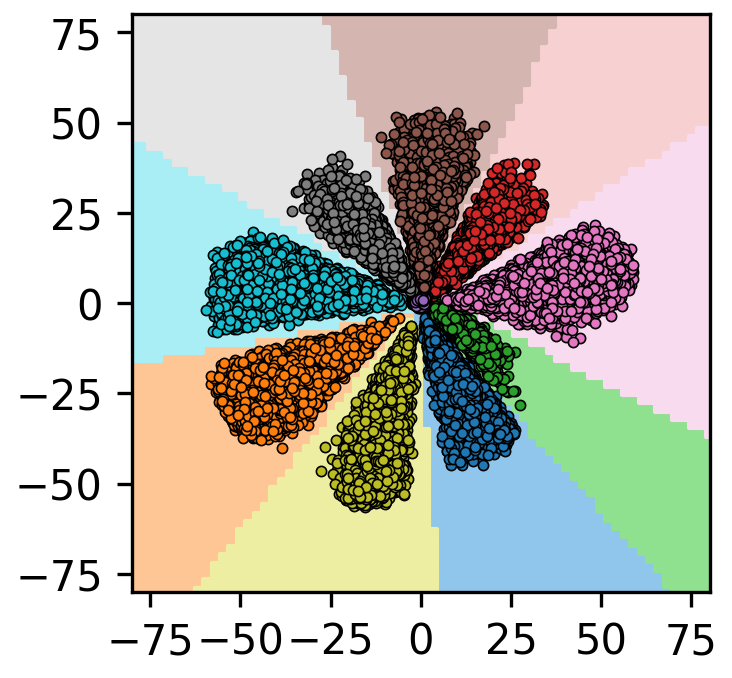}}
    }
    \subfloat[{\begin{tabular}{c}Swin-T \\ Label smoothing \\ w/ bias\end{tabular}}]
    {
        {\includegraphics[width=0.232\columnwidth]{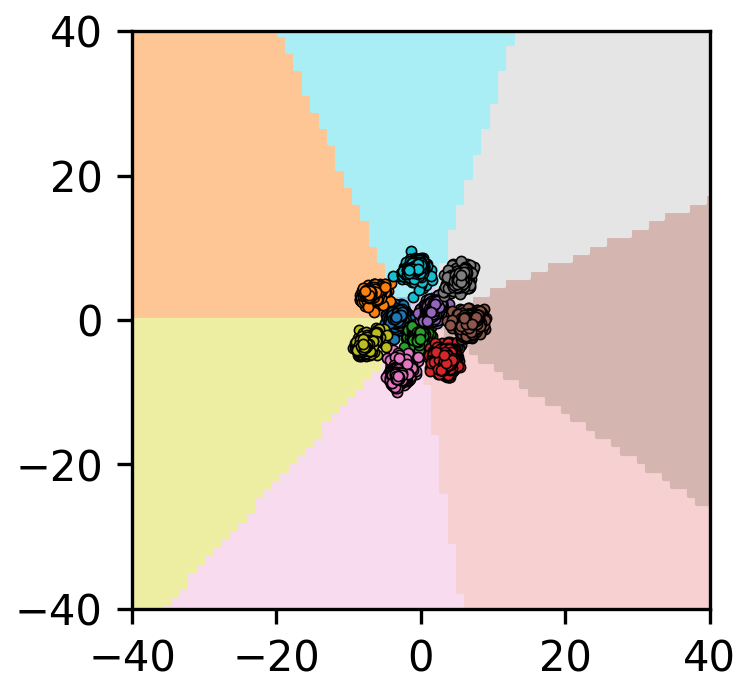}}
    }
    \subfloat[{\begin{tabular}{c}Swin-T \\ Mixup \\ w/ bias\end{tabular}}]
    {
        {\includegraphics[width=0.232\columnwidth]{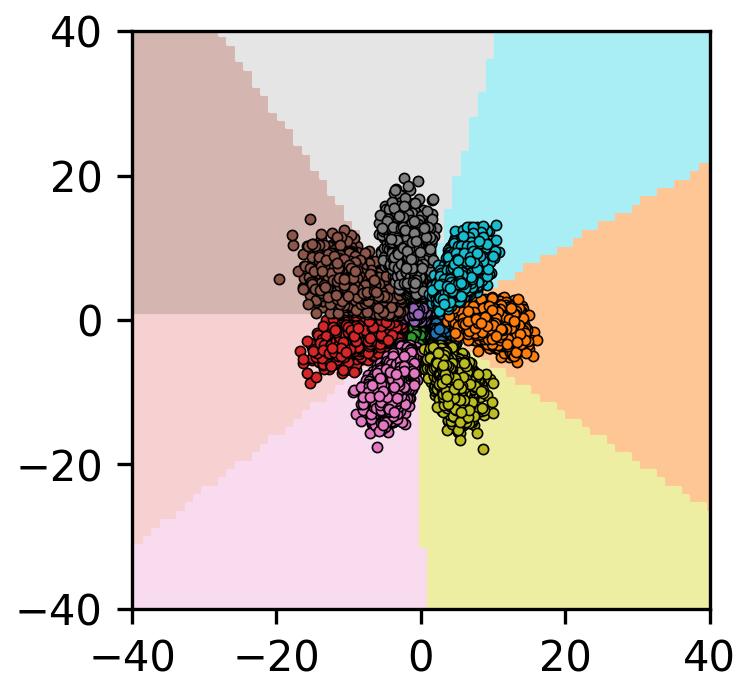}}
    }
    \subfloat[{\begin{tabular}{c}Swin-T \\ CutMix \\ w/ bias\end{tabular}}]
    {
        {\includegraphics[width=0.232\columnwidth]{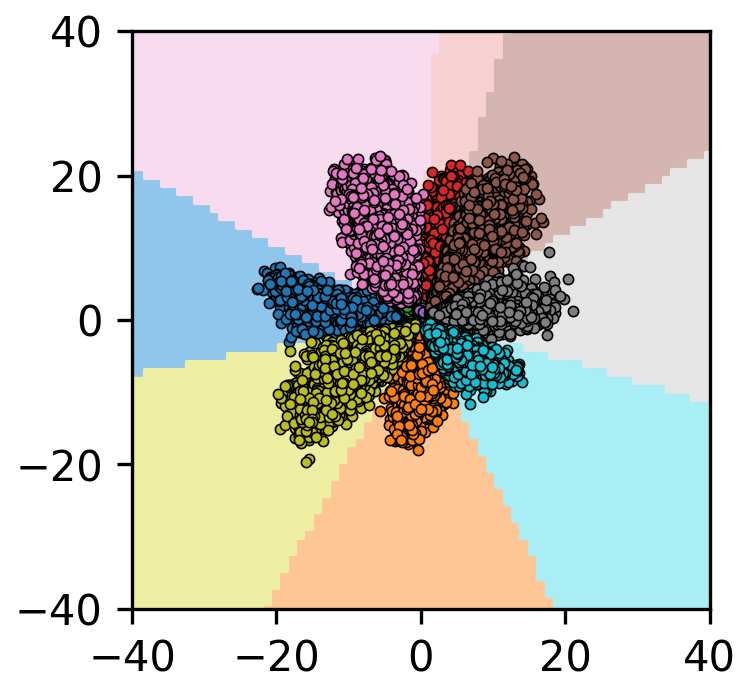}}
    }

    \caption{
        Decision regions and feature distribution in the 2D representation space for ResNet50 and Swin-T on CIFAR-10. Note that the scales differ across figures.
    }
    \label{fig:10}
\end{figure*}

Fig.~\ref{fig:10} provides further examples on how regularization and classification layer bias influence the decision regions and feature distributions in the 2D representation space. As discussed in Section~\ref{sec:3.1} and Appendix~\ref{appendix:bias}, when the bias term is present in the visualizable low-dimensional representation space, some decision regions deviate from cone shapes, as the bias value may have a greater effect than the angular differences between features. Regarding the effect of regularization, as explained in Section~\ref{sec:4_2}, the use of soft labels leads to a significant reduction in the magnitude of the features.

\newpage

\section{Effect of Weight Initialization}
\label{appendix:initialization}

\begin{figure*}[h]
    \centering
    \subfloat[]
    {
        {\includegraphics[width=0.25\columnwidth]{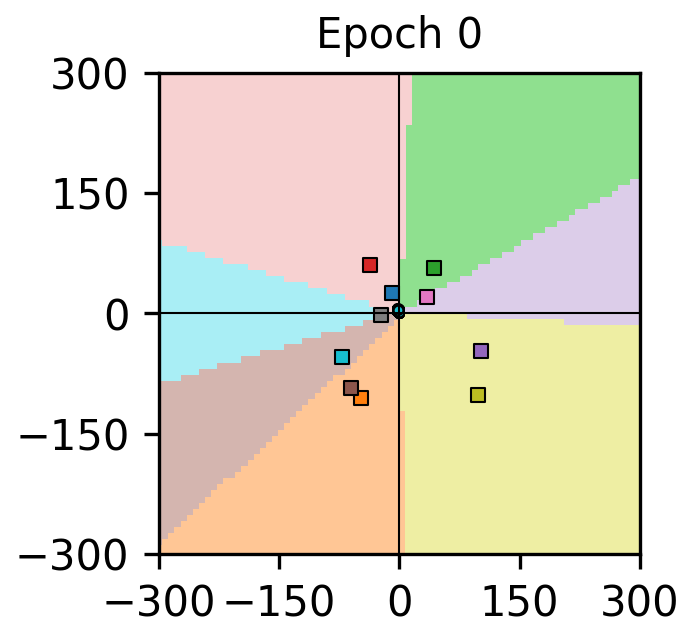}}
        {\includegraphics[width=0.25\columnwidth]{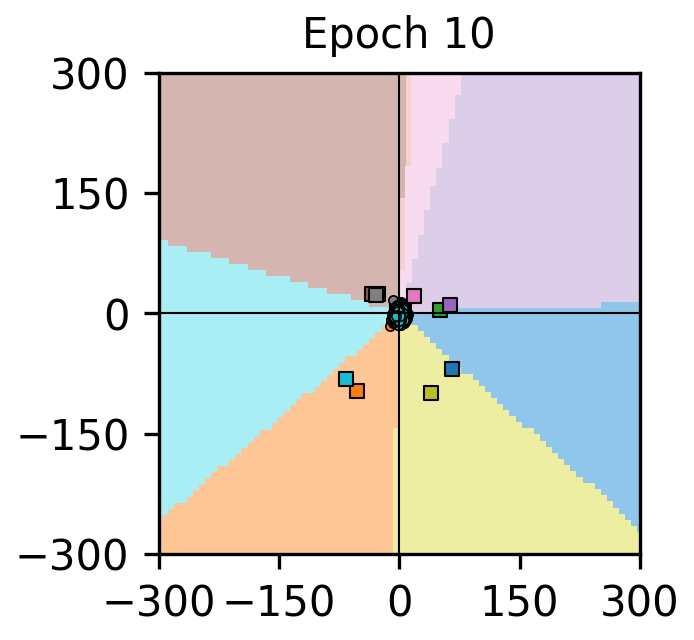}}
        {\includegraphics[width=0.25\columnwidth]{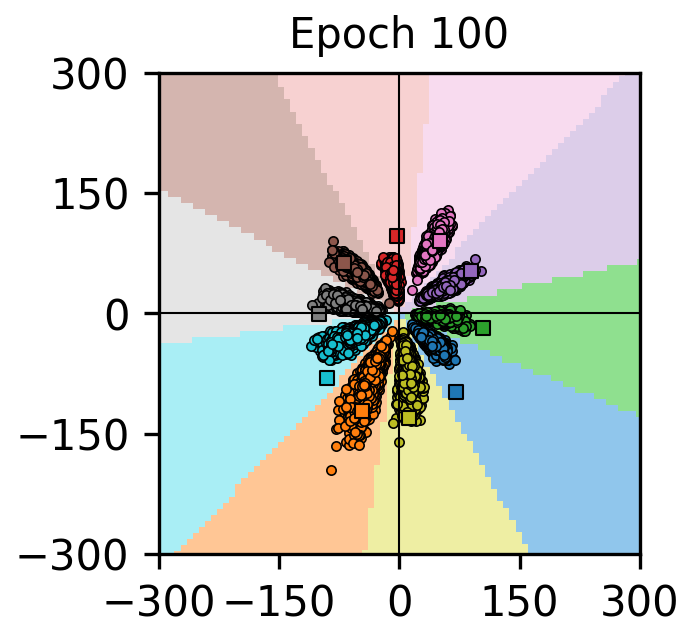}}
        \label{fig:2_a}
    }
    \\
    \subfloat[]
    {
        {\includegraphics[width=0.25\columnwidth]{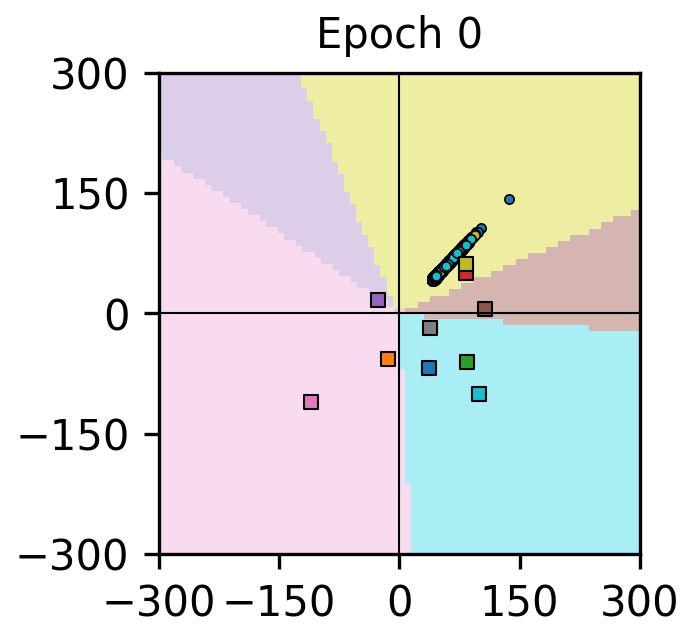}}
        {\includegraphics[width=0.25\columnwidth]{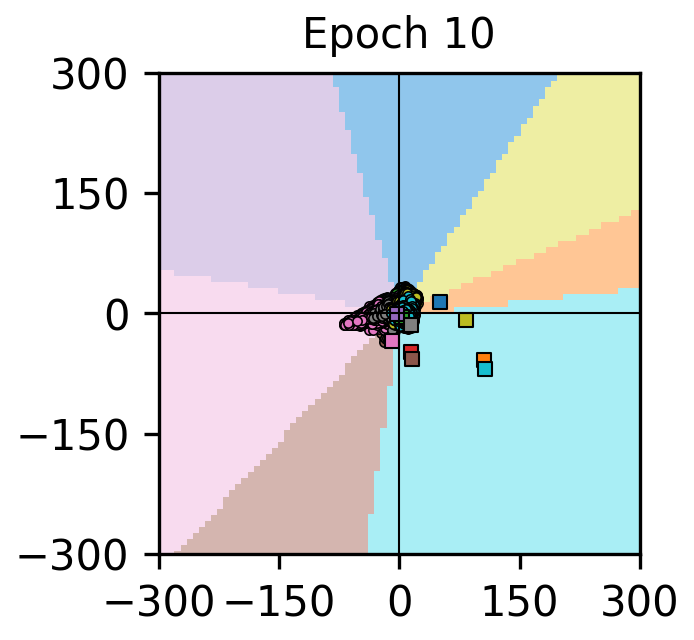}}
        {\includegraphics[width=0.25\columnwidth]{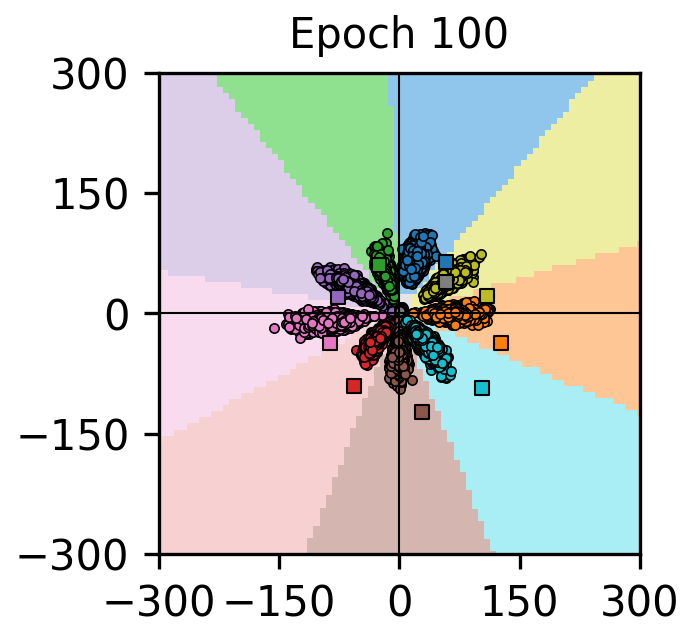}}
        \label{fig:2_b}
    }
    \\
    \subfloat[]
    {
        {\includegraphics[width=0.25\columnwidth]{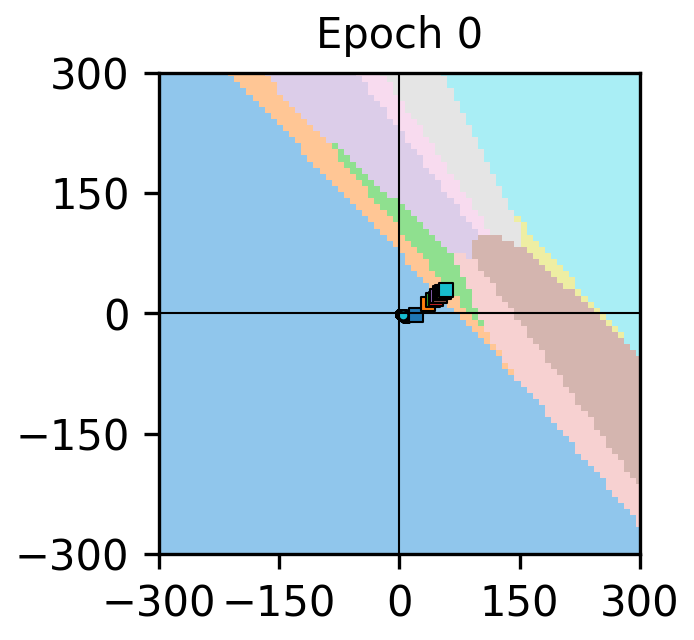}}
        {\includegraphics[width=0.25\columnwidth]{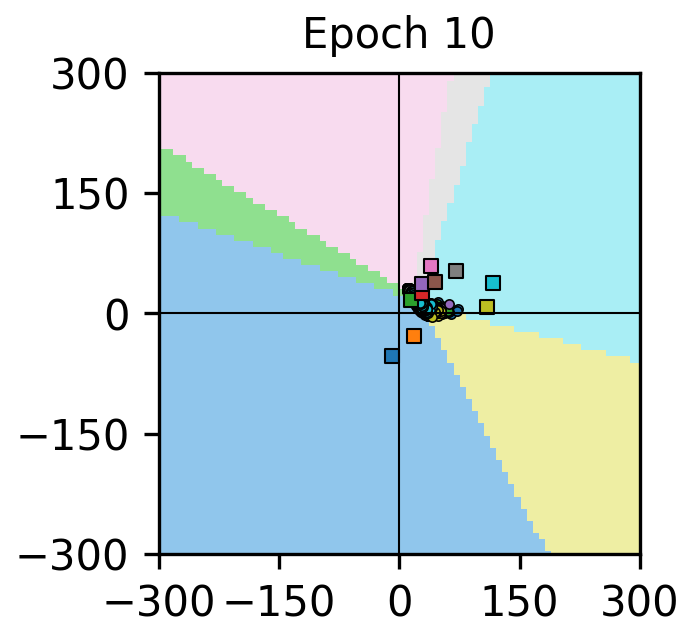}}
        {\includegraphics[width=0.25\columnwidth]{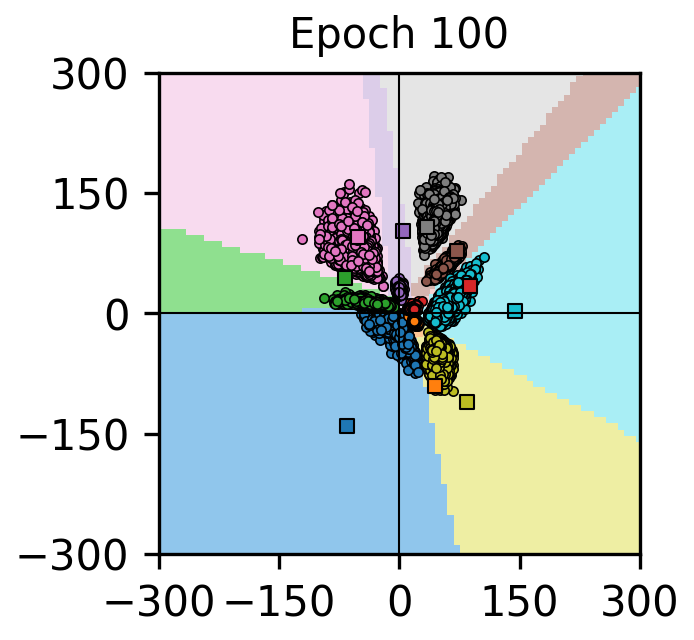}}
        \label{fig:2_c}
    }
    \\
    \subfloat[]
    {
        {\includegraphics[width=0.25\columnwidth]{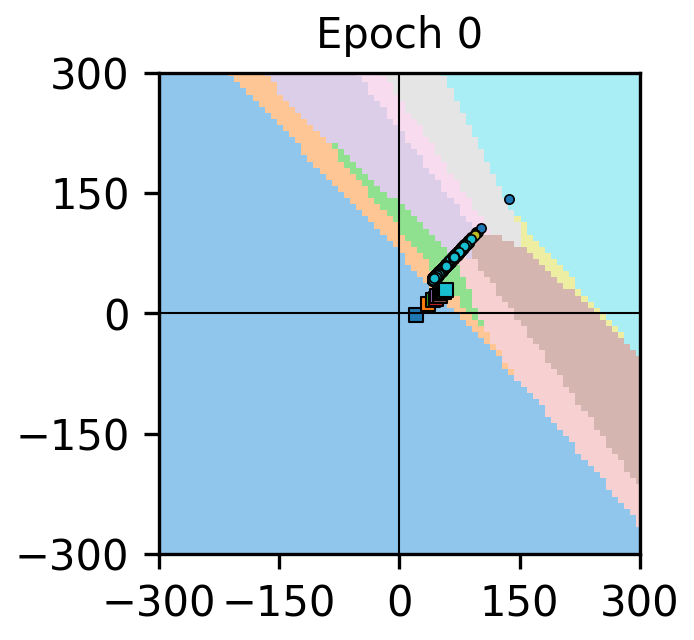}}
        {\includegraphics[width=0.25\columnwidth]{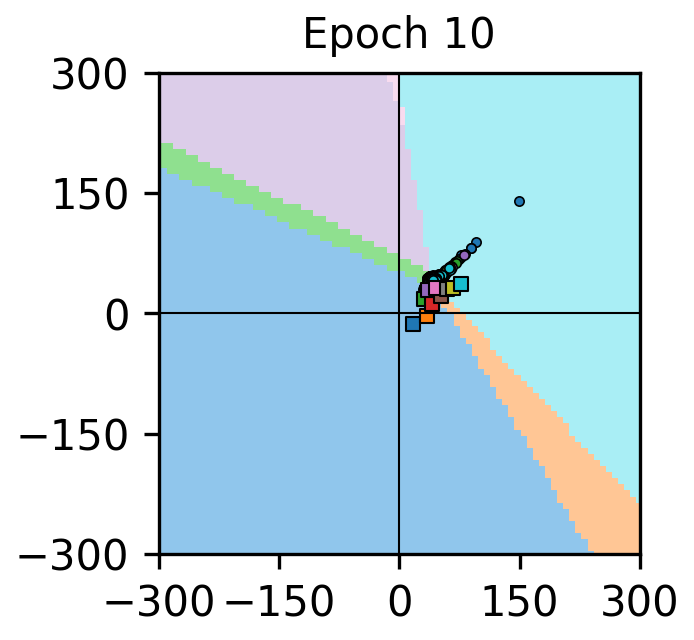}}
        {\includegraphics[width=0.25\columnwidth]{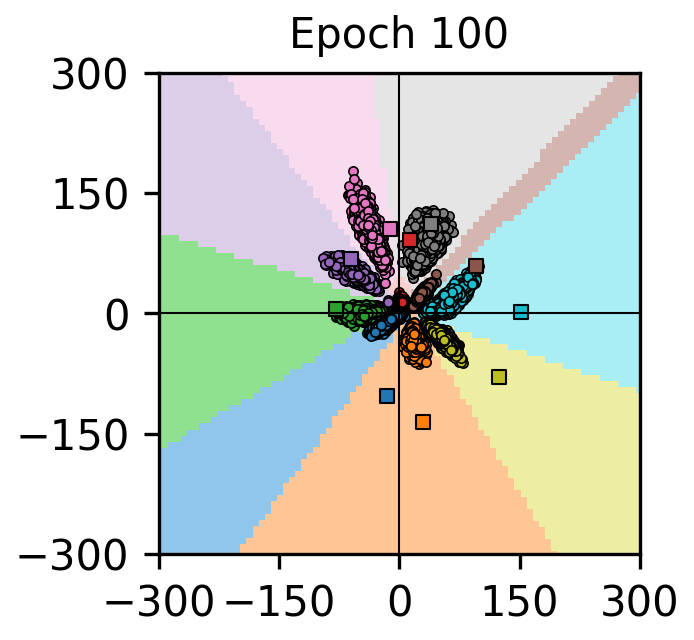}}
        \label{fig:2_d}
    }
    \caption{
        Changes in the 2D representation space as training progresses for ResNet50 on CIFAR-10 with different weight initializations. Features and weight vectors are represented as circles and squares, respectively (weight vectors are scaled to indicate directions).
    }
    \label{fig:2}
\end{figure*}

In this section, we verify whether the formation of cone-shaped decision regions around the origin depends on weight initialization.
Fig.~\ref{fig:2} shows the process of how the decision regions change during training of ResNet50 (with the classification layer bias) on CIFAR-10 with four different weight initializations (implementation details can be found in Appendix~\ref{appendix:implementation}).
In Fig.~\ref{fig:2_a}, we use the default Kaiming uniform weight initialization \citep{he2015delving}. Before training, the features are distributed close to the origin, as the model's weights are initialized as small values; some decision regions are divided around it, while for the other decision regions, due to random weight initialization, they have not yet been assigned to regions in the representation space.
In Fig.~\ref{fig:2_b}, we set the initial weights of the feature extractor in such a way that the features are positioned far from the origin while the weights of the classification layer are initialized as in Fig.~\ref{fig:2_a} so that the decision regions remain divided around the origin.
In Fig.~\ref{fig:2_c}, we set the initial weights of the classification layer so that the decision regions are divided far from the origin (but the initial features are located around the origin due to the default initialization of the feature extractor).
Finally, in Fig.~\ref{fig:2_d}, we adjust the initial weights of the entire model so that the center of the decision regions is shifted and the features are distributed far from the origin.
It can be observed from the figure that in all four cases, the decision regions eventually evolve into cone-like shapes around the origin as training progresses.

\newpage

\section{Implementation Details}
\label{appendix:implementation}

Here, we provide the implementation and training details for the models used in the main paper and Appendix.

\subsection{2D and 3D Representation Spaces}
\label{appendix:2d_3d_rep_space}

We train models on the CIFAR-10 dataset for 100 epochs using the SGD optimizer with a momentum of 0.9 and a weight decay of 0.0001. For learning rate scheduling, we apply a linear warmup from 0 to 0.01 over the first 10 epochs, followed by a cosine annealing scheduler for the remaining 90 epochs. Regarding regularization hyperparameters, we set the label smoothing value to 0.1 and use an alpha value of 0.2 for both Mixup and CutMix. However, for Mixup in Figs.~\ref{fig:rms_decrease_mixup}, \ref{fig:7_b}, and \ref{fig:8_b}, the alpha value is set to 0.05 for better visualization.

Regarding weight initialization in Fig.~\ref{fig:2}, to initialize the model to distribute features far from the origin, we first train the feature extractor alone, using the L2 norm between the extracted features and $\mathbf{v} = [80, 80]$ as the loss function for 1 epoch using the SGD optimizer with a learning rate of 0.0005. After this, we attach a linear classifier and train the full model using the training settings listed above.
To initialize the classifier so that the decision regions are divided far from the origin, we use hand-crafted weight matrix $\mathbf{W}$ and bias vector $\mathbf{b}$ in the classification layer, which are given as follows:

\[
\mathbf{W} = 
\begin{bmatrix}
  0.1318 & -0.0165 \\
  0.2245 &  0.0709 \\
  0.2630 &  0.1030 \\
  0.2881 &  0.1138 \\
  0.2947 &  0.1381 \\
  0.3189 &  0.1362 \\
  0.3195 &  0.1567 \\
  0.3323 &  0.1654 \\
  0.3482 &  0.1684 \\
  0.3619 &  0.1806 \\
\end{bmatrix}, \quad
\mathbf{b} = 
\begin{bmatrix}
  15.705 \\
   9.045 \\
   5.644 \\
   3.144 \\
   0.834 \\
  -1.286 \\
  -3.442 \\
  -5.799 \\
  -8.489 \\
 -11.972 \\
\end{bmatrix}.
\]



\subsection{Original Representation Spaces}
\label{appendix:orig_rep_space}

We train models on the CIFAR-10, CIFAR-100, and ImageNet datasets for 300 epochs using the AdamW optimizer \citep{loshchilov2017decoupled} with a weight decay of 0.0005. For learning rate scheduling, we apply a linear warmup from 0 to 0.001 over the first 20 epochs, followed by a cosine annealing scheduler for the remaining 280 epochs. When training Swin-T on CIFAR-10 and CIFAR-100, we increase the learning rate from 0 to 0.01 during the warmup phase to improve test accuracy. Regarding the regularization hyperparameters, we set the label smoothing value to 0.1 and use an alpha value of 1.0 for both Mixup and CutMix.

We trained models using a 12th Gen Intel® Core™ i7-12700K CPU and a NVIDIA RTX A6000 GPU. Specifically, training MobileNetV2 on the ImageNet dataset for 300 epochs required approximately 7 days.

The weights used for MobileNetV2, EfficientNet-B1 \citep{tan2019efficientnet}, and ViT-B/16 \citep{dosovitskiy2021an} in Tabs.~\ref{tab:appendix_bias_check}, \ref{tab:appendix_pytorch_settings}, and \ref{tab:appendix_more_results} are pretrained weights from PyTorch. In the PyTorch framework, these pretrained weights can be accessed using the names \texttt{IMAGENET1K\_V1}, \texttt{IMAGENET1K\_V2}, and \texttt{IMAGENET1K\_SWAG\_LINEAR\_V1}. Details on the regularization hyperparameters (using soft labels) used to train these weights can be found in Tab.~\ref{tab:appendix_pytorch_settings}.

\section{Representation Space Visualization}
\label{appendix:visualization}

To visualize the representation space in 2D (or 3D), we insert a linear layer between the feature extractor and the classification layer, mapping the output features into 2D (or 3D) vectors, and adjust the input dimension of the classification layer accordingly. As a result, the new representation space is 2D (or 3D), which facilitates visual examination, as in Figs.~\ref{fig:boundary_comparison}, \ref{fig:class_center_options}, \ref{fig:rms_decrease_2d}, \ref{fig:appendix_bias_2d_a}, \ref{fig:1}, \ref{fig:10}, and \ref{fig:2}.
To distinguish between different decision regions in the 2D (or 3D) representation space, we input a 2D grid (or 3D cube) with a fixed range into the classification layer. Each point in the grid or cube is then classified into a specific class. We visualize the decision regions by coloring each point in the grid or cube according to its predicted class. Next, using the values of the 2D (or 3D) feature vectors as coordinates, we plot their locations in the representation space, marking them with black-bordered circles colored by their predicted class.

\begin{table}[t]
\centering
\caption{Regularization hyperparameters to train models on ImageNet.}
\resizebox{0.7\columnwidth}{!}{
    \begin{tabular}{c c c c c}
    \toprule
    
    \bf Model & \bf Method & \bf Label smoothing & \bf Mixup & \bf CutMix \\ \hline
      \multirow{2}{*}{MobileNetV2}
      & PyTorch V1 & - & - & -  \\
     & PyTorch V2 & 0.10 & 0.2 & 1.0 \\
     \hline
    
      \multirow{2}{*}{EfficientNet-B1}
      & PyTorch V1 & - & - & - \\
     & PyTorch V2 & 0.10 & 0.2 & 1.0 \\
     \hline
    
      \multirow{2}{*}{ViT-B/16}
     & PyTorch Swag Linear V1 & - & 0.1 & - \\
      & PyTorch V1 & 0.11 & 0.2 & 1.0 \\
    
    \bottomrule
     
    \end{tabular}
}
\label{tab:appendix_pytorch_settings}
\end{table}

\begin{table*}[!t]
\centering
\caption{Overall performance and feature statistics (mean and standard deviation values) across various models and training methods. Red ECE values indicate overconfidence, while blue ECE values indicate underconfidence. For PGD and AutoAttack, we present results under two hyperparameter settings, with detailed configurations provided in Section~\ref{sec:4_4}.}
\resizebox{\textwidth}{!}{
\setlength{\tabcolsep}{4pt}
\begin{tabular}{c c c c c c c c c c c}
\toprule
\multirow{2.5}{*}{\bf Model} & \multirow{2.5}{*}{\bf Method} & \multirow{2.5}{*}{\shortstack{\bf Validation \\ \bf Accuracy (\%)}} & \multirow{2.5}{*}{\bf RMS} & \multirow{2.5}{*}{\shortstack{\bf Cosine \\ \bf Similarity}} & \multirow{2.5}{*}{\bf ECE} & \multicolumn{5}{c}{\bf Attack Success Rate (\%)}  \\
\cmidrule(lr){7-11} 
& & & & & & \bf FGSM & \bf PGD$^\text{w}$ & \bf PGD$^\text{s}$ & \bf AutoAttack$^\text{w}$ & \bf AutoAttack$^\text{s}$ \\
\midrule

  \multirow{2}{*}{MobileNetV2}
  & PyTorch V1 & 71.9 & 0.78 \textpm 0.09 & 0.31 \textpm 0.07 & \textcolor{red}{0.028} & 85.7 & 77.6 & 94.3 & 90.7 & 99.8 \\
 & PyTorch V2 & 72.0 & 0.28 \textpm 0.05 & 0.36 \textpm 0.09 & \textcolor{blue}{0.367} & 73.2 & 64.6 & 88.2 & 90.2 & 99.8\\
 \hline

  \multirow{2}{*}{EfficientNet-B1}
  & PyTorch V1 & 77.6 & 0.34 \textpm 0.08 & 0.32 \textpm 0.09 & \textcolor{blue}{0.091} & 65.1 & 48.8 & 81.4 & 69.3 & 99.6\\
 & PyTorch V2 & 78.9 & 0.15 \textpm 0.02 & 0.34 \textpm 0.07 & \textcolor{blue}{0.271} & 62.1 & 46.2 & 78.4 & 65.0 & 99.1 \\
 \hline

  \multirow{2}{*}{ViT-B/16}
 & PyTorch Swag Linear V1 & 81.8 & 1.28 \textpm 0.08 & 0.33 \textpm 0.08 & \textcolor{red}{0.018} & 58.8 & 48.5 & 79.5 & 71.3 & 99.7 \\
  & PyTorch V1 & 81.1 & 0.56 \textpm 0.09 & 0.57 \textpm 0.11 & \textcolor{blue}{0.055} & 54.9 & 37.6 & 65.2 & 51.3 & 98.5 \\

\bottomrule
 
\end{tabular}
}
\label{tab:appendix_more_results}
\end{table*}

\newpage

\section{Proof of Theorem 1}
\label{appendix:proof}

Let $y$ be the ground-truth class from the set of classes $c\in\{1,\dots,K\}$.
Define $z_c=\mathbf{w}_c^{\top}\mathbf{f}=\|\mathbf{w}\| \cdot \|\mathbf{f}\|\cos\theta_c$, where $\theta_c=\angle(\mathbf{w}_c,\mathbf{f})$.
Let the target vector be $\mathbf{t}=(t_1,\dots,t_K)$, where $t_y>\max_{c\neq y}t_c$ and $\sum_{c=1}^{K}t_c=1$.
The training objective to minimize cross-entropy can be written as

\begin{align}
\min \mathrm{CE}
&=\min\;-\sum_{c=1}^{K}
       t_c\,\log\frac{\exp(\mathbf{w}_c^{\top}\mathbf{f})}
                      {\sum_{j=1}^{K}\exp(\mathbf{w}_{j}^{\top}\mathbf{f})}
                      \tag{1}\\[2pt]
&=\min\;-\sum_{c=1}^{K}
       \log\!\left(
       \frac{\exp(\mathbf{w}_c^{\top}\mathbf{f})}
            {\sum_{j=1}^{K}\exp(\mathbf{w}_{j}^{\top}\mathbf{f})}
       \right)^{t_c}                                   \tag{2}\\[2pt]
&=\min\;-\prod_{c=1}^{K}
       \left(
       \frac{\exp(\mathbf{w}_c^{\top}\mathbf{f})}
            {\sum_{j=1}^{K}\exp(\mathbf{w}_{j}^{\top}\mathbf{f})}
       \right)^{t_c}                                   \tag{3}\\[4pt]
&=\min\;-\prod_{c=1}^{K}
       \frac{\exp\!\bigl(t_c\mathbf{w}_c^{\top}\mathbf{f}\bigr)}
            {\Bigl(\sum_{j=1}^{K}\exp(\mathbf{w}_{j}^{\top}\mathbf{f})\Bigr)^{t_c}}
                                                    \tag{4}\\[4pt]
&=\min\;-
       \frac{\exp\!\bigl(\sum_{c=1}^{K}t_c\mathbf{w}_c^{\top}\mathbf{f}\bigr)}
            {\Bigl(\sum_{c=1}^{K}\exp(\mathbf{w}_{c}^{\top}\mathbf{f})\Bigr)^{\sum_{c}t_c}}
                                                    \tag{5}\\[4pt]
&=\min\;-
       \frac{\exp\!\bigl(\sum_{c=1}^{K}t_c\mathbf{w}_c^{\top}\mathbf{f}\bigr)}
            {\sum_{c=1}^{K}\exp(\mathbf{w}_{c}^{\top}\mathbf{f})}
                                                    \tag{6}\\[4pt]
&=\min\;\underbrace{-
       \frac{\exp\!\bigl(\|\mathbf{w}\|\cdot\|\mathbf{f}\|
                 \sum_{c=1}^{K}t_c\cos\theta_c\bigr)}
            {\sum_{c=1}^{K}\exp\!\bigl(\|\mathbf{w}\|\cdot\|\mathbf{f}\|
                                   \cos\theta_c\bigr)}
                                                    \tag{7}
                                                    }_{\triangleq L}\\
&= \min L                                    \tag{8}
\end{align}
we use $||\mathbf{w}_{c}||= ||\mathbf{w}||$ as assumed.


To solve this optimization problem, let us compute the derivative of $L$ with respect to $||\mathbf{f}||$. For the sake of notation simplicity, let us denote $A
\triangleq\|\mathbf{w}\| \cdot \|\mathbf{f}\|\sum_{c}t_c\cos\theta_c$ and
$B\triangleq\sum_{c}\exp\!\bigl(\|\mathbf{w}\|\cdot\|\mathbf{f}\|\cos\theta_c\bigr)$.
Then, $L=-\exp(A)/B$.  Using
$\partial z_c/\partial\|\mathbf{f}\|=\|\mathbf{w}\|\cos\theta_c$, we get

\begin{align}
\frac{\partial L}{\partial\|\mathbf{f}\|}
&=-\frac{\|\mathbf{w}\|
        \Bigl(\sum_{c}t_c\cos\theta_c\Bigr)\,
        \exp(A)\,B}
       {B^{2}}
  \notag\\
&\quad
      +\frac{\exp(A)\,
        \Bigl(\|\mathbf{w}\|\sum_{c}\cos\theta_c\,
              \exp(z_c)\Bigr)}
            {B^{2}}                                    \tag{9}\\[4pt]
&=\underbrace{\frac{\|\mathbf{w}\|\exp(A)}{B^{2}}}_{\triangleq C}
  \notag\\
&\quad
   \times \underbrace{\Bigl\{
     -\Bigl(\sum_{c}t_c\cos\theta_c\Bigr)B
     +\sum_{c}\cos\theta_c\,\exp(z_c)
   \Bigr\}}_{\triangleq D} \tag{10}
\end{align}
Note that $C>0$.

$D$ can be further written as

\begin{align}
D
&=\sum_{c}\exp(z_c)\,\Bigl(\cos\theta_c-\sum_{j}t_j\cos\theta_j\Bigr)
                                                    \tag{11}\\[2pt]
&=B\,\Bigl(\sum_{c}p_c\cos\theta_c-\sum_{c}t_c\cos\theta_c\Bigr),
                     \tag{12}
\end{align}
where we define $p_c\triangleq\frac{\exp(z_c)}{B}$.

Thus,

\begin{equation}
\frac{\partial L}{\partial\|\mathbf{f}\|}
=C\,B\Bigl(\sum_{c}p_c\cos\theta_c-\sum_{c}t_c\cos\theta_c\Bigr)
.                                                 \tag{13}
\end{equation}

\paragraph{(i) Hard‑label training.}
For hard labels, \ie, $t_y=1$ and $t_{c\neq y}=0$, we have
$\sum_{c}t_c\cos\theta_c=\cos\theta_y$.
Hence,

\begin{align}
\frac{\partial L_{hard}}{\partial\|\mathbf{f}\|}
&=C\,B\Bigl(\sum_{c}p_c\cos\theta_c-\cos\theta_y\Bigr)      \tag{14}\\
&=C\,B\sum_{c\neq y}(\cos\theta_c-\cos\theta_y)\,p_c        \tag{15}\\
&<0 \notag
\end{align}
as $B>0$ from definition and $\cos\theta_y>\cos\theta_{c\neq y}$ under the assumption that the feature vector is better aligned with the correct class than with incorrect classes.

Therefore, the loss \emph{always} decreases as $\|\mathbf{f}\|$ increases, 
and the optimal solution is achieved at $\|\mathbf{f}\|\to\infty$. 
In other words,
\[
\|\mathbf{f}_{hard}^{*}\|\to\infty.
\]

\paragraph{(ii) Soft‑label training.}
Eq. (13) applies directly to soft targets:

\begin{align}
\frac{\partial L_{soft}}{\partial\|\mathbf{f}\|}
&=C\,B\Bigl(\sum_{c}p_c\cos\theta_c-\sum_{c}t_c\cos\theta_c\Bigr).
                                                        \tag{16}
\end{align}

\begin{itemize}
\item \textbf{Small $\|\mathbf{f}\|$ ($\|\mathbf{f}\|\to0$):}\; 
Since $p_c\approx\frac{1}{K},~\forall c,$ we have
\begin{equation}
\sum_c p_c\cos\theta_c
\approx\frac{1}{K}\sum_c\cos\theta_c
<\sum_c t_c\cos\theta_c,
\tag{17}
\end{equation}
by assuming that $\cos\theta_y$ is sufficiently larger than $\cos\theta_{c\neq y}$. Then, eq. (16) becomes negative.
\item \textbf{Large $\|\mathbf{f}\|$ ($\|\mathbf{f}\|\to\infty$):}\;
Since $p_y\to1$ and $p_{c\neq y}\to0$, we have
\begin{equation}
\sum_c p_c\cos\theta_c\to\cos\theta_y>\sum_c t_c\cos\theta_c,
\tag{18}
\end{equation}
thus eq. (16) is positive.
\end{itemize}

By continuity, there exists $\|\mathbf{f}_{soft}^{*}\|<\infty$ such that the gradient becomes 0, \ie,
\begin{equation}
\sum_{c=1}^{K}
\Bigl(t_c - p_c^{*}\Bigr)\cos\theta_c
\;=\;0,
\tag{19}
\end{equation}
where
\begin{equation}
p_c^{*}=
\frac{\exp\!\bigl(\|\mathbf{w}\|\cdot\|\mathbf{f}_{soft}^{*}\|\cos\theta_c\bigr)}
     {\sum_{j}\exp\!\bigl(\|\mathbf{w}\|\cdot\|\mathbf{f}_{soft}^{*}\|\cos\theta_j\bigr)} \tag{20}
\end{equation}
The solution $||\mathbf{f}_{soft}^*||$ is \emph{finite}, and thereafter, $L$ no longer decreases.

\paragraph{(iii) Conclusion.}
Hard-label training pushes $\|\mathbf{f}\|$ toward infinity,  
but soft-label training stops at a \emph{finite} magnitude satisfying eq. (20). Thus,

\[
\|\mathbf{f}_{hard}^{*}\| \;>\;
\|\mathbf{f}_{soft}^{*}\|.
\]

\section{More Results on the Effect of Regularization}
\label{appendix:more_result_regularization}

\begin{figure*}
    \centering
    \subfloat[Baseline]
    {
        {\includegraphics[width=0.22\columnwidth]{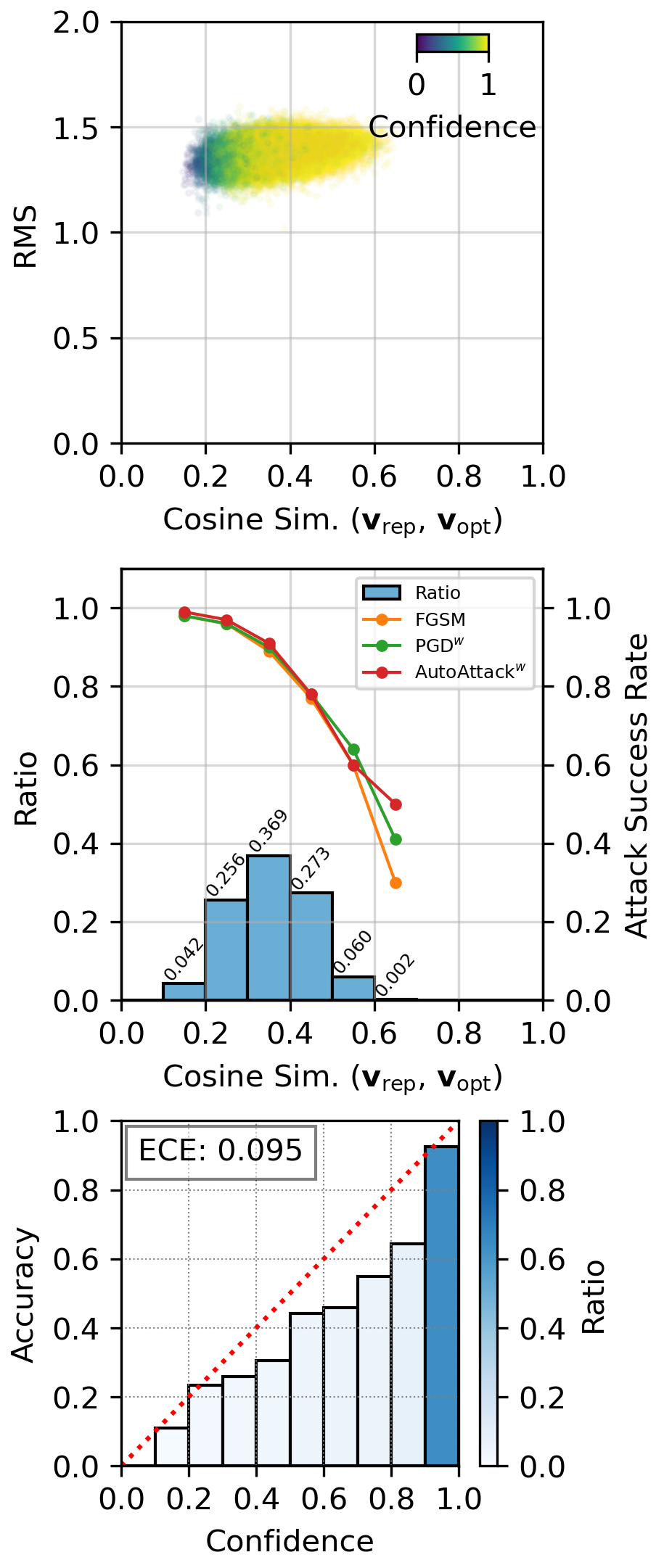}}
    }
    \subfloat[Label smoothing]
    {
        {\includegraphics[width=0.22\columnwidth]{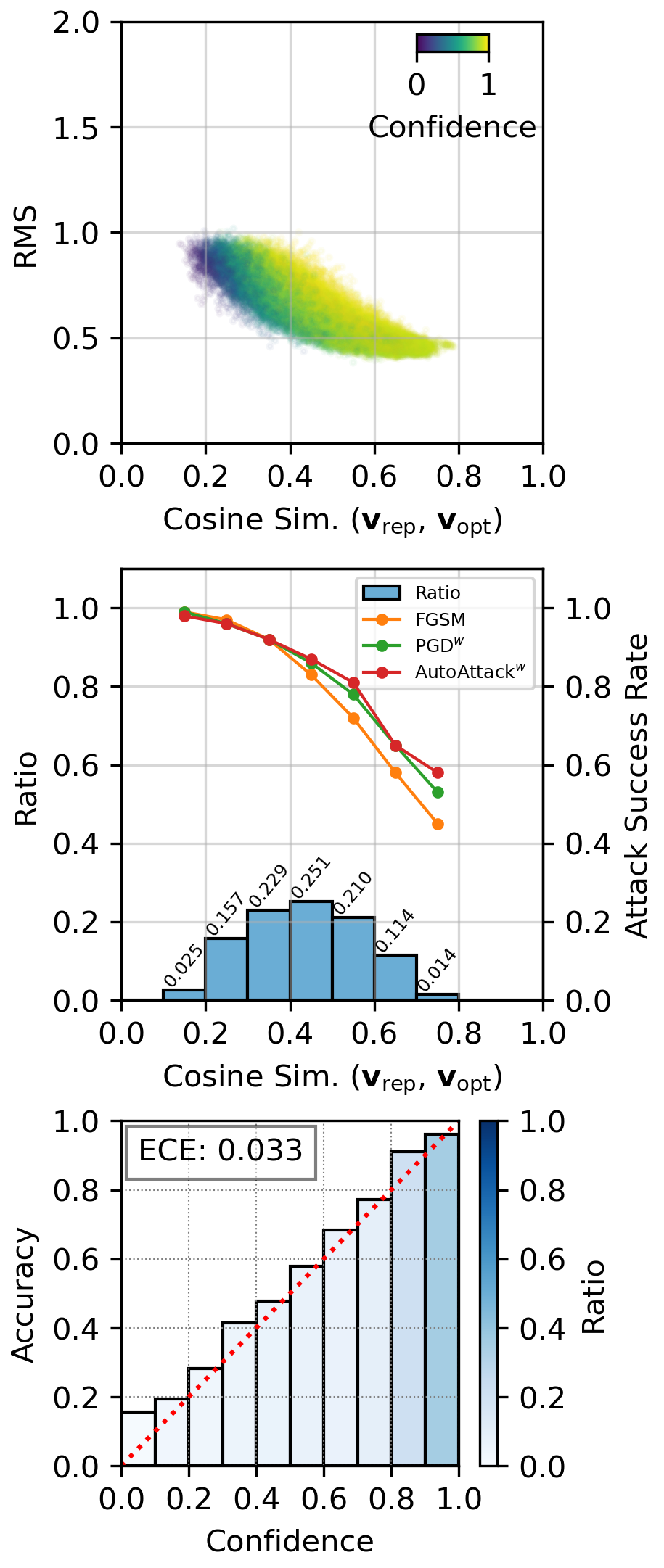}}
    }
    \subfloat[Mixup]
    {
        {\includegraphics[width=0.22\columnwidth]{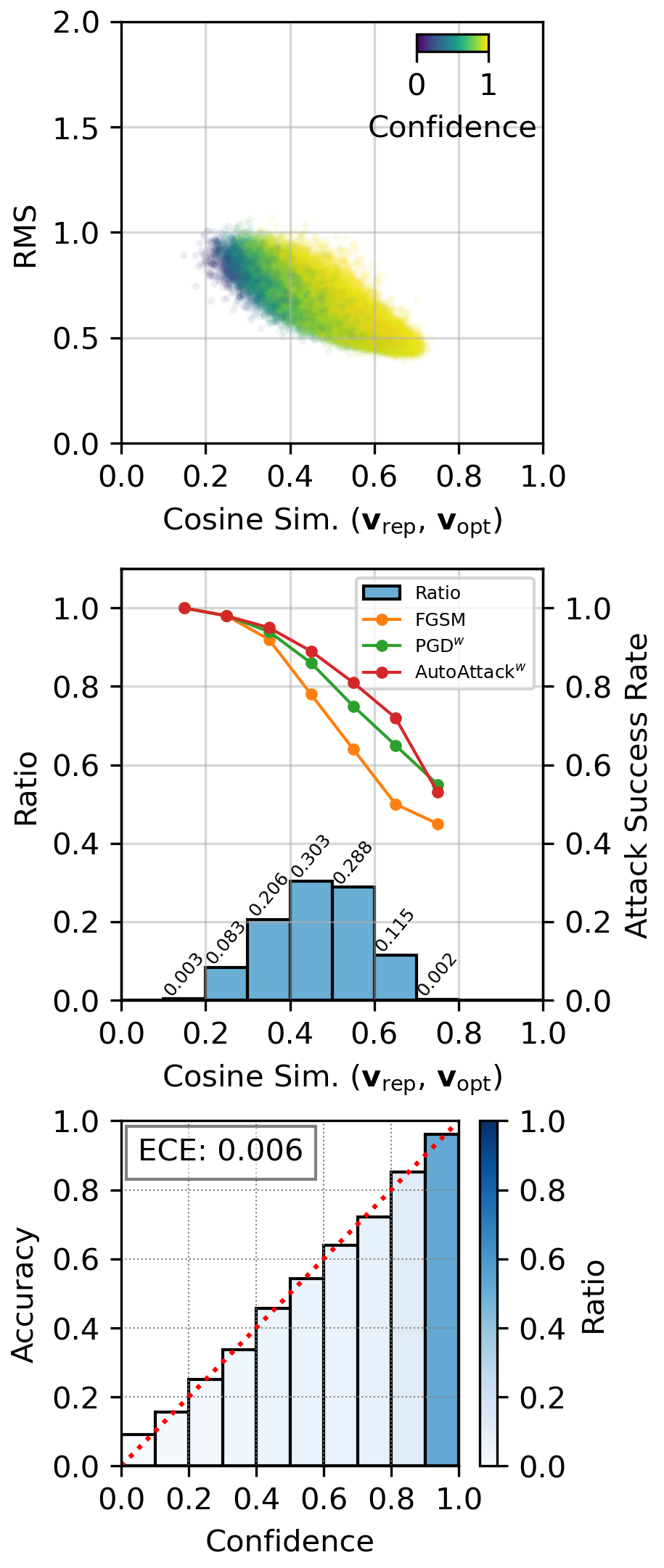}}
    }
    \subfloat[CutMix]
    {
        {\includegraphics[width=0.22\columnwidth]{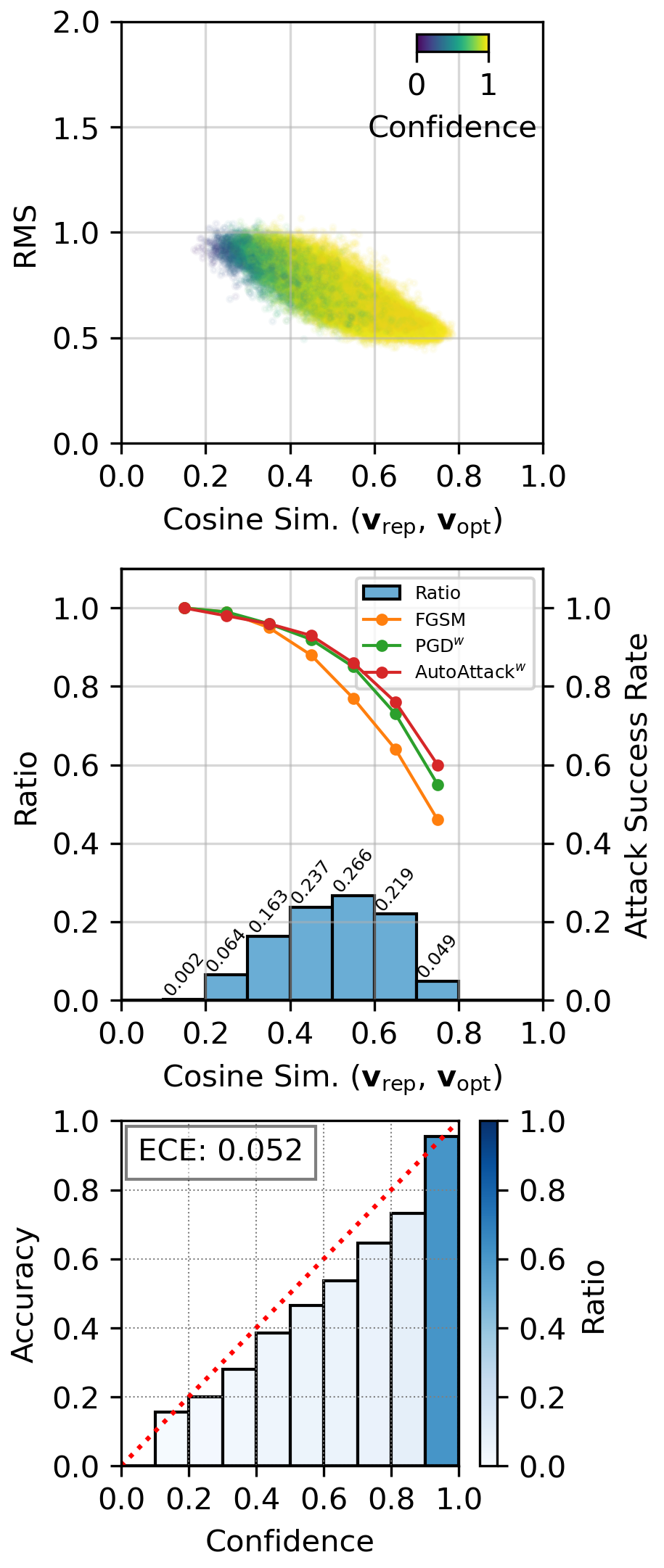}}
    }

    \caption{
    Evaluation results of Swin-T on the ImageNet validation data.
    \textbf{Top.} Scatter plots of feature RMS and cosine similarities of features ($\mathbf{v}_\text{rep}$) with the class center ($\mathbf{v}_\text{opt}$). Colors represent confidence values.
    \textbf{Middle.} Histograms of cosine similarities of features to class centers, along with the attack success rates of FGSM, PGD$^\text{w}$, and AutoAttack$^\text{w}$ for each bin (hyperparameters settings for the attacks can be found in Section~\ref{sec:4_4}).
    \textbf{Bottom.} Reliability diagrams, where the transparency of bars represents the ratio of data in each confidence bin. Expected calibration error (ECE) \citep{guo2017calibration} values are shown for each case.
    }
    \label{fig:swin_t}
\end{figure*}

\begin{figure*}
    \centering
    \subfloat[Baseline]
    {
        {\includegraphics[width=0.22\columnwidth]{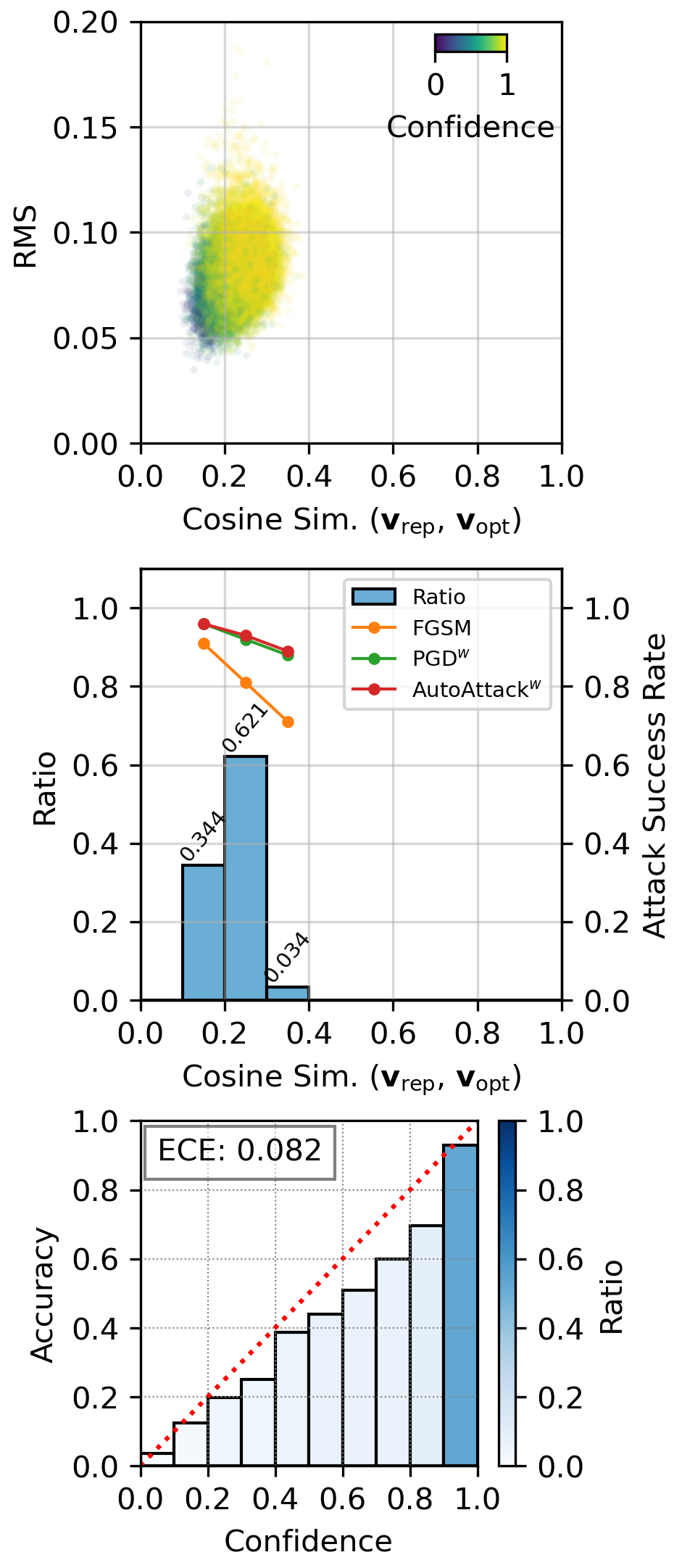}}
    }
    \subfloat[Label smoothing]
    {
        {\includegraphics[width=0.22\columnwidth]{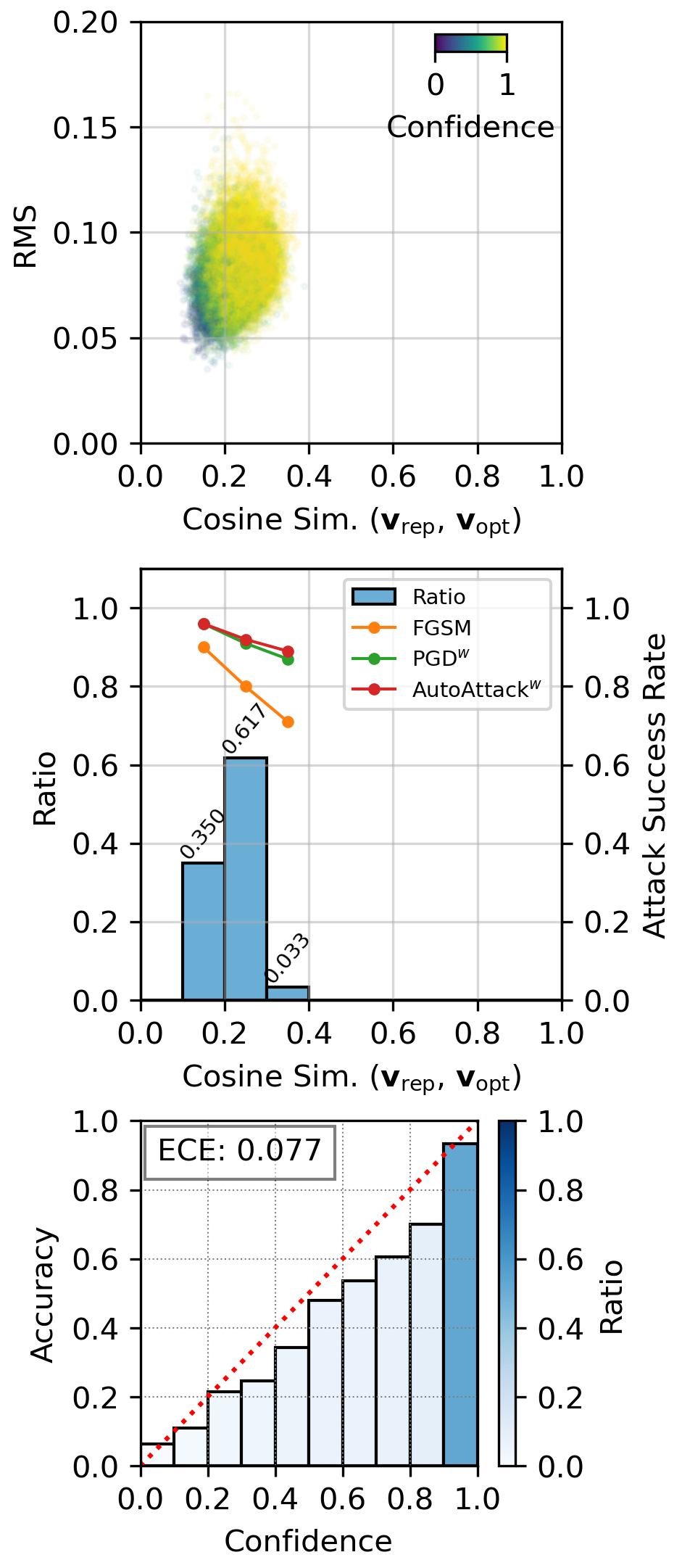}}
    }
    \subfloat[Mixup]
    {
        {\includegraphics[width=0.22\columnwidth]{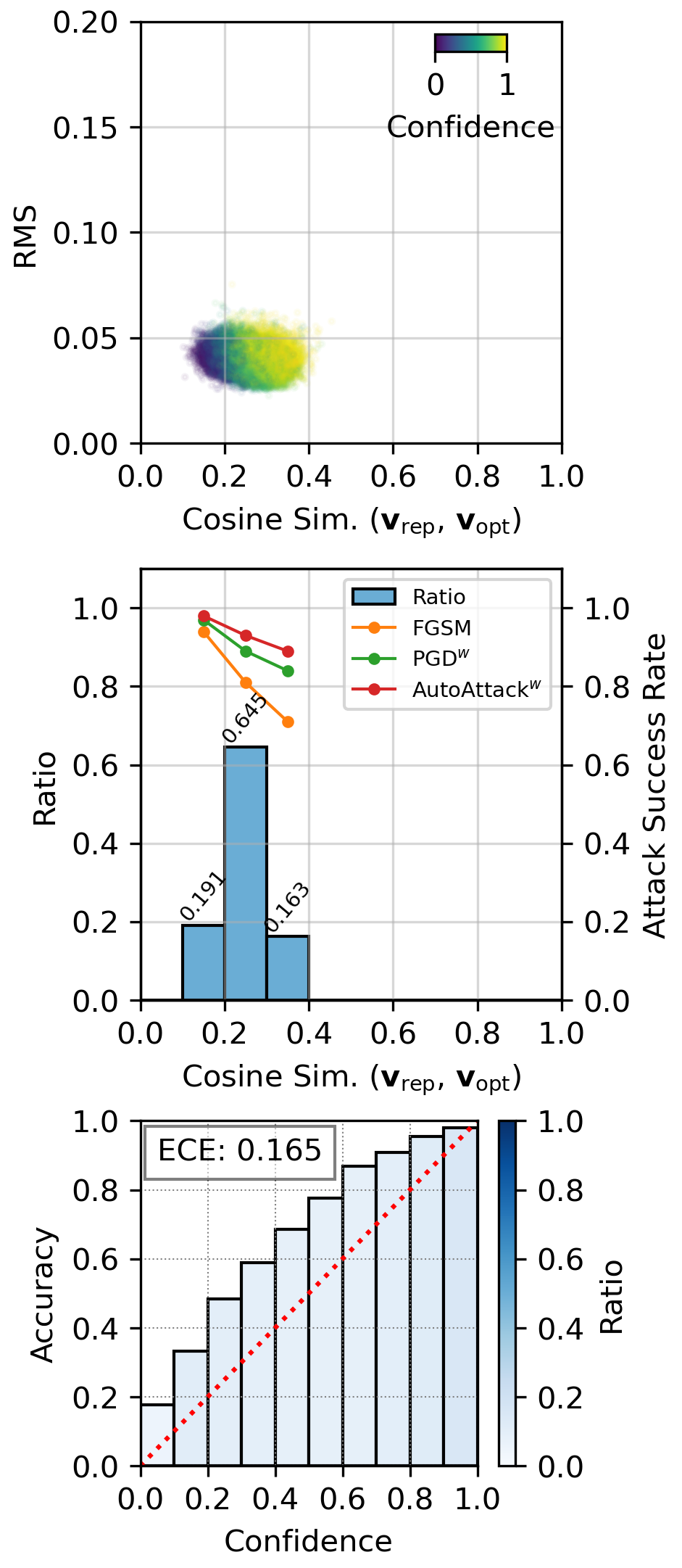}}
    }
    \subfloat[CutMix]
    {
        {\includegraphics[width=0.22\columnwidth]{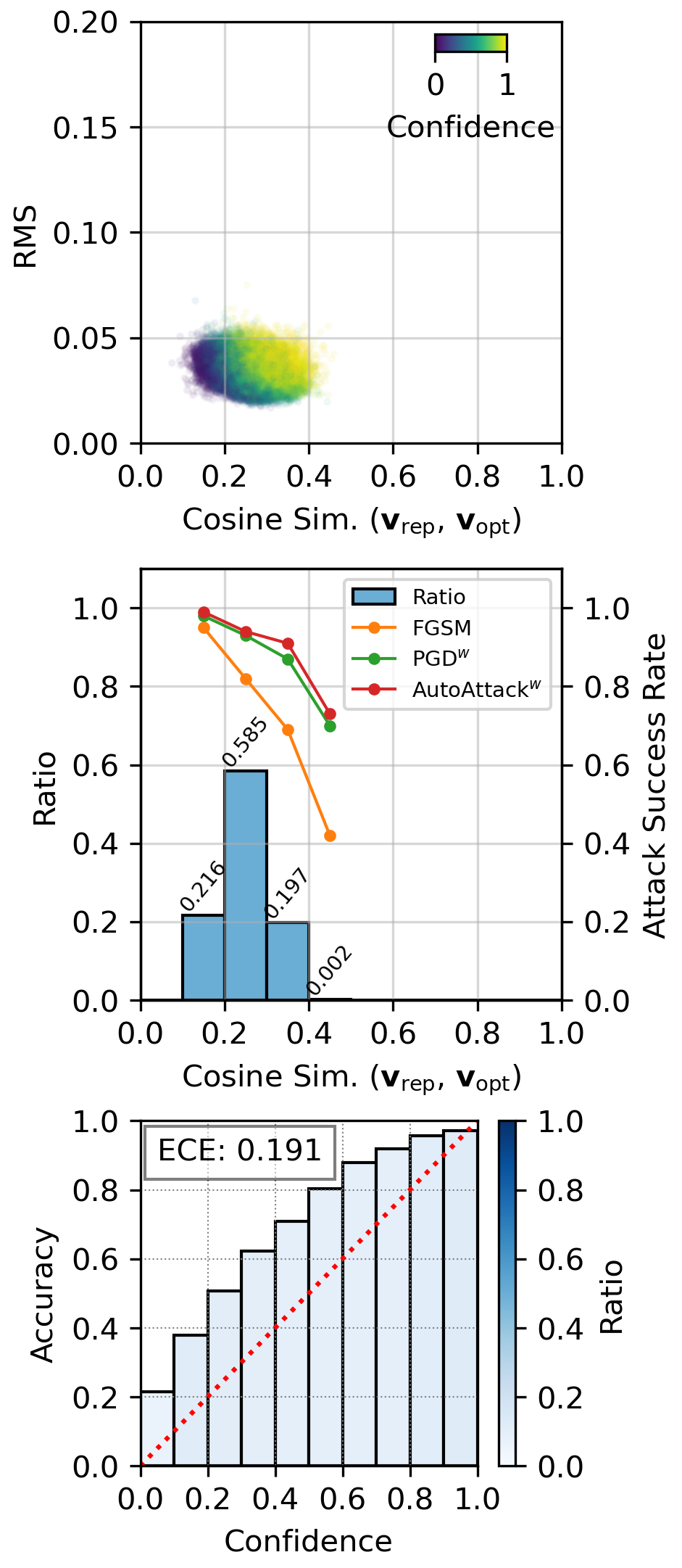}}
    }

    \caption{
    Evaluation results of MobileNetV2 on the ImageNet validation data.
    \textbf{Top.} Scatter plots of feature RMS and cosine similarities of features ($\mathbf{v}_\text{rep}$) with the class center ($\mathbf{v}_\text{opt}$). Colors represent confidence values.
    \textbf{Middle.} Histograms of cosine similarities of features to class centers, along with the attack success rates of FGSM, PGD$^\text{w}$, and AutoAttack$^\text{w}$ for each bin (hyperparameters settings for the attacks can be found in Section~\ref{sec:4_4}).
    \textbf{Bottom.} Reliability diagrams, where the transparency of bars represents the ratio of data in each confidence bin. Expected calibration error (ECE) \citep{guo2017calibration} values are shown for each case.
    }
    \label{fig:mobilenet_v2}
\end{figure*}

\begin{figure*}
    \centering
    \subfloat[Baseline]
    {
        {\includegraphics[width=0.22\columnwidth]{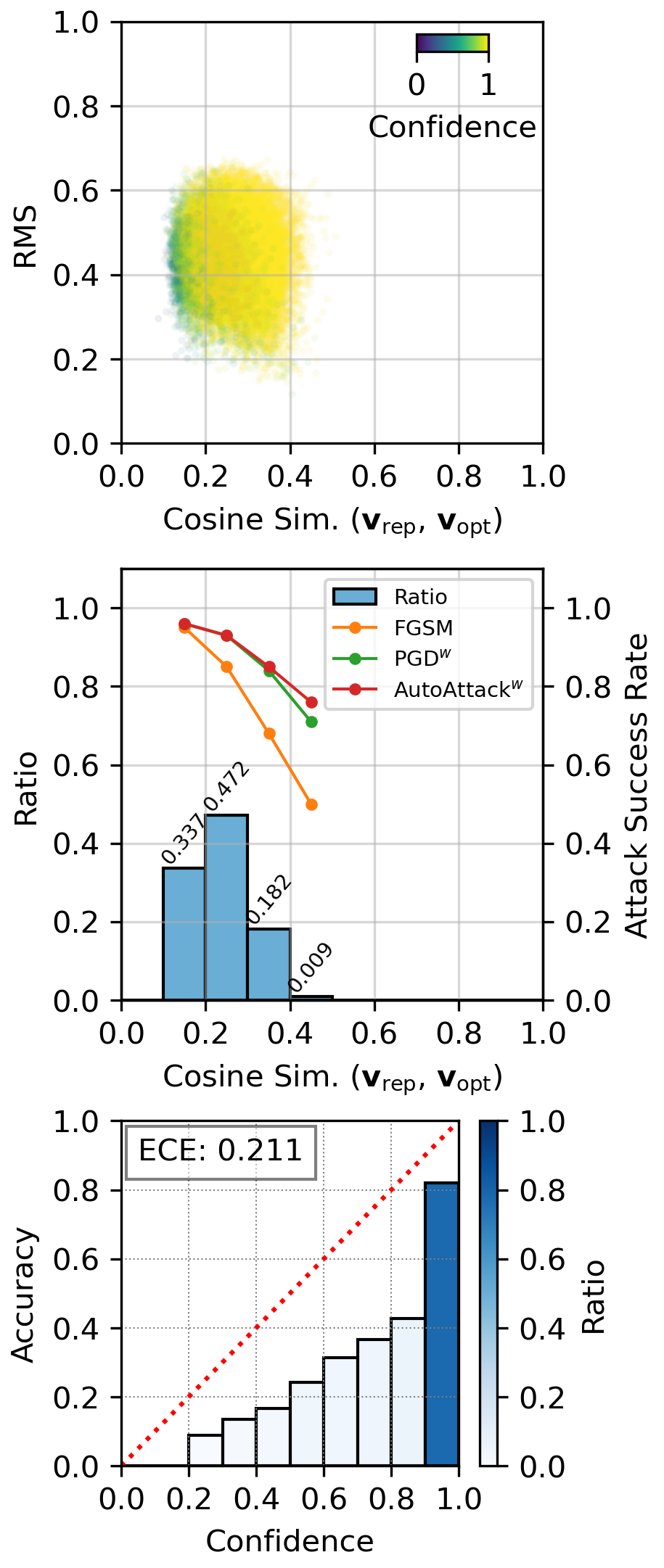}}
    }
    \subfloat[Label smoothing]
    {
        {\includegraphics[width=0.22\columnwidth]{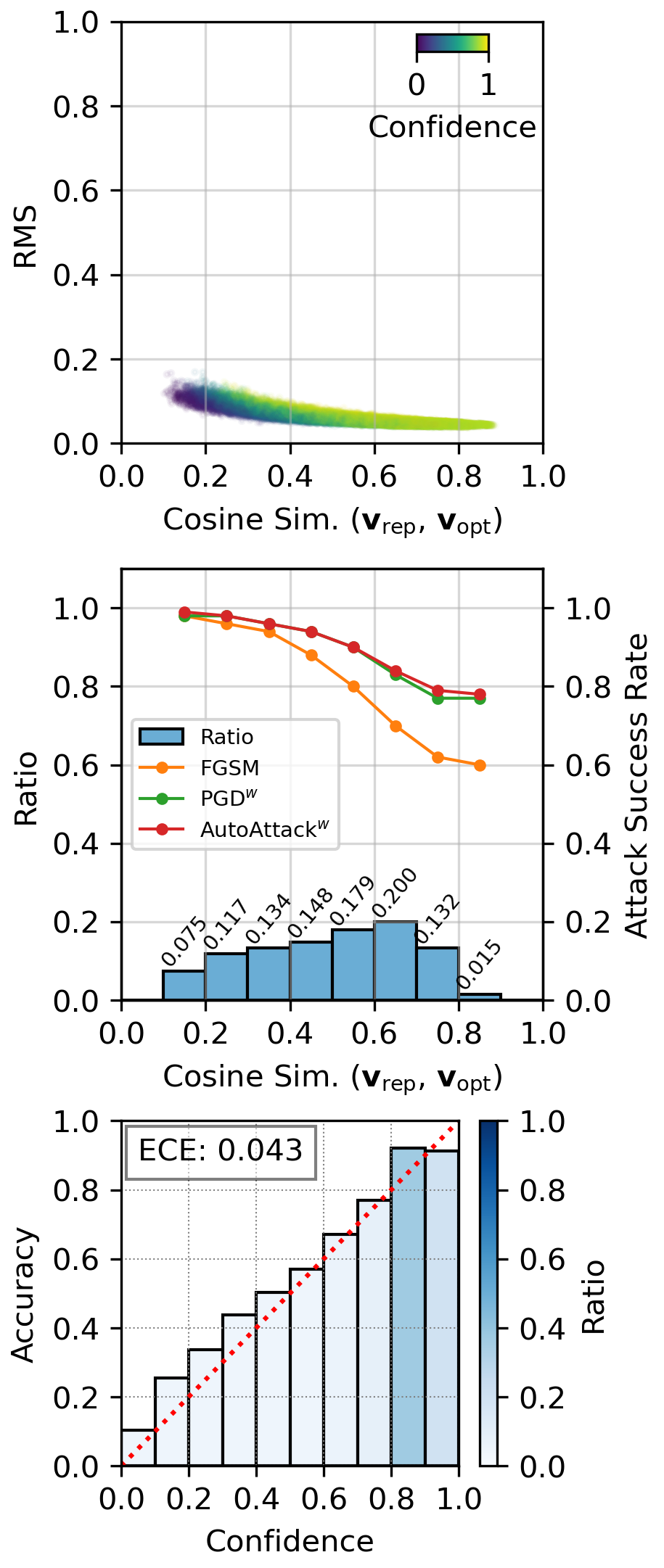}}
    }
    \subfloat[Mixup]
    {
        {\includegraphics[width=0.22\columnwidth]{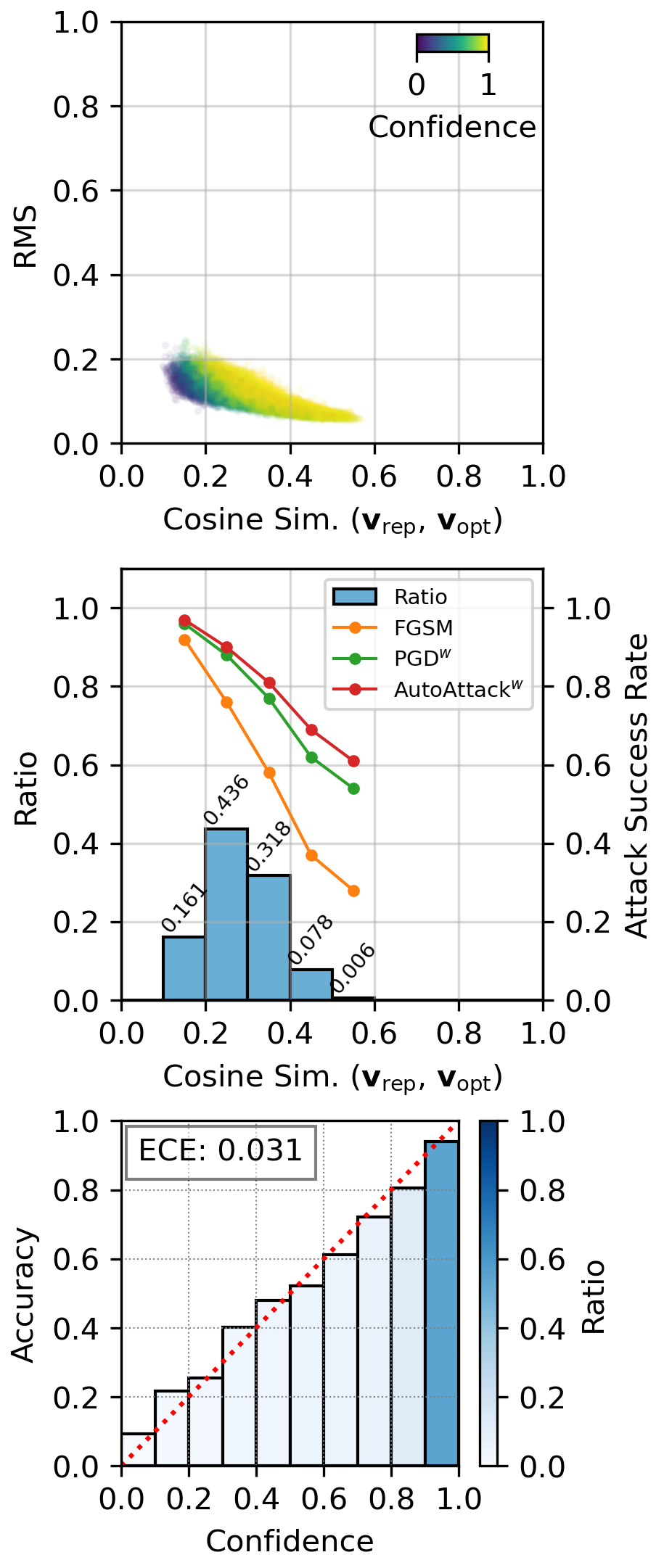}}
    }

    \caption{
    Evaluation results of ConvNeXt-T on the ImageNet validation data.
    \textbf{Top.} Scatter plots of feature RMS and cosine similarities of features ($\mathbf{v}_\text{rep}$) with the class center ($\mathbf{v}_\text{opt}$). Colors represent confidence values.
    \textbf{Middle.} Histograms of cosine similarities of features to class centers, along with the attack success rates of FGSM, PGD$^\text{w}$, and AutoAttack$^\text{w}$ for each bin (hyperparameters settings for the attacks can be found in Section~\ref{sec:4_4}).
    \textbf{Bottom.} Reliability diagrams, where the transparency of bars represents the ratio of data in each confidence bin. Expected calibration error (ECE) \citep{guo2017calibration} values are shown for each case.
    }
    \label{fig:convnext_tiny}
\end{figure*}

\begin{figure*}
    \centering

    \subfloat[{\begin{tabular}{c}MobileNetV2 \\ V1\end{tabular}}]
    {
        {\includegraphics[width=0.22\columnwidth]{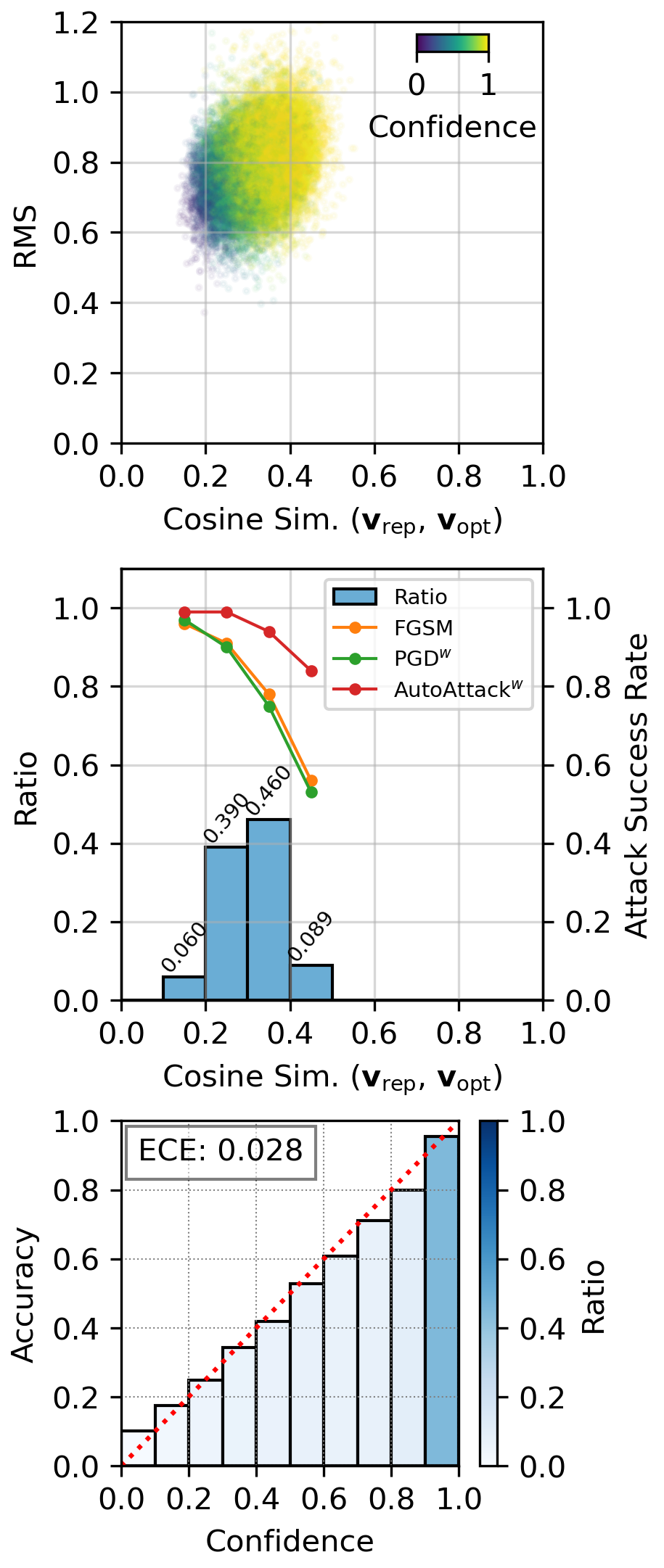}}
    }
    \subfloat[{\begin{tabular}{c}MobileNetV2 \\ V2\end{tabular}}]
    {
        {\includegraphics[width=0.22\columnwidth]{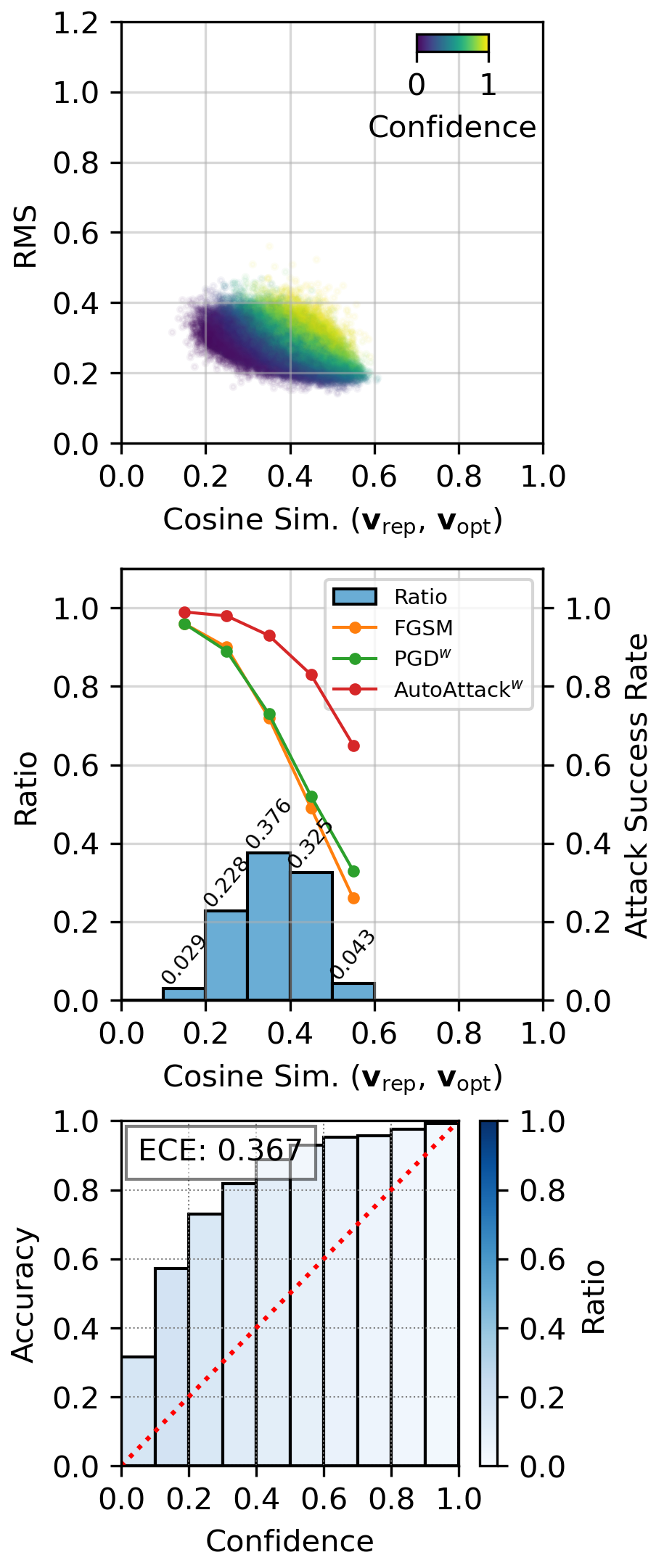}}
    }
    \subfloat[{\begin{tabular}{c}EfficientNet-B1 \\ V1\end{tabular}}]
    {
        {\includegraphics[width=0.22\columnwidth]{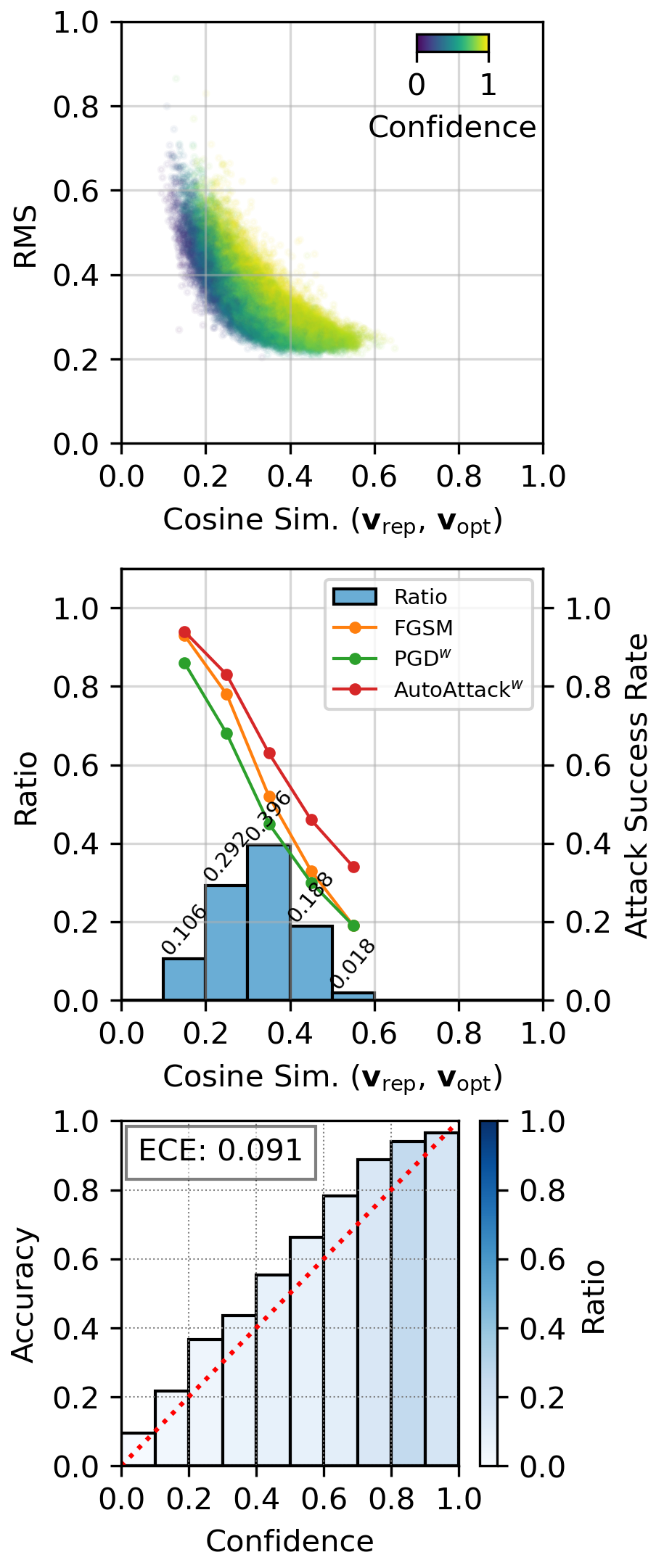}}
    }
    \subfloat[{\begin{tabular}{c}EfficientNet-B1 \\ V2\end{tabular}}]
    {
        {\includegraphics[width=0.22\columnwidth]{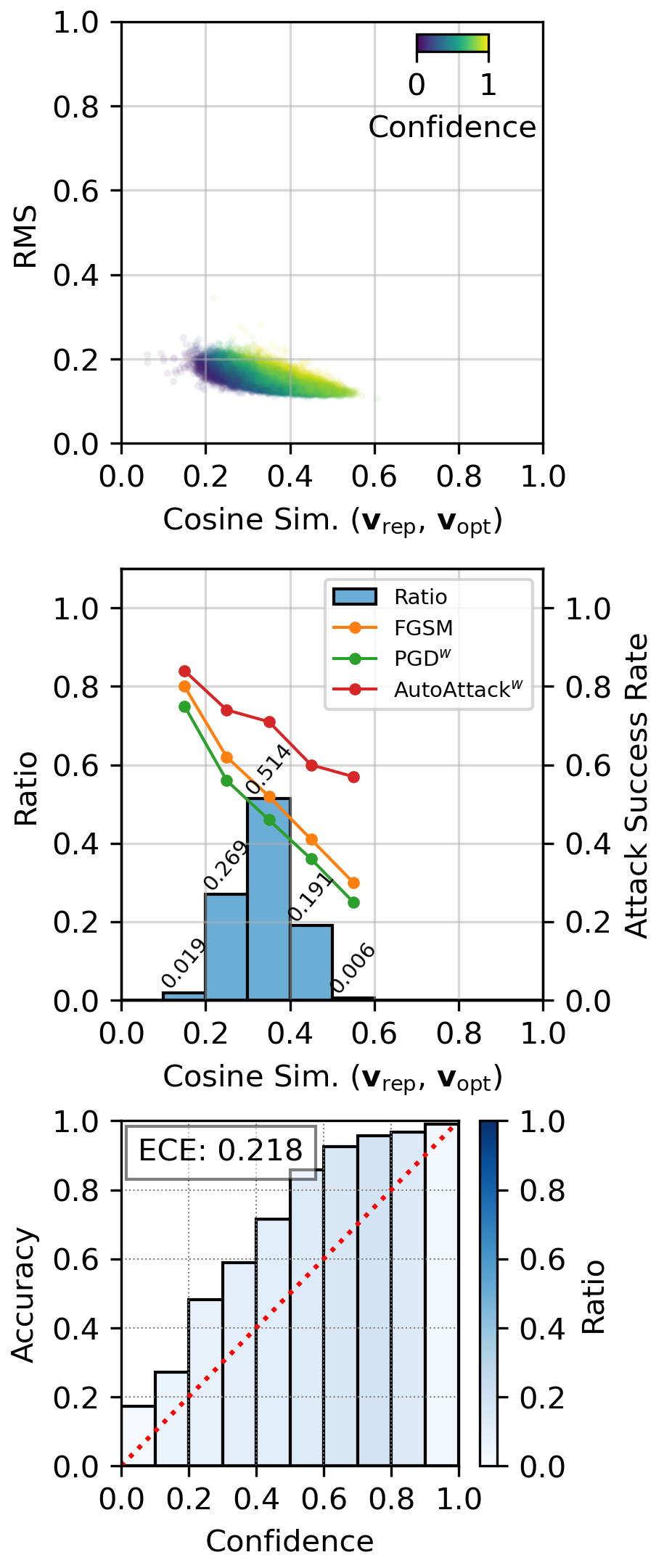}}
    }\hspace{0.0cm}
    \subfloat[{\begin{tabular}{c}ViT-B/16 \\ Swag Linear V1\end{tabular}}]
    {
        \includegraphics[width=0.22\columnwidth]{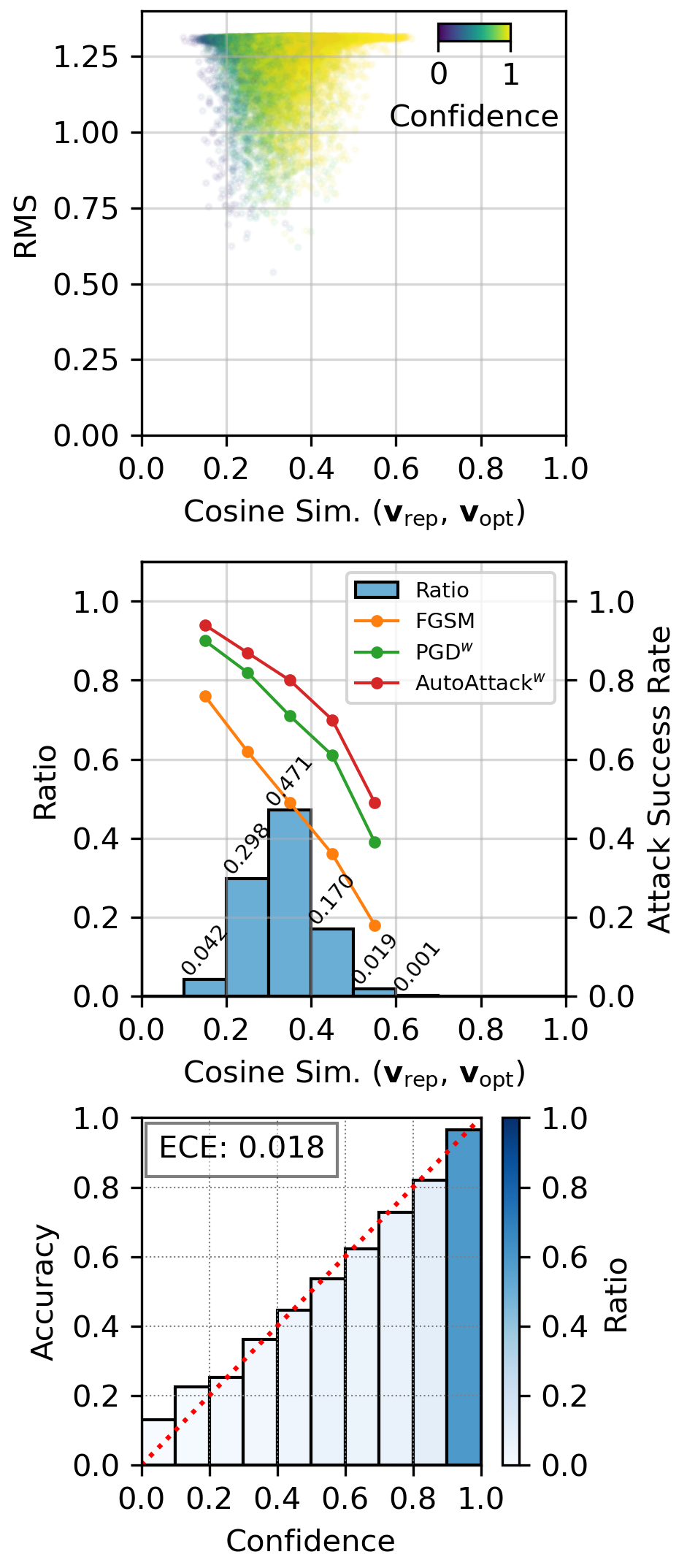}
    }
    \subfloat[{\begin{tabular}{c}ViT-B/16 \\ V1\end{tabular}}]
    {
        {\includegraphics[width=0.22\columnwidth]{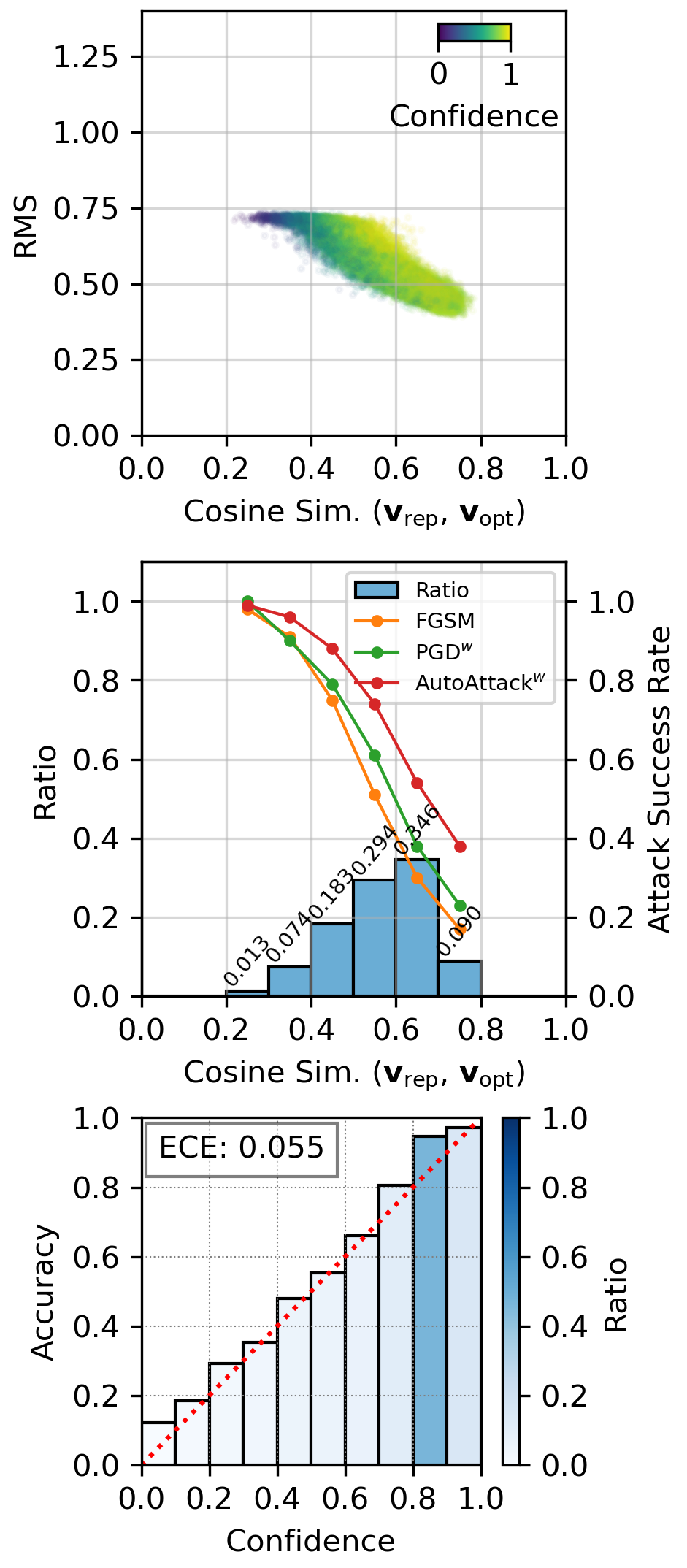}}
    }

    \caption{
    Evaluation results of MobileNetV2, EfficientNet-B1, and ViT-B/16 on the ImageNet validation data.
    \textbf{Top.} Scatter plots of feature RMS and cosine similarities of features ($\mathbf{v}_\text{rep}$) with the class center ($\mathbf{v}_\text{opt}$). Colors represent confidence values.
    \textbf{Middle.} Histograms of cosine similarities of features to class centers, along with the attack success rates of FGSM, PGD$^\text{w}$, and AutoAttack$^\text{w}$ for each bin (hyperparameters settings for the attacks can be found in Section~\ref{sec:4_4}).
    \textbf{Bottom.} Reliability diagrams, where the transparency of bars represents the ratio of data in each confidence bin. Expected calibration error (ECE) \citep{guo2017calibration} values are shown for each case.
    }
    \label{fig:13}
\end{figure*}

\begin{figure*}
    \centering
    \subfloat[Baseline]
    {
        {\includegraphics[width=0.22\columnwidth]{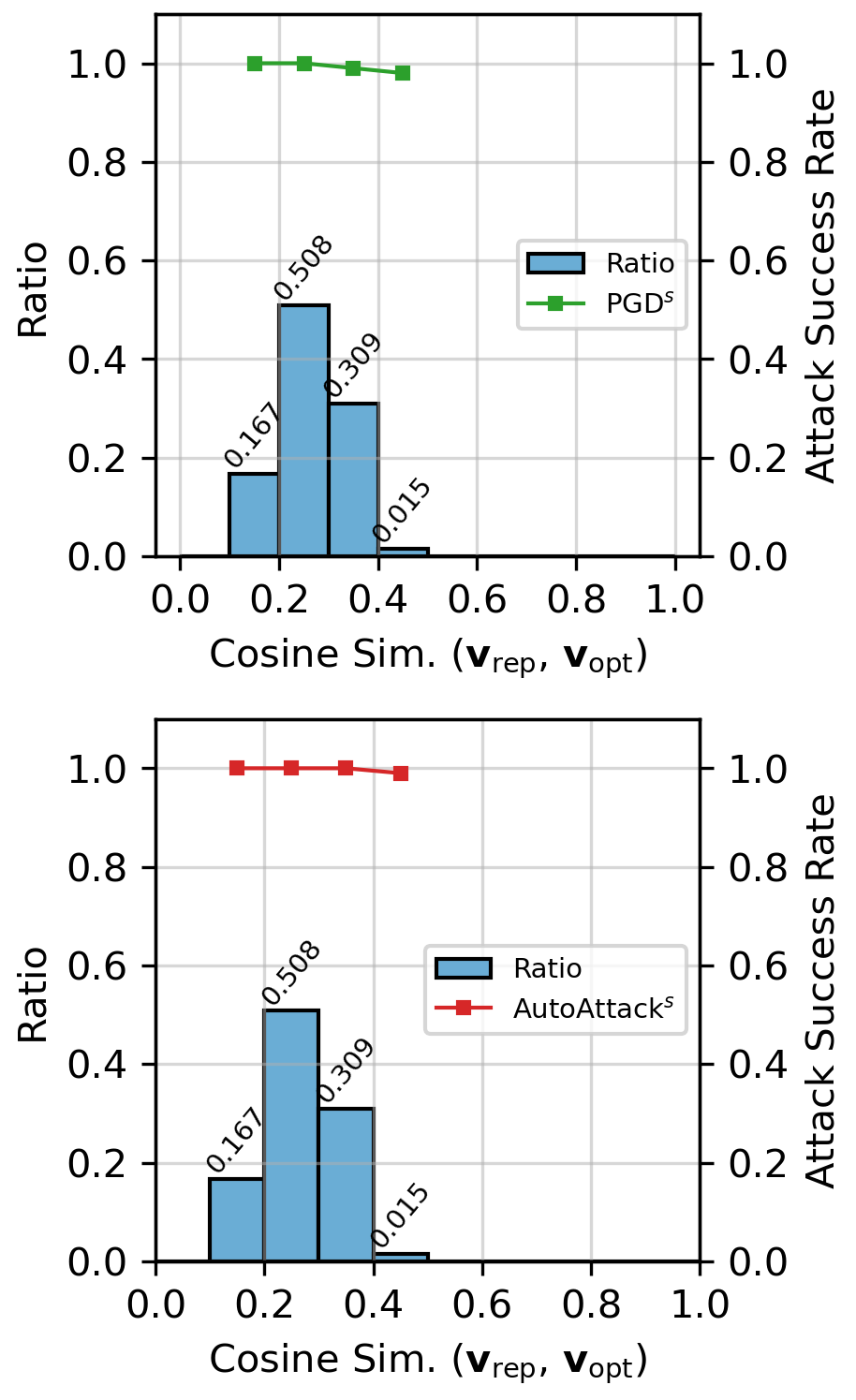}}
    }
    \subfloat[Label smoothing]
    {
        {\includegraphics[width=0.22\columnwidth]{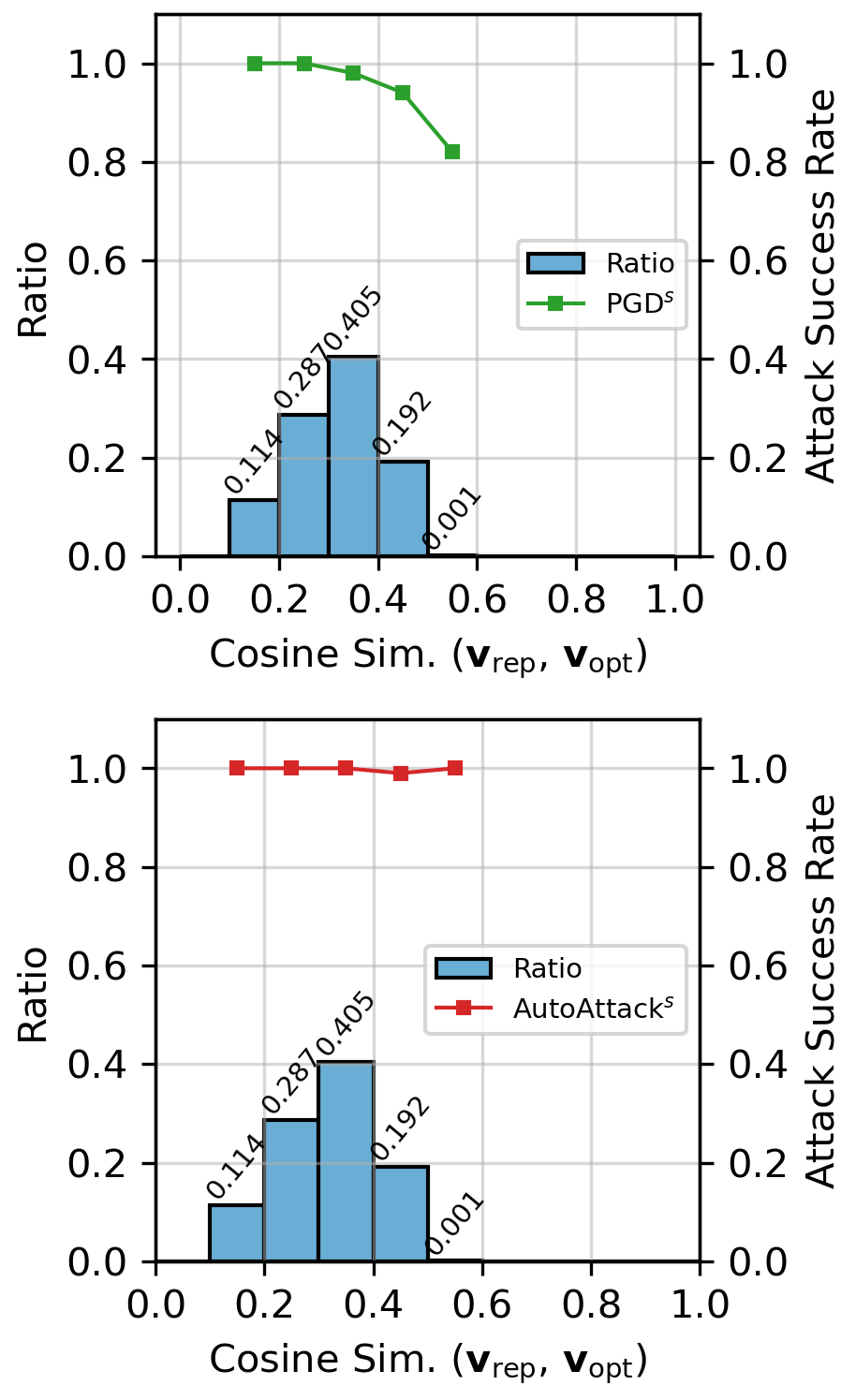}}
    }
    \subfloat[Mixup]
    {
        {\includegraphics[width=0.22\columnwidth]{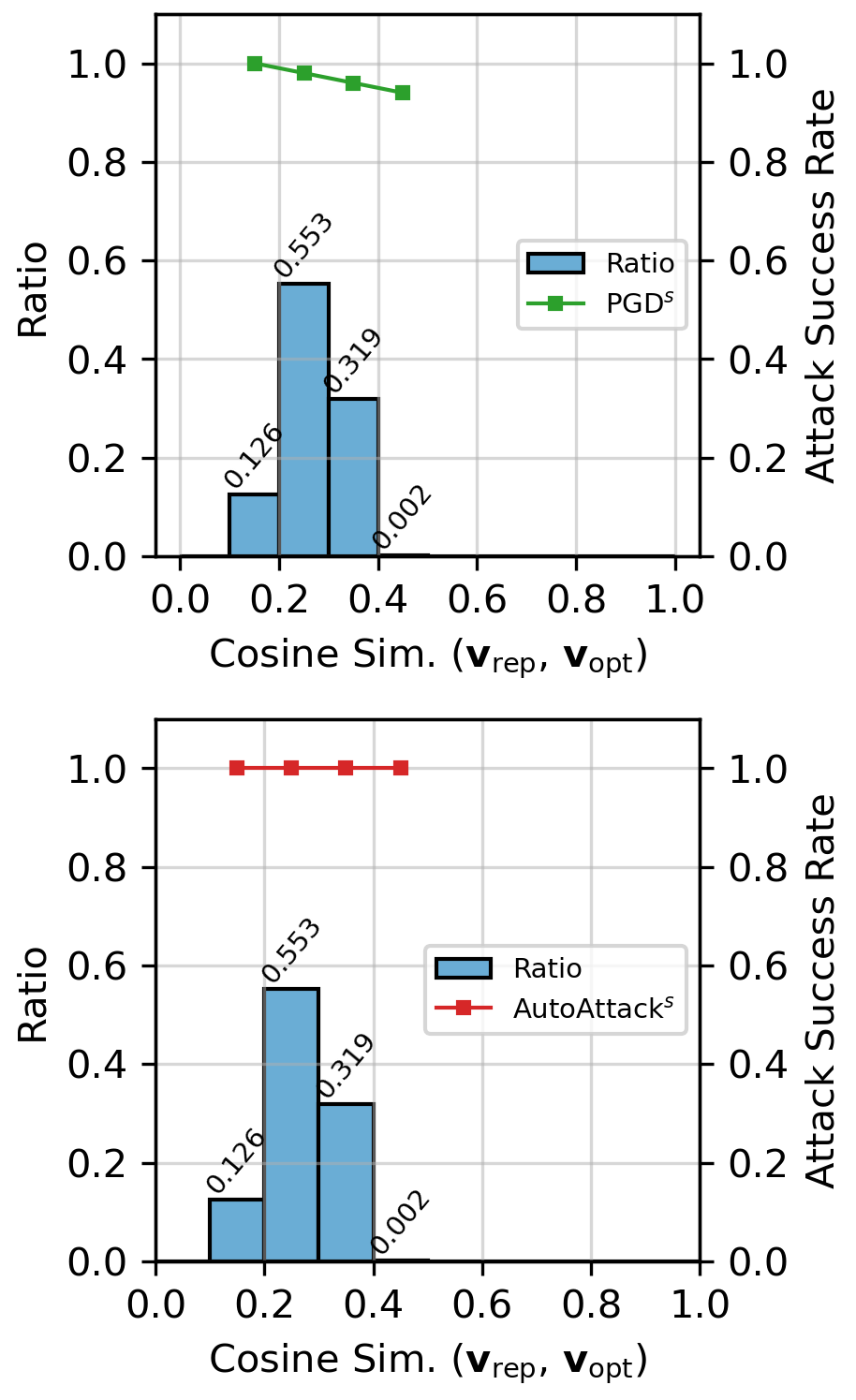}}
    }
    \subfloat[CutMix]
    {
        {\includegraphics[width=0.22\columnwidth]{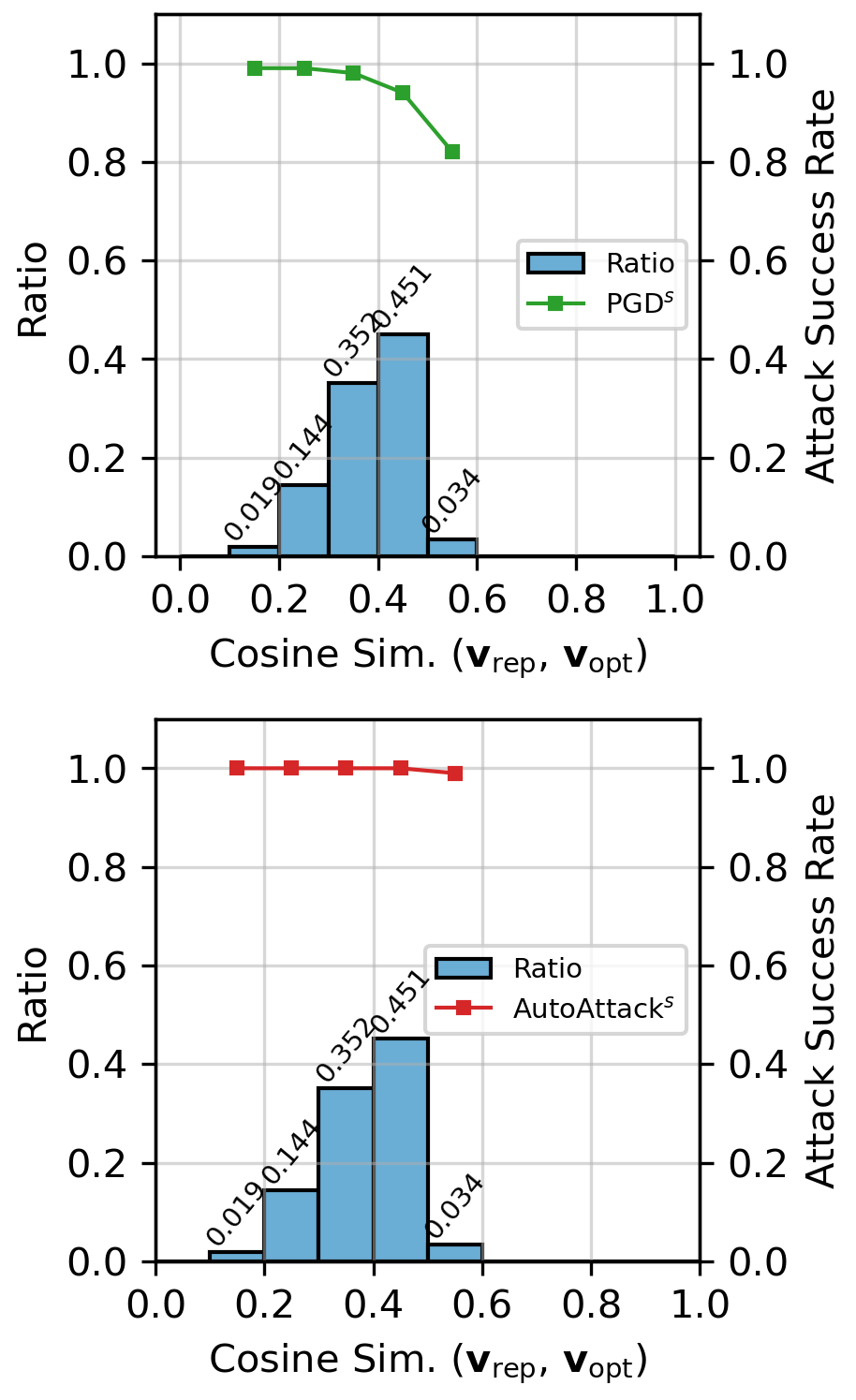}}
    }

    \caption{
    Evaluation results of ResNet50 on the ImageNet validation data.
    We show histograms of cosine similarities of features to class centers, along with the attack success rates for PGD$^\text{s}$ (top) and AutoAttack$^\text{s}$ (bottom) for each bin (hyperparameters settings for the attacks can be found in Section~\ref{sec:4_4}).
    }
    \label{fig:stronger_settings1}
\end{figure*}

\begin{figure*}
    \centering
    \subfloat[Baseline]
    {
        {\includegraphics[width=0.22\columnwidth]{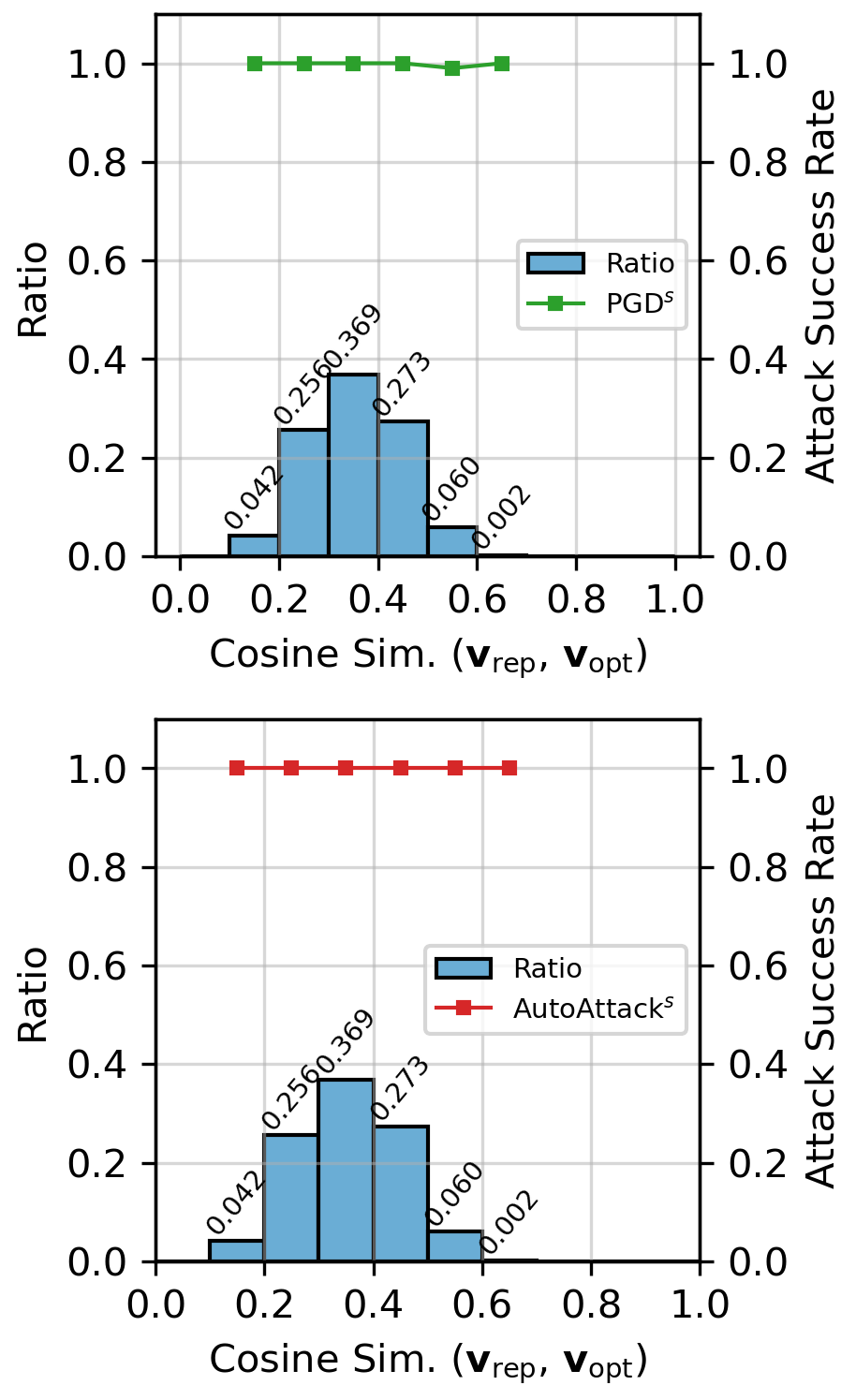}}
    }
    \subfloat[Label smoothing]
    {
        {\includegraphics[width=0.22\columnwidth]{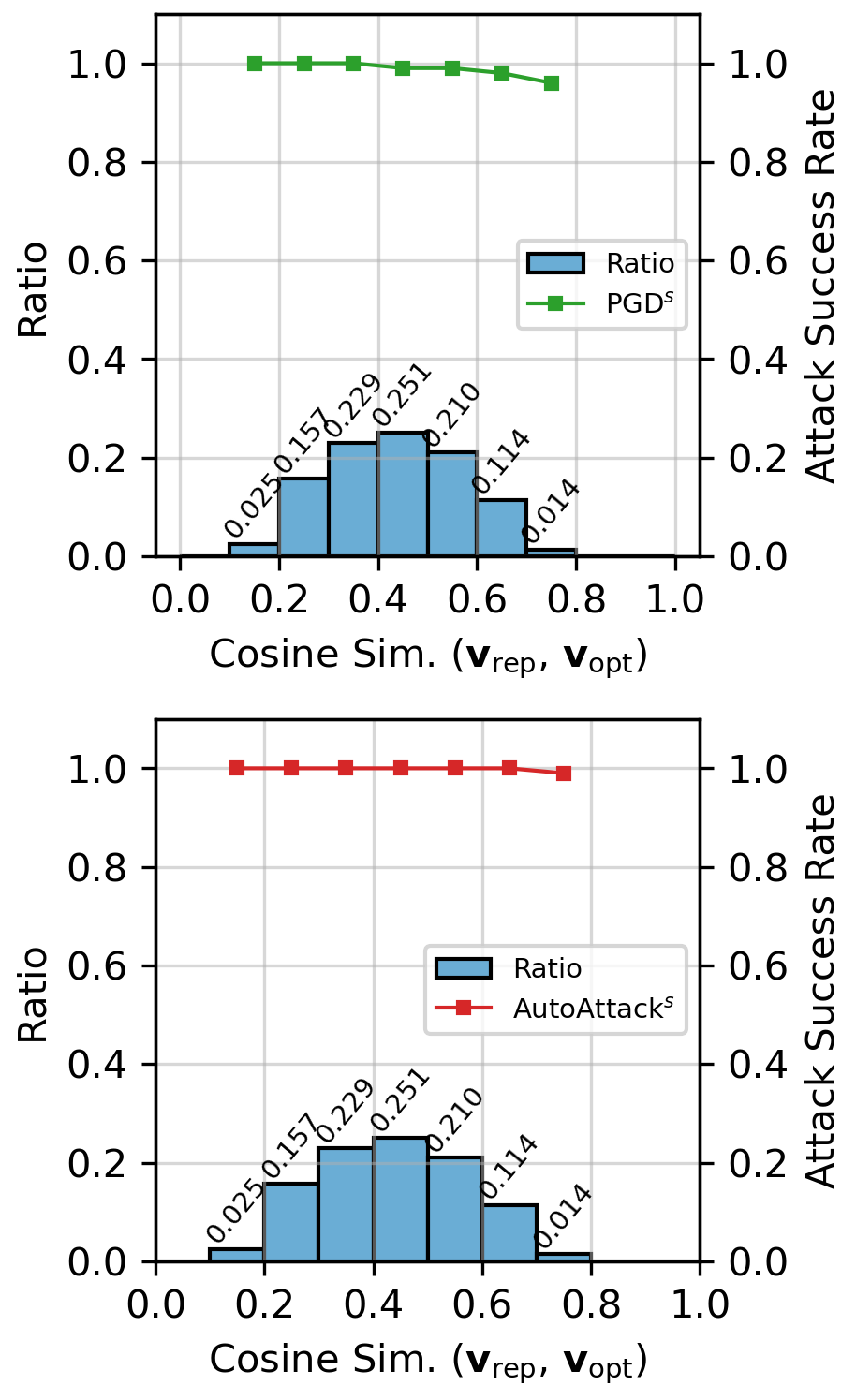}}
    }
    \subfloat[Mixup]
    {
        {\includegraphics[width=0.22\columnwidth]{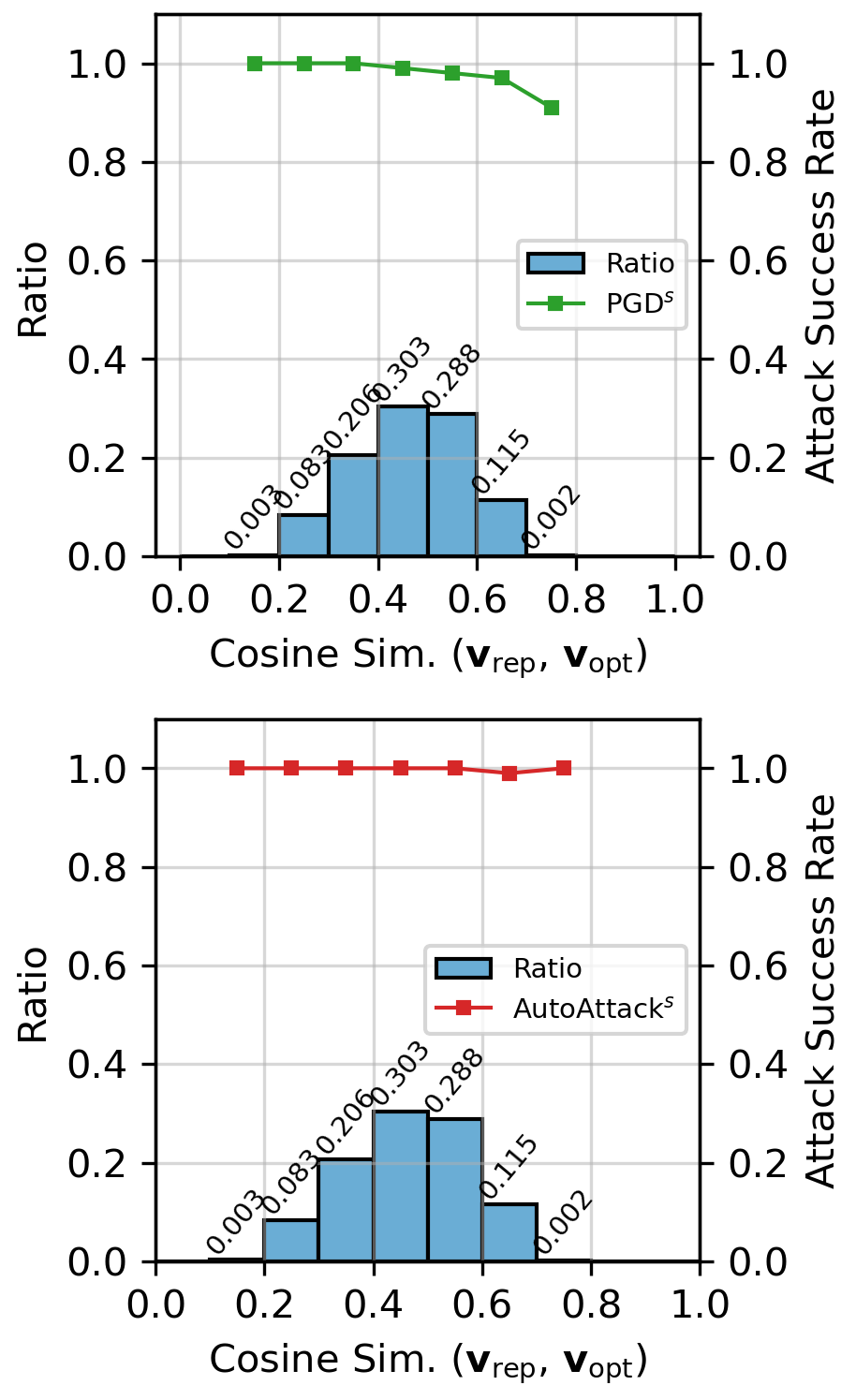}}
    }
    \subfloat[CutMix]
    {
        {\includegraphics[width=0.22\columnwidth]{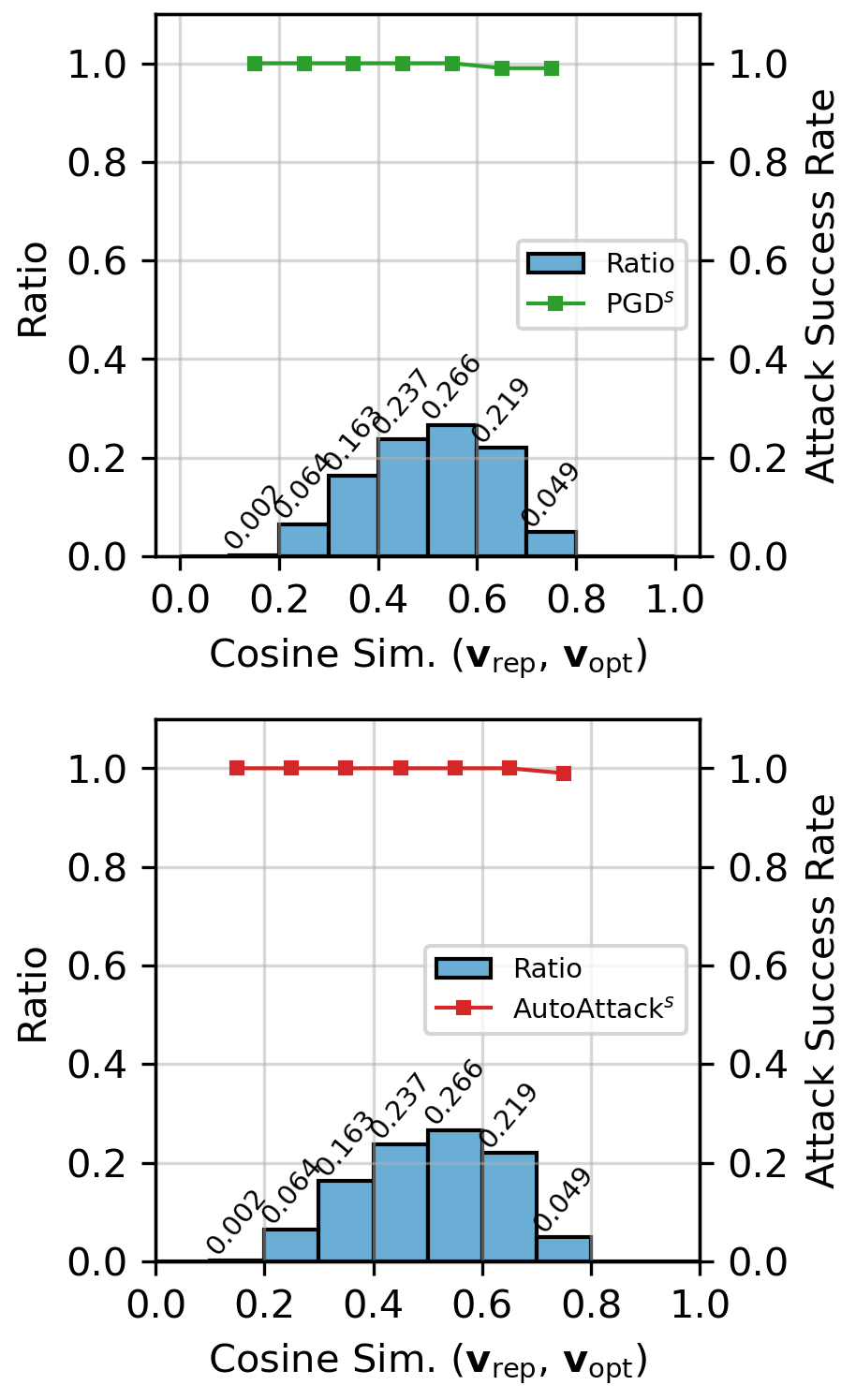}}
    }

    \caption{
    Evaluation results of Swin-T on the ImageNet validation data.
    We show histograms of cosine similarities of features to class centers, along with the attack success rates for PGD$^\text{s}$ (top) and AutoAttack$^\text{s}$ (bottom) for each bin (hyperparameters settings for the attacks can be found in Section~\ref{sec:4_4}).
    }
    \label{fig:stronger_settings2}
\end{figure*}

\begin{figure*}
    \centering
    \subfloat[Baseline]
    {
        {\includegraphics[width=0.22\columnwidth]{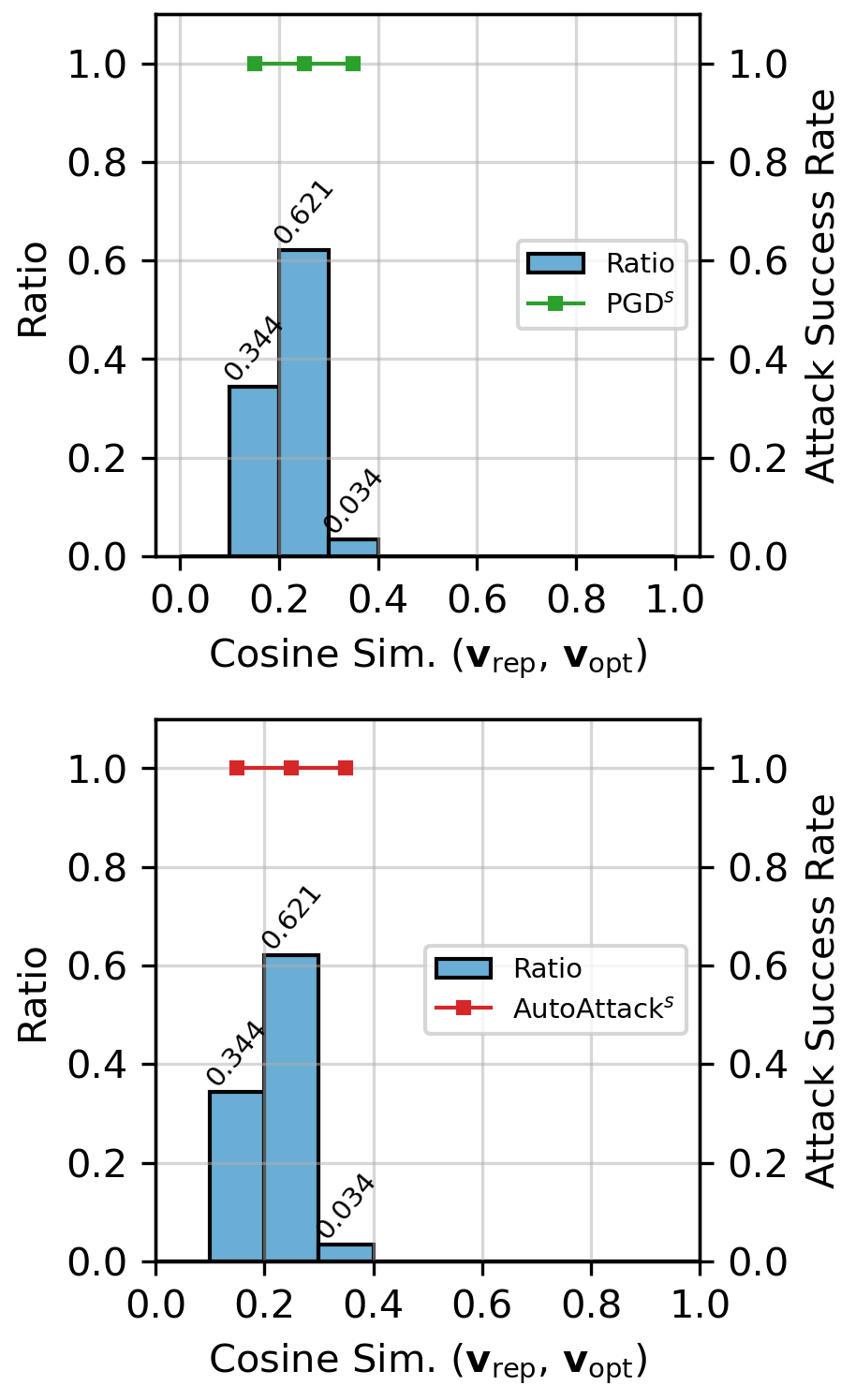}}
    }
    \subfloat[Label smoothing]
    {
        {\includegraphics[width=0.22\columnwidth]{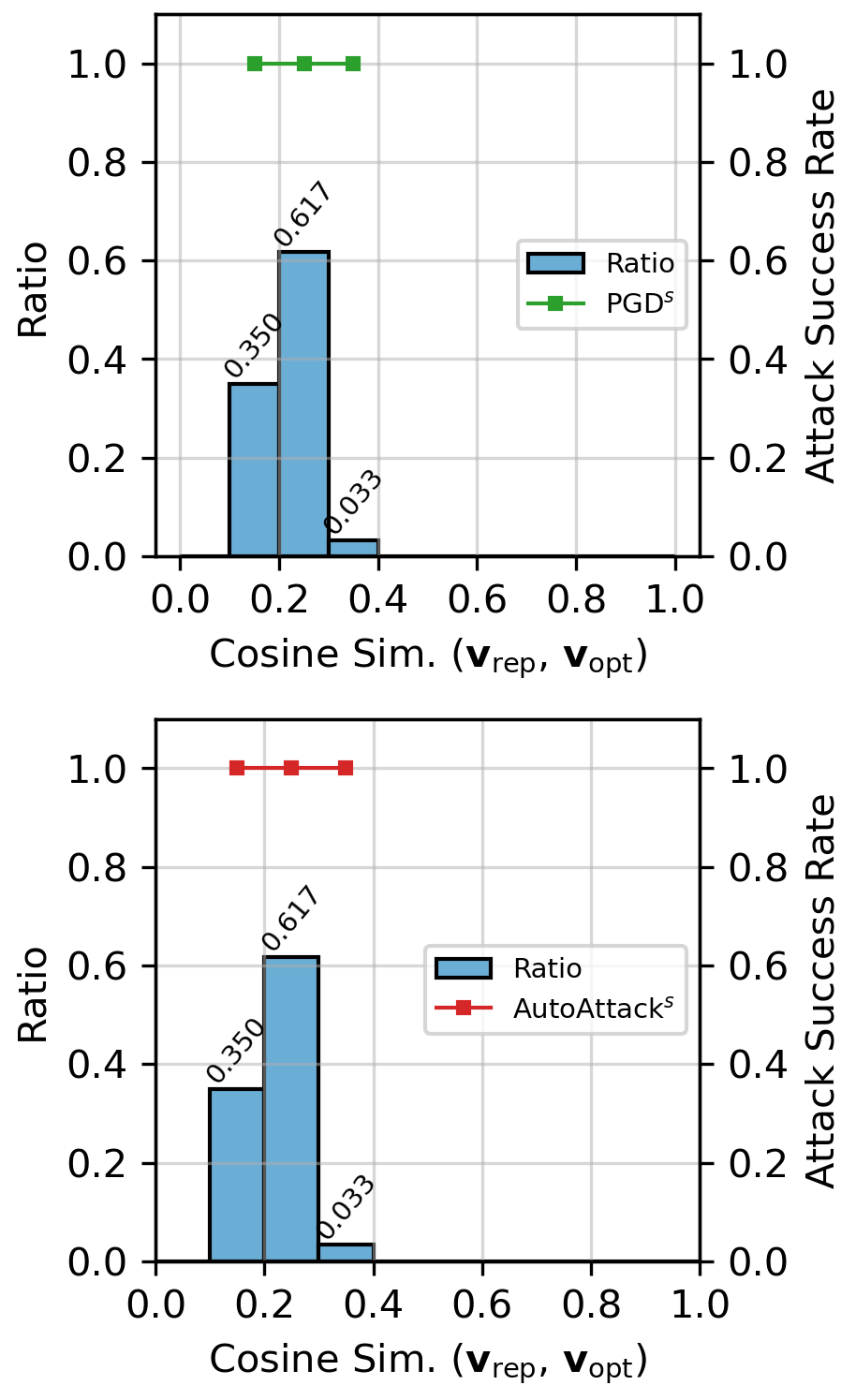}}
    }
    \subfloat[Mixup]
    {
        {\includegraphics[width=0.22\columnwidth]{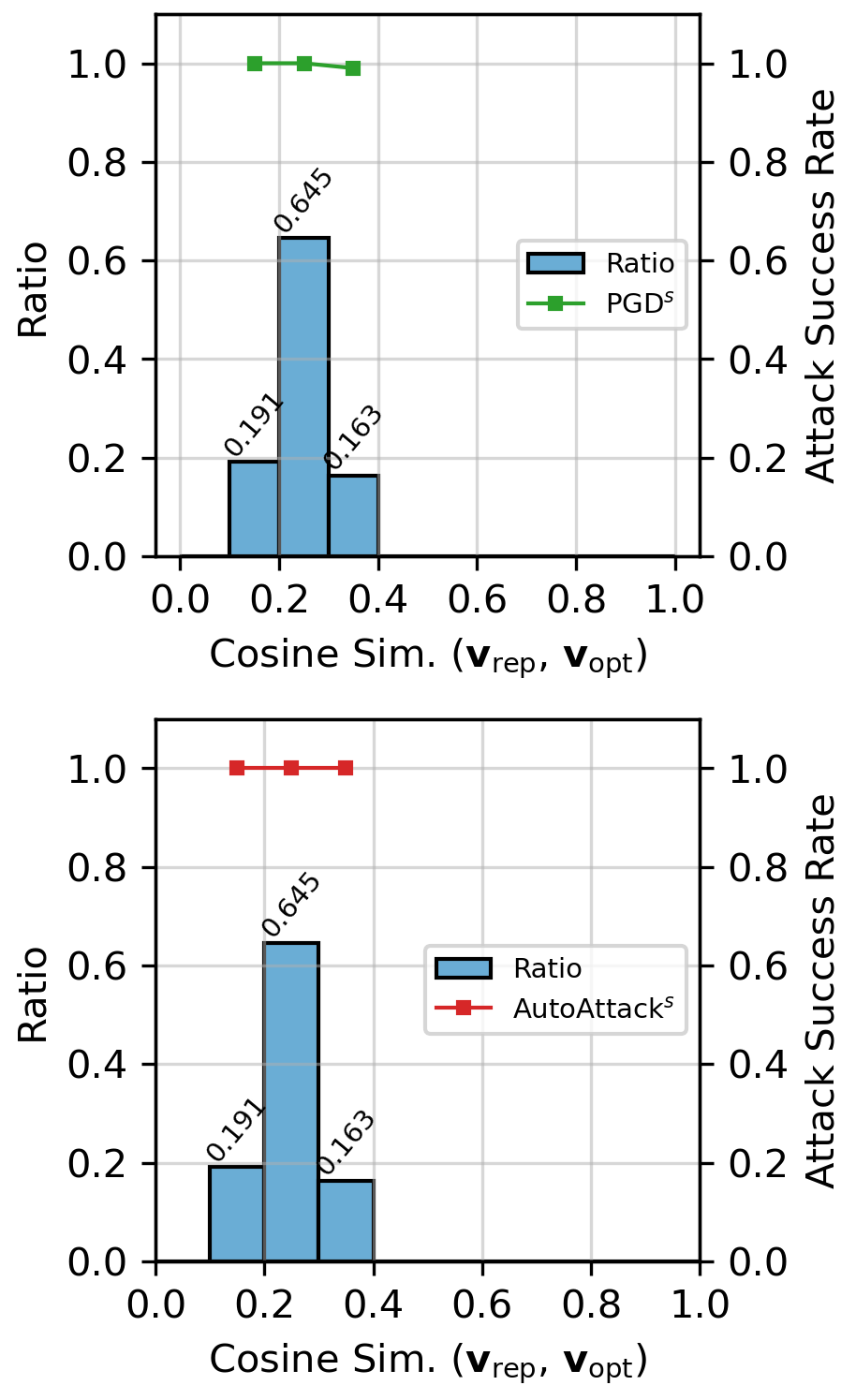}}
    }
    \subfloat[CutMix]
    {
        {\includegraphics[width=0.22\columnwidth]{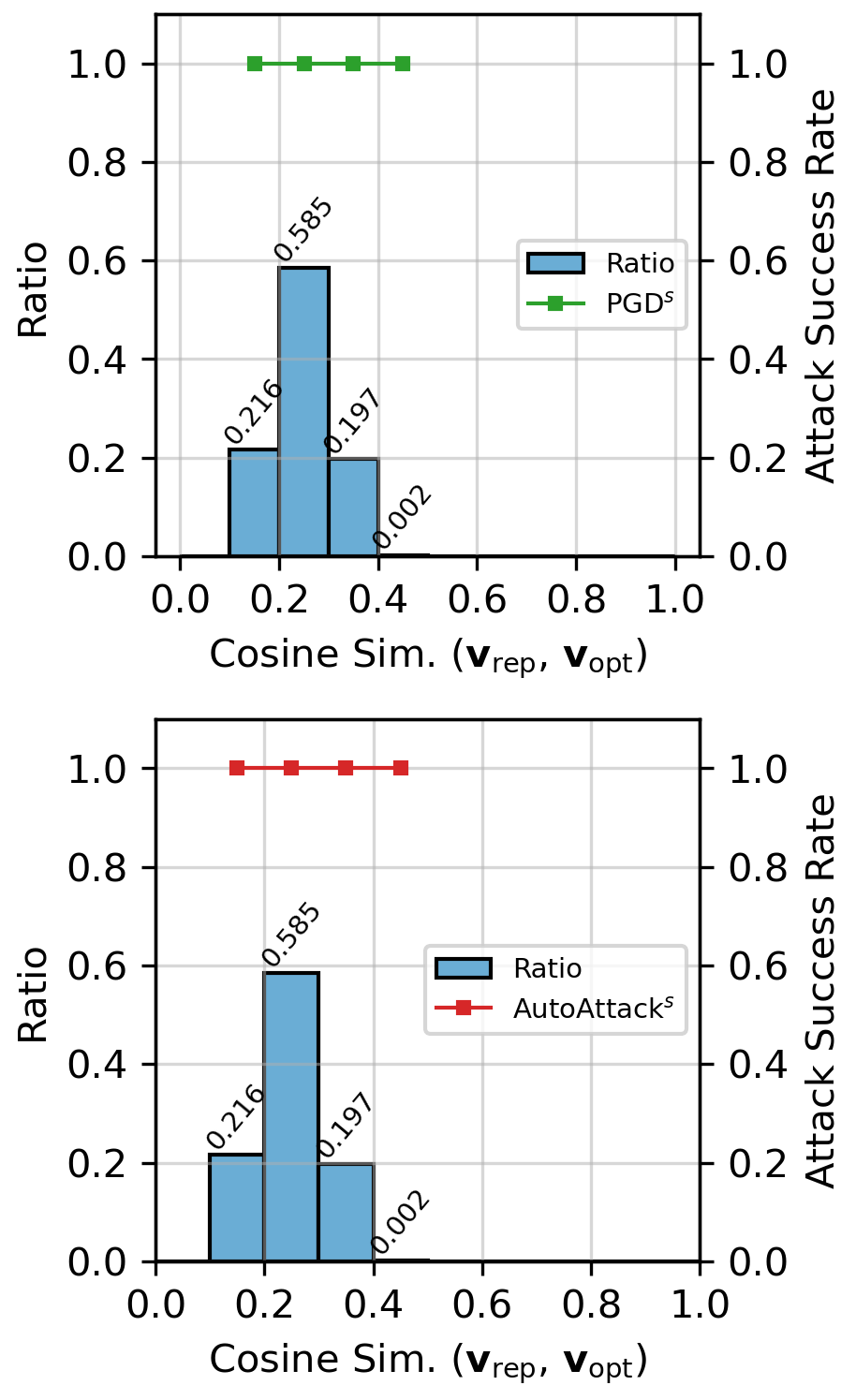}}
    }

    \caption{
    Evaluation results of MobileNetV2 on the ImageNet validation data.
    We show histograms of cosine similarities of features to class centers, along with the attack success rates for PGD$^\text{s}$ (top) and AutoAttack$^\text{s}$ (bottom) for each bin (hyperparameters settings for the attacks can be found in Section~\ref{sec:4_4}).
    }
    \label{fig:stronger_settings3}
\end{figure*}

\begin{figure*}
    \centering
    \subfloat[Baseline]
    {
        {\includegraphics[width=0.22\columnwidth]{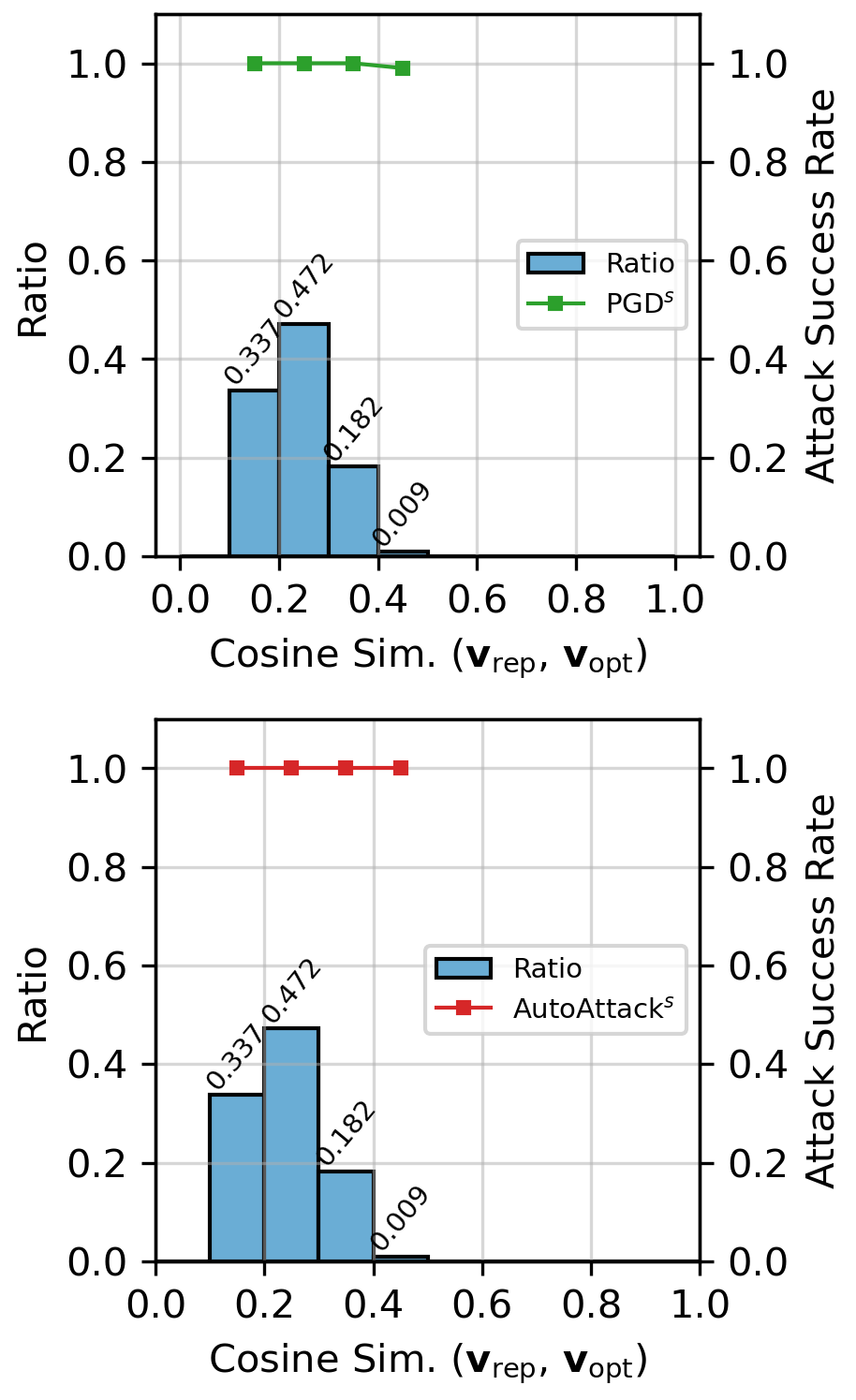}}
    }
    \subfloat[Label smoothing]
    {
        {\includegraphics[width=0.22\columnwidth]{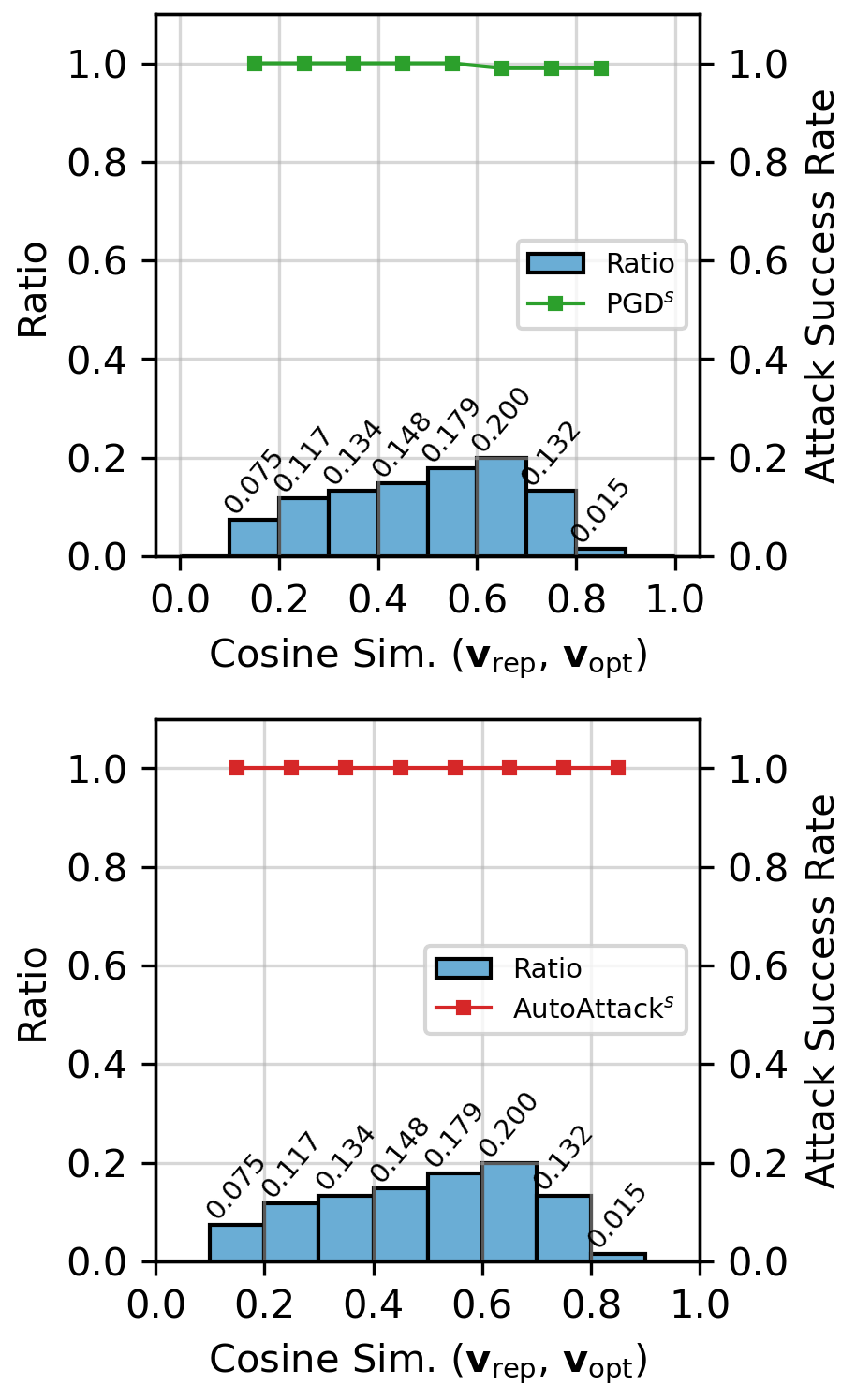}}
    }
    \subfloat[Mixup]
    {
        {\includegraphics[width=0.22\columnwidth]{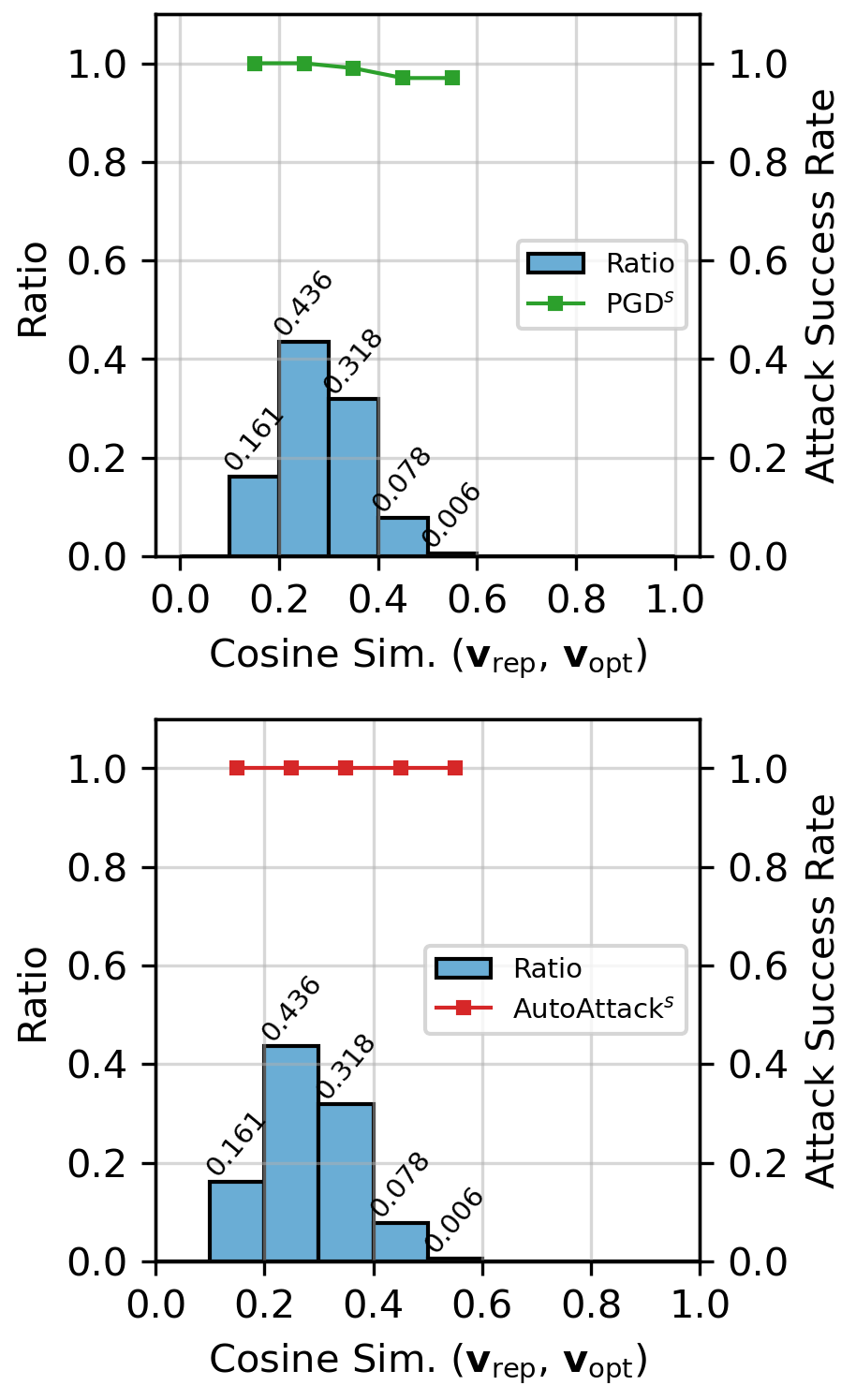}}
    }

    \caption{
    Evaluation results of ConvNeXt-T on the ImageNet validation data.
    We show histograms of cosine similarities of features to class centers, along with the attack success rates for PGD$^\text{s}$ (top) and AutoAttack$^\text{s}$ (bottom) for each bin (hyperparameters settings for the attacks can be found in Section~\ref{sec:4_4}).
    }
    \label{fig:stronger_settings4}
\end{figure*}

\begin{figure*}
    \centering
    \subfloat[{\begin{tabular}{c}MobileNetV2 \\ V1\end{tabular}}]
    {
        {\includegraphics[width=0.22\columnwidth]{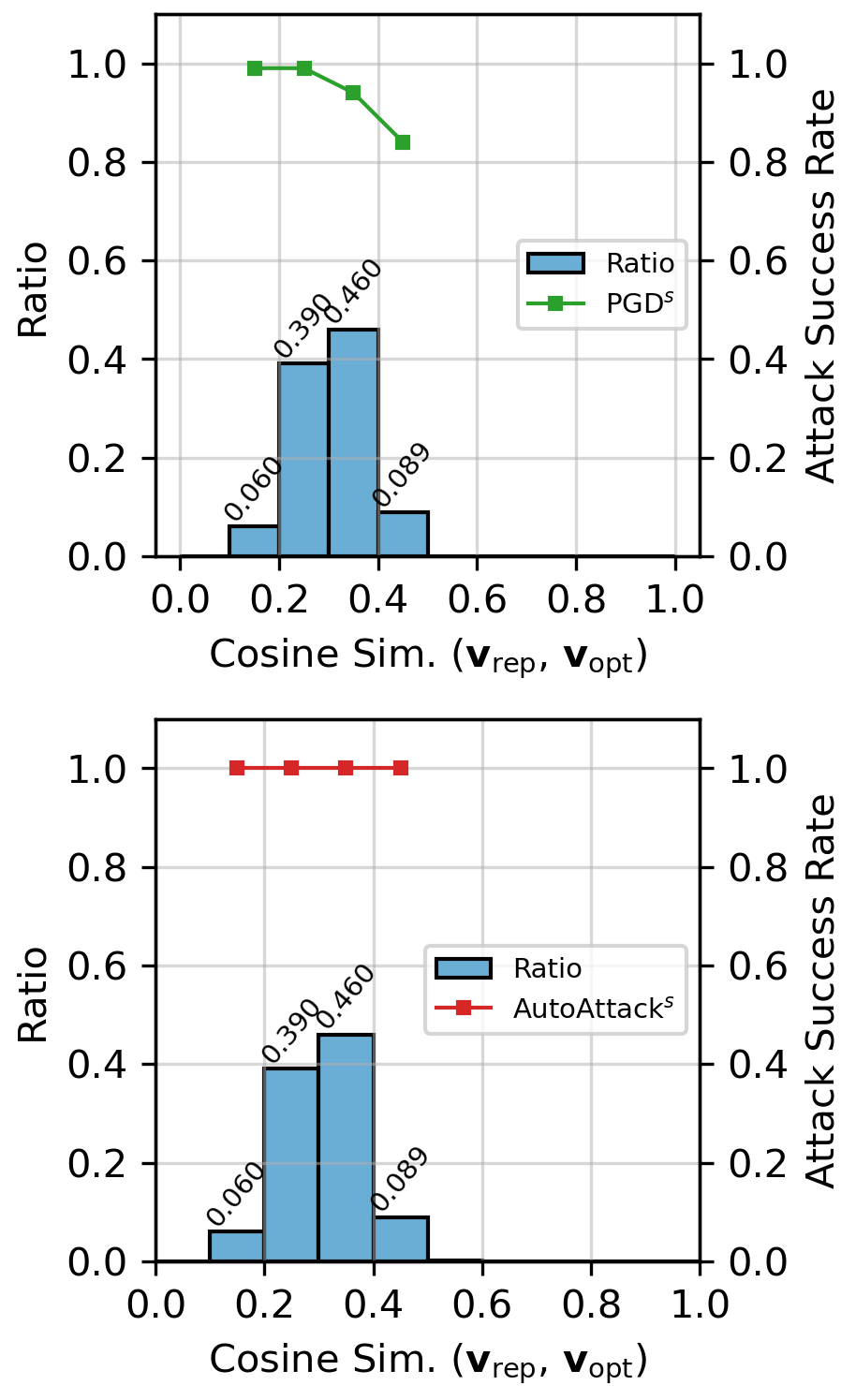}}
    }
    \subfloat[{\begin{tabular}{c}MobileNetV2 \\ V2\end{tabular}}]
    {
        {\includegraphics[width=0.22\columnwidth]{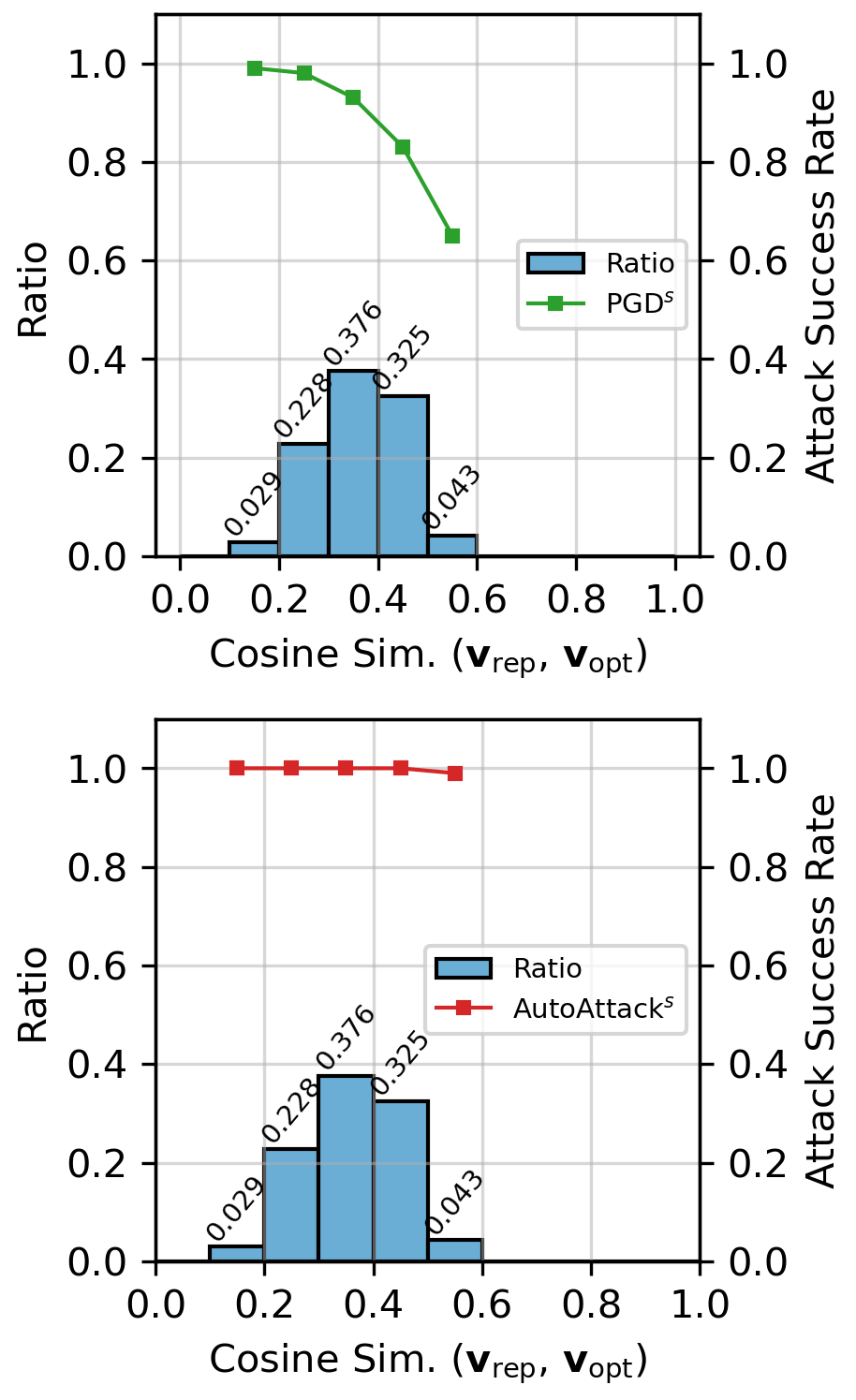}}
    }
    \subfloat[{\begin{tabular}{c}EfficientNet-B1 \\ V1\end{tabular}}]
    {
        {\includegraphics[width=0.22\columnwidth]{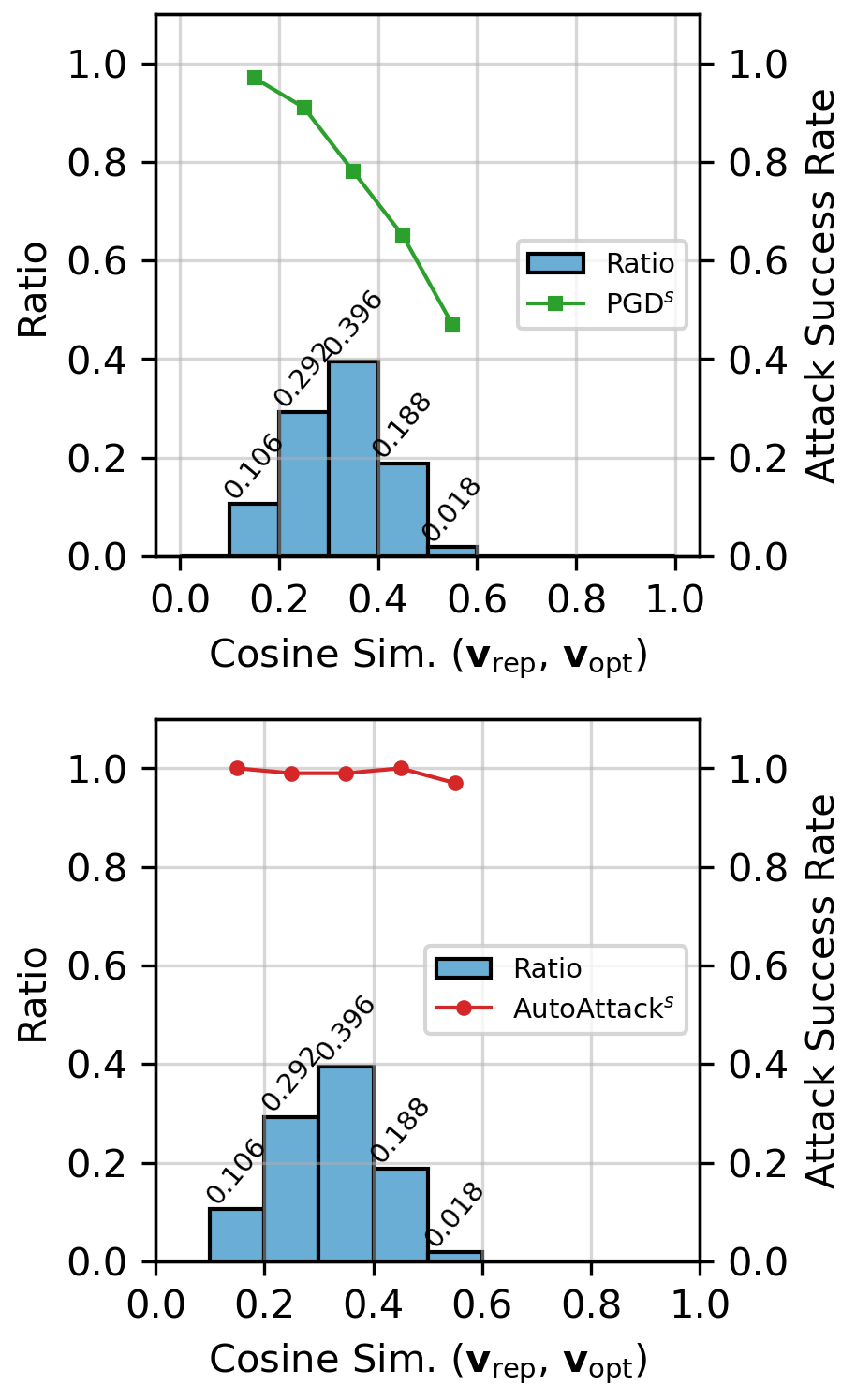}}
    }
    \subfloat[{\begin{tabular}{c}EfficientNet-B1 \\ V2\end{tabular}}]
    {
        {\includegraphics[width=0.22\columnwidth]{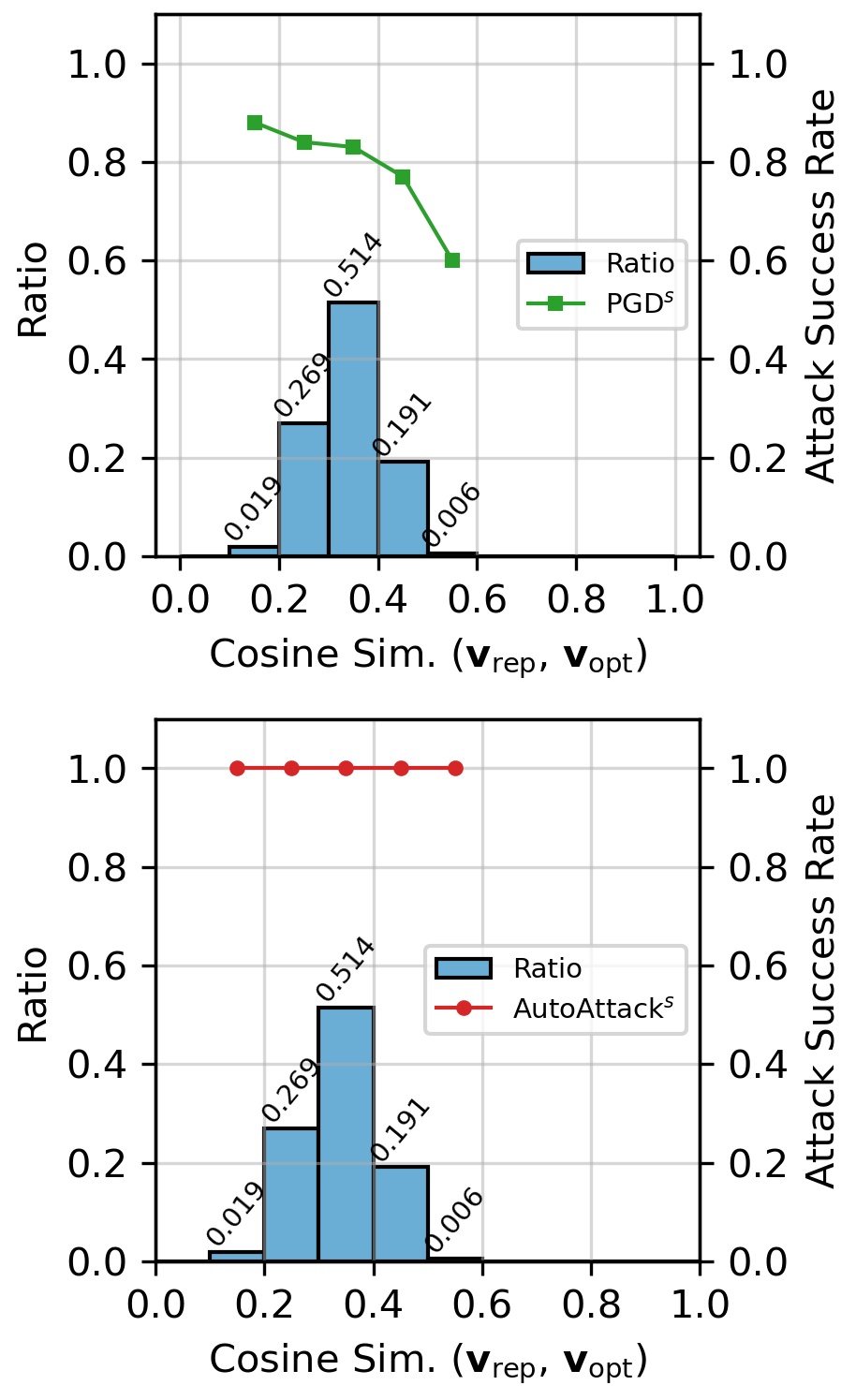}}
    }\hspace{0.0cm}
    \subfloat[{\begin{tabular}{c}ViT-B/16 \\ Swag Linear V1\end{tabular}}]
    {
        \includegraphics[width=0.22\columnwidth]{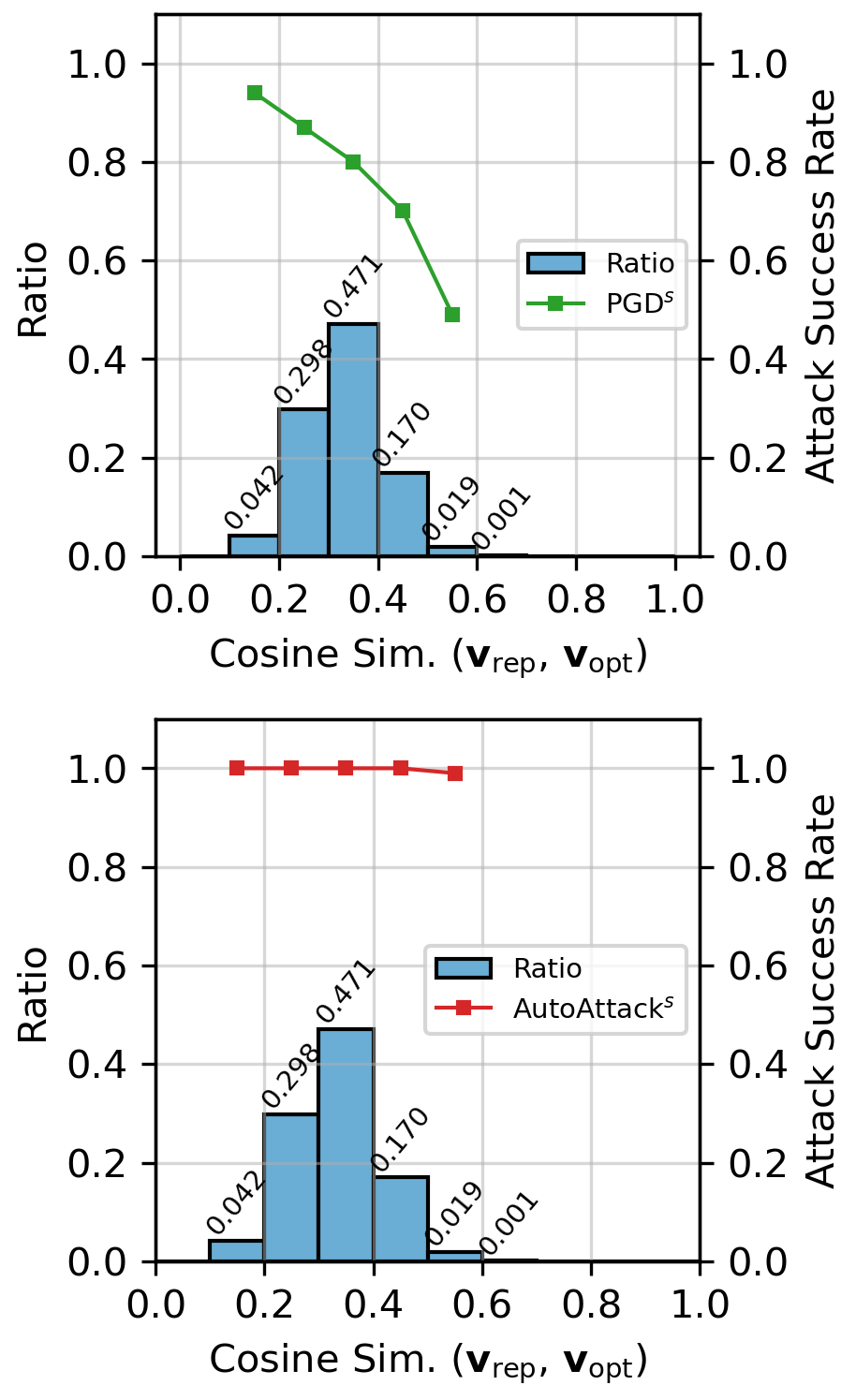}
    }
    \subfloat[{\begin{tabular}{c}ViT-B/16 \\ V1\end{tabular}}]
    {
        {\includegraphics[width=0.22\columnwidth]{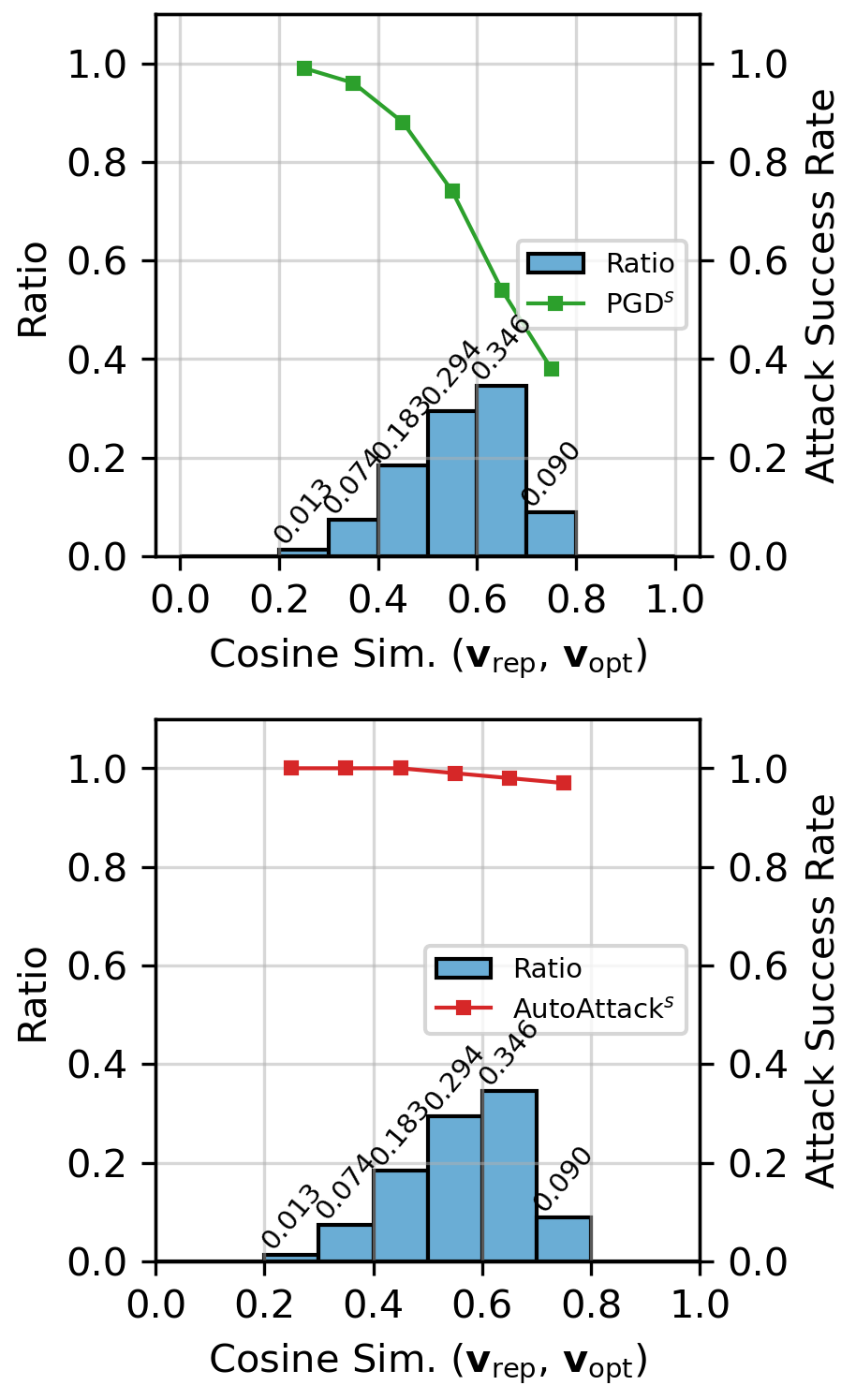}}
    }

    \caption{
    Evaluation results of MobileNetV2, EfficientNet-B1, and ViT-B/16  on the ImageNet validation data.
    We show histograms of cosine similarities of features to class centers, along with the attack success rates for PGD$^\text{s}$ (top) and AutoAttack$^\text{s}$ (bottom) for each bin (hyperparameters settings for the attacks can be found in Section~\ref{sec:4_4}).
    }
    \label{fig:stronger_settings5}
\end{figure*}

Figs.~\ref{fig:swin_t}-\ref{fig:13} present additional examples on how regularization affects feature distributions in the original representation space, across various models.
As discussed in Section~\ref{sec:4}, regularization reduces the RMS of features (top row), leading to less confident predictions (bottom row). Furthermore, the overall cosine similarity between features and class centers increases with regularization (middle row).
We show attack success rates for the stronger attack configurations (PGD$^\text{s}$ and AutoAttack$^\text{s}$) in Figs.~\ref{fig:stronger_settings1}-\ref{fig:stronger_settings5}.

In Tab.~\ref{tab:appendix_more_results}, we provide additional experimental results for MobileNetV2, EfficientNet-B1, and ViT-B/16 using pretrained weights from PyTorch.
In the PyTorch framework, these pretrained weights can be accessed using the names \texttt{IMAGENET1K\_V1}, \texttt{IMAGENET1K\_V2}, and \texttt{IMAGENET1K\_SWAG\_LINEAR\_V1}.
For MobileNetV2 and EfficientNet-B1, the V2 weights result from stronger regularization compared to V1, whereas for ViT-B/16, the V1 weights are derived from stronger regularization compared to Swag Linear V1.
Details on the regularization hyperparameters (using soft labels) used to train these weights can be found in Tab.~\ref{tab:appendix_pytorch_settings}.

Consistent with the findings in Section~\ref{sec:4}, applying stronger regularization reduces the RMS of the features and increases the cosine similarity between the features and class centers. 
Moreover, the models become more underconfident, and the attack success rates of both FGSM and PGD attacks decrease.

\section{Feature Scaling}
\label{appendix:feature_scaling}
\begin{figure*}
    \centering

    \subfloat[{\begin{tabular}{c}ResNet50 \\ Baseline \\ $S=1$ \\ Acc=76.07\%\end{tabular}}]
    {
        {\includegraphics[width=0.2\columnwidth]{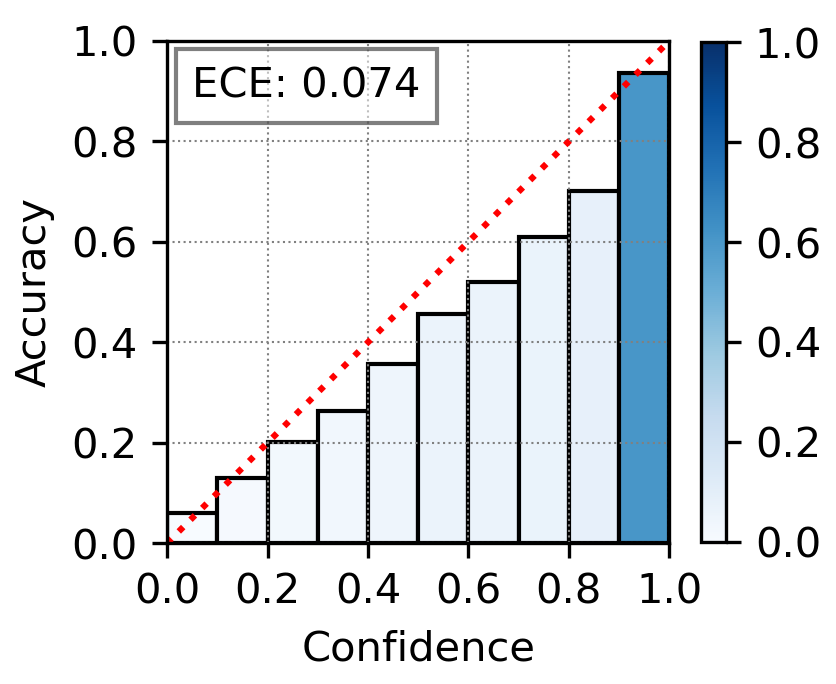}}
        \label{fig:14_a}
    }\hspace{0.0cm}
    \subfloat[{\begin{tabular}{c}ResNet50 \\ Baseline \\ $S=0.75$ \\ Acc=76.08\%\end{tabular}}]
    {
        {\includegraphics[width=0.2\columnwidth]{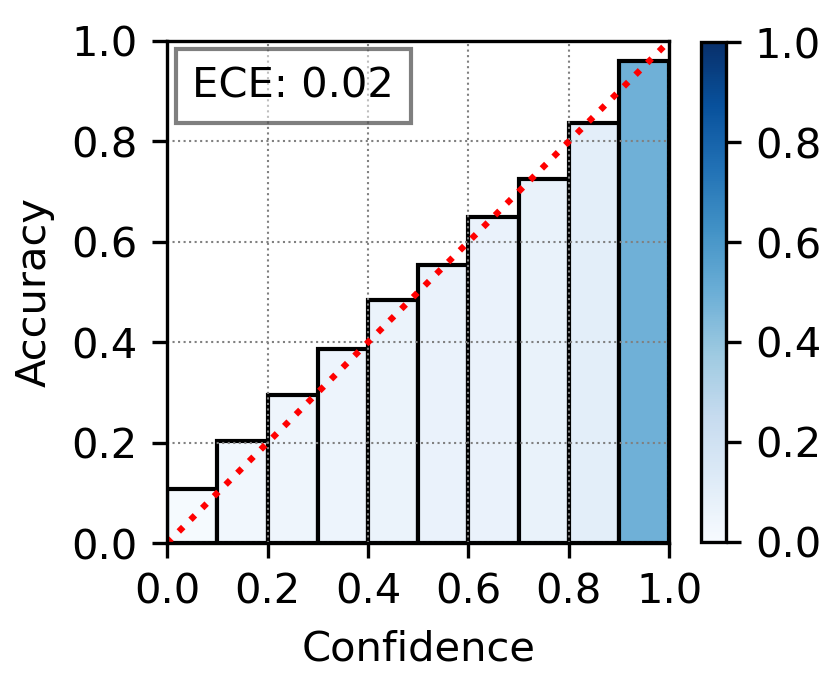}}
        \label{fig:14_b}
    }\hspace{0.0cm}
    \subfloat[{\begin{tabular}{c}ResNet50 \\ Mixup \\ $S=1$ \\ Acc=76.6\%\end{tabular}}]
    {
        {\includegraphics[width=0.2\columnwidth]{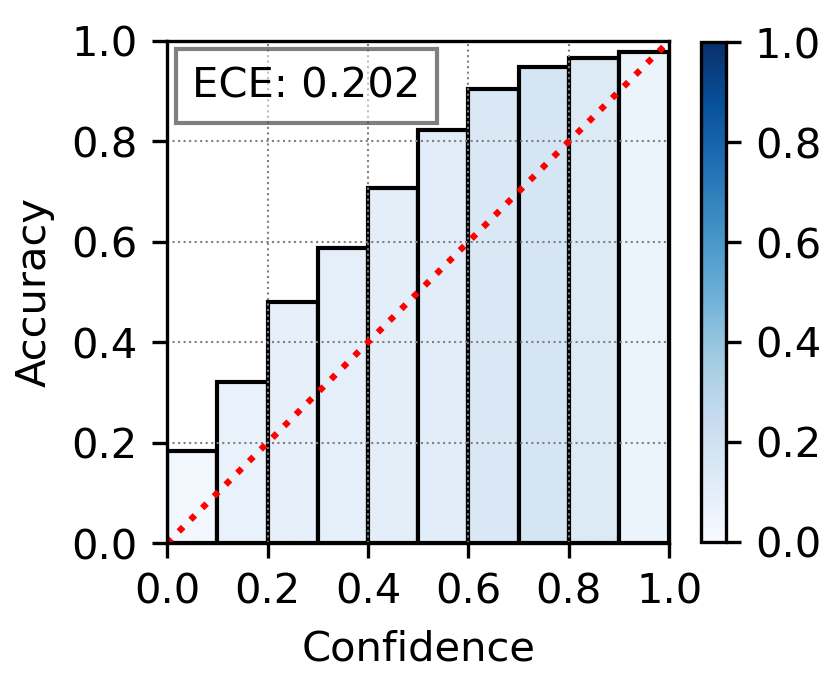}}
        \label{fig:14_c}
    }\hspace{0.0cm}
    \subfloat[{\begin{tabular}{c}ResNet50 \\ Mixup \\ $S=1.3$ \\ Acc=76.62\%\end{tabular}}]
    {
        {\includegraphics[width=0.2\columnwidth]{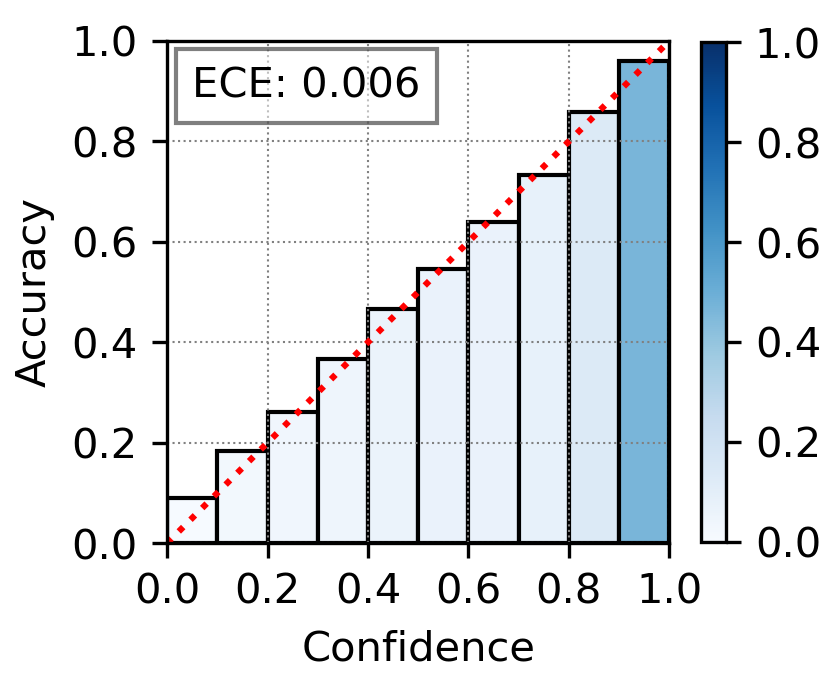}}
        \label{fig:14_d}
    }

    \subfloat[{\begin{tabular}{c}Swin-T \\ Baseline \\ $S=1$ \\ Acc=76.07\%\end{tabular}}]
    {
        {\includegraphics[width=0.2\columnwidth]{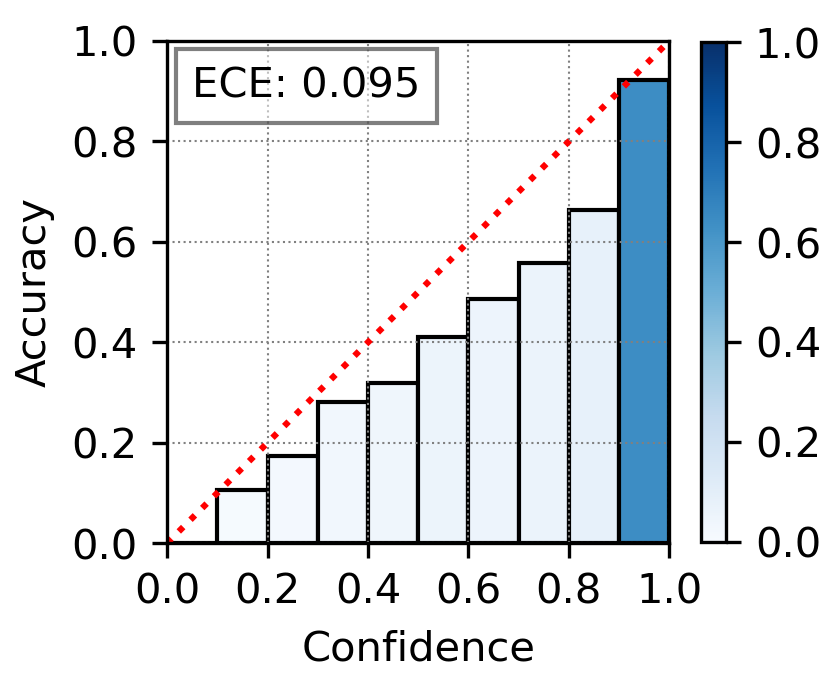}}
        \label{fig:14_e}
    }\hspace{0.0cm}
    \subfloat[{\begin{tabular}{c}Swin-T \\ Baseline \\ $S=0.7$ \\ Acc=76.08\%\end{tabular}}]
    {
        {\includegraphics[width=0.2\columnwidth]{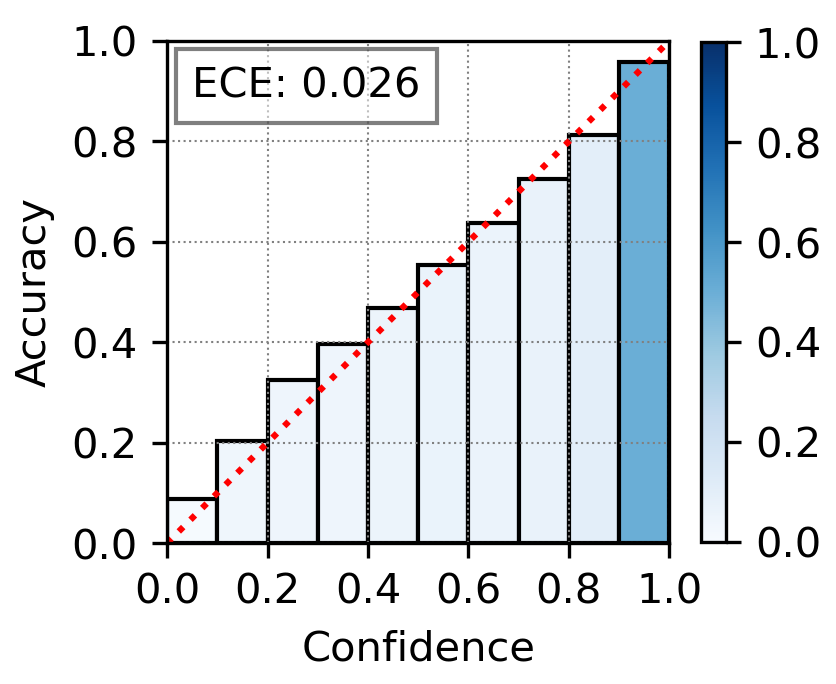}}
        \label{fig:14_f}
    }\hspace{0.0cm}
    \subfloat[{\begin{tabular}{c}Swin-T \\ Mixup \\ $S=1$ \\ Acc=78.19\%\end{tabular}}]
    {
        {\includegraphics[width=0.2\columnwidth]{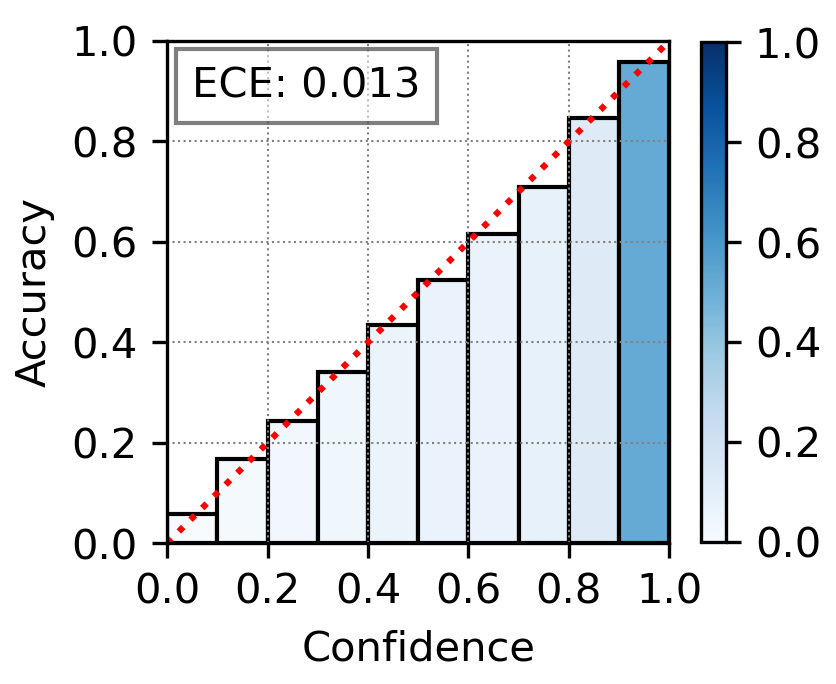}}
        \label{fig:14_g}
    }\hspace{0.0cm}
    \subfloat[{\begin{tabular}{c}Swin-T \\ Mixup \\ $S=0.95$ \\ Acc=78.19\%\end{tabular}}]
    {
        {\includegraphics[width=0.2\columnwidth]{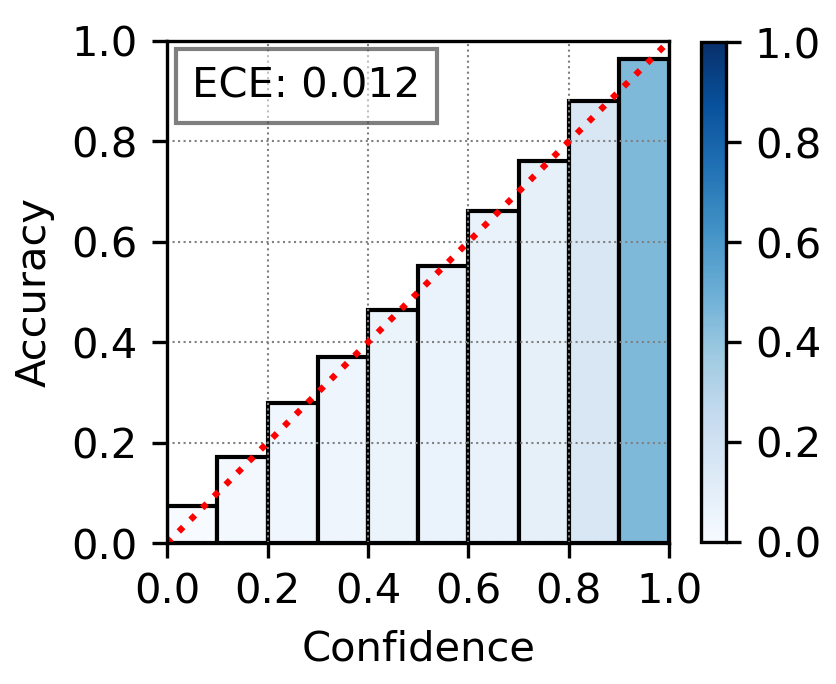}}
        \label{fig:14_h}
    }

    \subfloat[{\begin{tabular}{c}MobileNetV2 \\ Baseline \\ $S=1$ \\ Acc=70.84\%\end{tabular}}]
    {
        {\includegraphics[width=0.2\columnwidth]{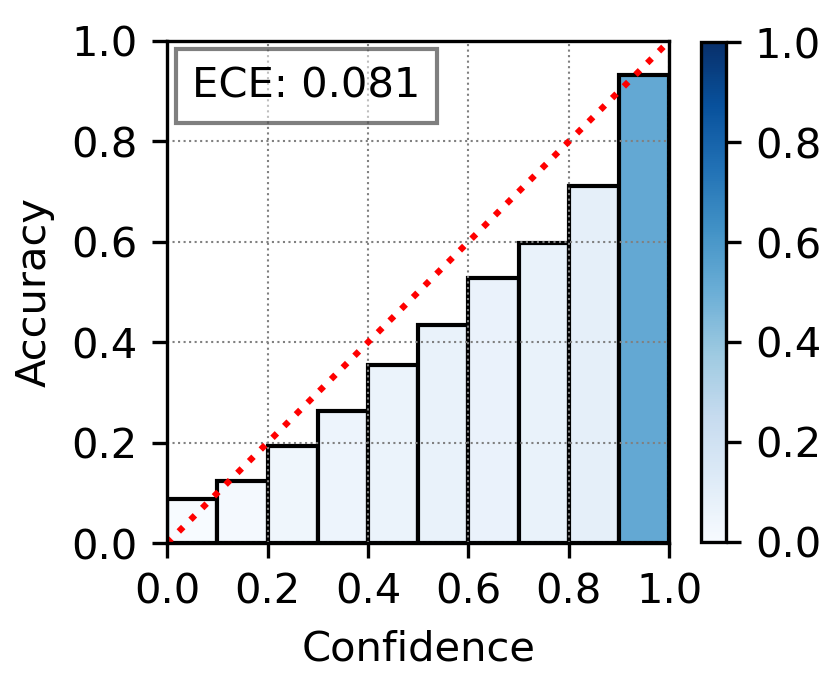}}
    }\hspace{0.0cm}
    \subfloat[{\begin{tabular}{c}MobileNetV2 \\ Baseline \\ $S=0.75$ \\ Acc=70.87\%\end{tabular}}]
    {
        {\includegraphics[width=0.2\columnwidth]{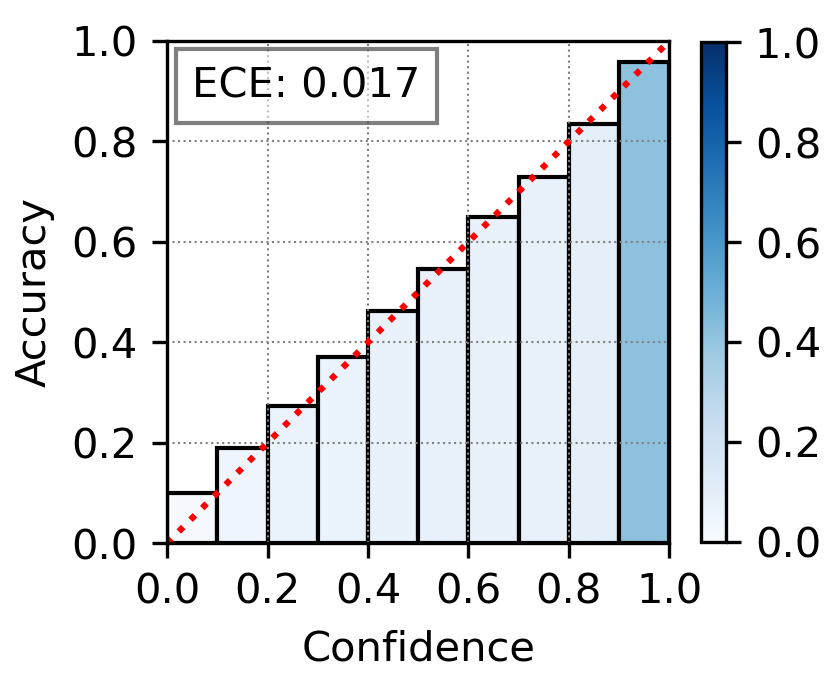}}
    }\hspace{0.0cm}
    \subfloat[{\begin{tabular}{c}MobileNetV2 \\ Mixup \\ $S=1$ \\ Acc=70.53\%\end{tabular}}]
    {
        {\includegraphics[width=0.2\columnwidth]{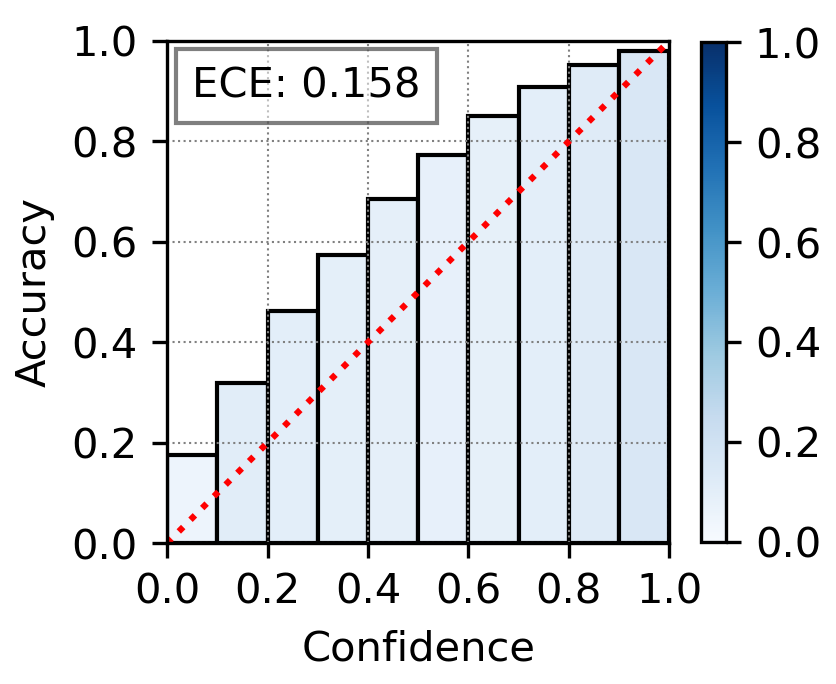}}
    }\hspace{0.0cm}
    \subfloat[{\begin{tabular}{c}MobileNetV2 \\ Mixup \\ $S=1.3$ \\ Acc=70.52\%\end{tabular}}]
    {
        {\includegraphics[width=0.2\columnwidth]{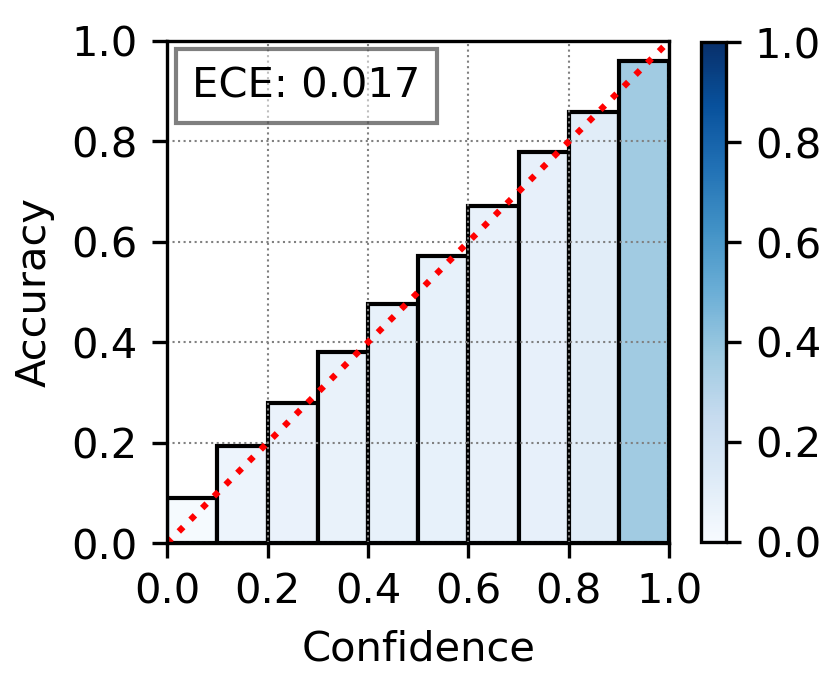}}
    }

    \subfloat[{\begin{tabular}{c}ConvNeXt-T \\ Baseline \\ $S=1$ \\ Acc=70.48\%\end{tabular}}]
    {
        {\includegraphics[width=0.2\columnwidth]{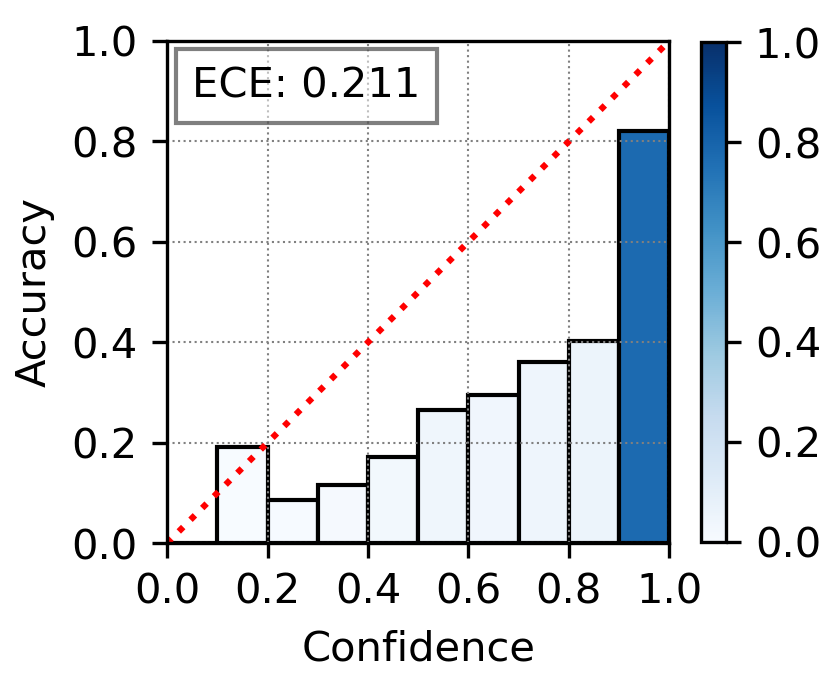}}
    }\hspace{0.0cm}
    \subfloat[{\begin{tabular}{c}ConvNeXt-T \\ Baseline \\ $S=0.35$ \\ Acc=70.0\%\end{tabular}}]
    {
        {\includegraphics[width=0.2\columnwidth]{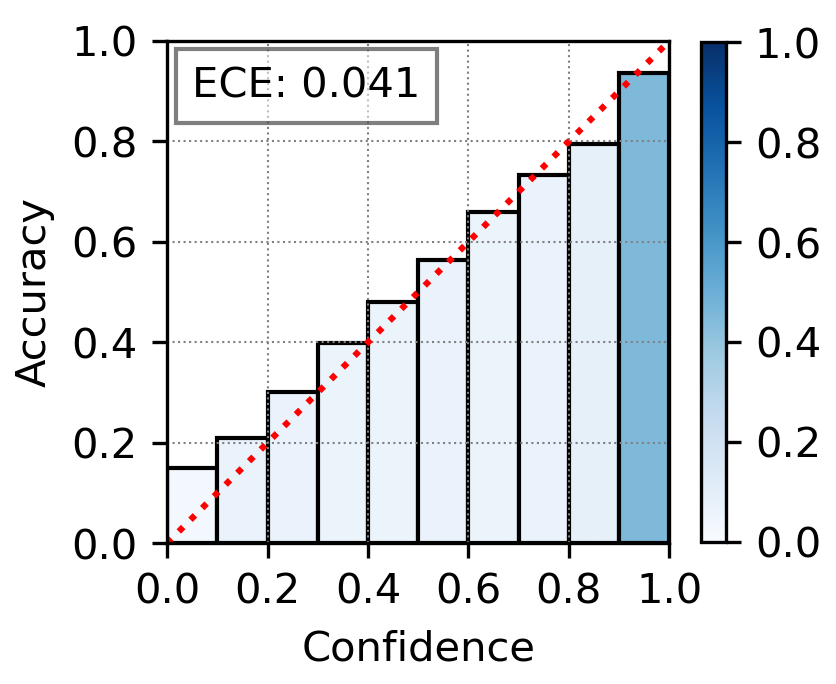}}
    }\hspace{0.0cm}
    \subfloat[{\begin{tabular}{c}ConvNeXt-T \\ Mixup \\ $S=1$ \\ Acc=77.97\%\end{tabular}}]
    {
        {\includegraphics[width=0.2\columnwidth]{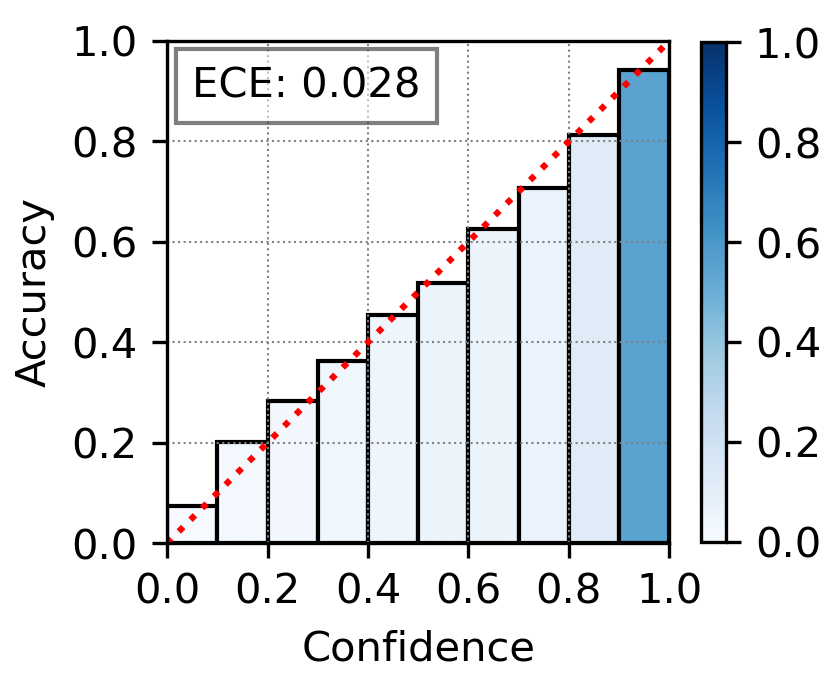}}
    }\hspace{0.0cm}
    \subfloat[{\begin{tabular}{c}ConvNeXt-T \\ Mixup \\ $S=0.95$ \\ Acc=77.97\%\end{tabular}}]
    {
        {\includegraphics[width=0.2\columnwidth]{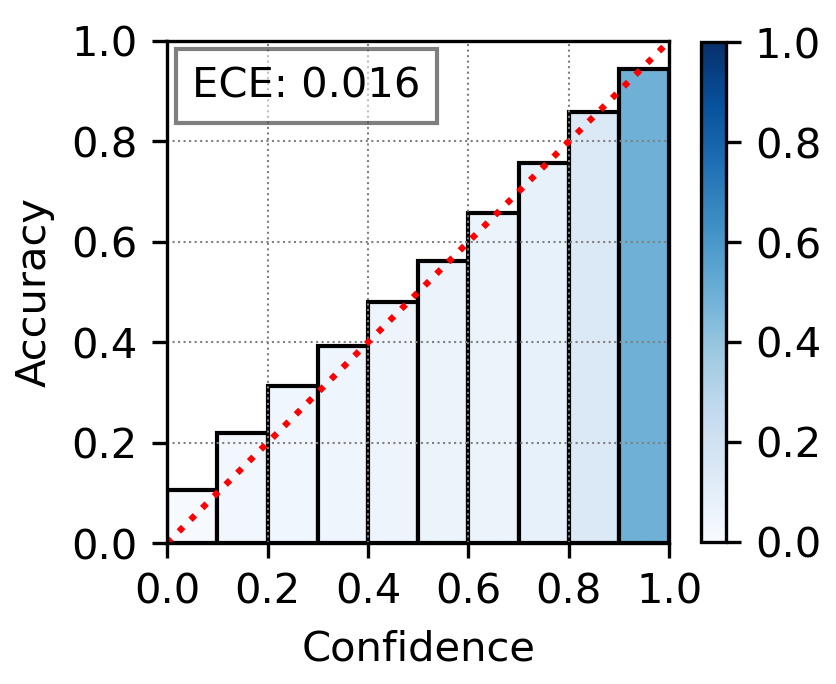}}
    }

    \caption{Calibration performances before and after manual feature scaling. $S$ represents the scaling factor, and Acc represents the accuracy on the ImageNet validation data.}
    \label{fig:14}
\end{figure*}

In Section~\ref{sec:4_3}, we show that regularization using soft labels has an effect of scaling down of features. Here, we examine the possibility of manual feature scaling  for calibration after training without regularization. In Fig.~\ref{fig:14}, we show the accuracy and calibration performance of ResNet50, Swin-T, MobileNetV2, and ConvNeXt-T on the ImageNet validation set, before and after manually scaling features across various models and weights. $S=1$ (where $S$ indicates the scaling factor) represents the use of original features. It can be observed that manual feature scaling does not affect classification accuracy but can improve calibration due to its similarity to temperature scaling.



\newpage

\newpage

\newpage

\newpage

\end{document}